\documentclass[twoside,11pt]{article}

\usepackage{times}
\usepackage{fullpage}
\usepackage{amsfonts}
\usepackage{amsmath}
\usepackage{latexsym}
\usepackage{color}
\usepackage{graphics}
\usepackage{enumerate}
\usepackage{amstext}
\usepackage{newclude}
\usepackage{psfrag}

\usepackage{amsthm}

\usepackage[round]{natbib}

\usepackage{graphicx} 
\usepackage{subfig,epstopdf}
\usepackage{rotating}

\usepackage{algorithm}
\usepackage{algorithmic}

\usepackage{hyperref}

\usepackage{commonopts}
\usepackage{commonmacros}

\newtheorem{theorem}{Theorem}
\newtheorem{lemma}{Lemma}
\newtheorem{definition}{Definition}

\begin{document}

\title{Robust Hierarchical Clustering
\thanks{A preliminary version of this article appeared under the title \emph{Robust Hierarchical Clustering} in the Proceedings of the Twenty-Third Conference on Learning Theory, 2010.}
}


\author{Maria-Florina Balcan\thanks{\texttt{\small ninamf@cs.cmu.edu}. School of Computer Science, Carnegie Mellon University.}
\and Yingyu Liang\thanks{\texttt{\small yliang39@gatech.edu}. College of Computing, Georgia Institute of Technology.}
\and Pramod Gupta\thanks{\texttt{\small pramodg@google.com}. Google, Inc..}
}

\date{}

\maketitle


\begin{abstract}
One of the most widely used techniques for data clustering is
agglomerative clustering. Such algorithms have been long used
across many different fields ranging from computational biology
to social sciences to computer vision in part because their
output is easy to interpret. Unfortunately, it is well known,
however, that many of the classic agglomerative clustering
algorithms are not robust to noise. In this
paper we propose and analyze a new robust algorithm for
bottom-up agglomerative clustering. We show that our algorithm
can be used to cluster accurately in cases where the data
satisfies a number of natural properties and where the
traditional agglomerative algorithms fail. We also show how to
adapt our algorithm to the inductive setting where our given
data is only a small random sample of the entire data set.
Experimental evaluations on synthetic and real world data sets
show that our algorithm  achieves better
performance than other hierarchical algorithms in the presence
of noise.

\end{abstract}


\section{Introduction}
\label{sec:introclust}

Many data mining and machine learning applications ranging from computer vision to biology problems have recently faced an explosion of data. As a consequence it has become increasingly important to develop effective, accurate, robust to noise, fast, and general clustering algorithms, accessible to developers and researchers in a diverse range of areas.

One of the oldest and most commonly used methods for clustering data, widely used in many
scientific applications, is hierarchical clustering~\citep{gower67, aprezian01, CKVW06, hierarchysanjoy05, dudahart2000, ravik06, Jain81dubes.algorithms,jain1999data, Johnson, qcluster2005}. In hierarchical clustering
the goal is not to find a single partitioning of the data, but a hierarchy (generally represented by a
tree) of partitions which may reveal interesting structure in the data at multiple levels of granularity.
The most widely used hierarchical methods are the agglomerative clustering techniques; most of these
techniques start with a separate cluster for each point and then progressively merge the two closest
clusters until only a single cluster remains. In all cases, we assume that we have a measure of
similarity between pairs of objects, but the different schemes are distinguished by how they convert
this into a measure of similarity between two clusters. For example, in single linkage the similarity
between two clusters is the maximum similarity between points in these two different clusters. In
complete linkage, the similarity between two clusters is the minimum similarity between points in
these two different clusters.
Average linkage defines the similarity between two clusters as the average similarity between points in these two different
clusters~\citep{hierarchysanjoy05,jain1999data}.

Such algorithms have been used in a wide range of application domains ranging from biology to social sciences to computer vision  mainly because they are quite fast
and the output is quite easy to interpret. It is well known, however, that one of the main limitations
of the agglomerative clustering algorithms is that they are not robust to noise~\citep{qcluster2005}.
In this paper we propose and analyze a robust algorithm
for bottom-up agglomerative clustering.
We show that our algorithm satisfies formal robustness guarantees and with proper parameter values, it will be successful in several natural cases where the traditional agglomerative algorithms fail.

In order to formally analyze correctness of our algorithm we use the framework introduced by~\citet{BBV08}.
In this framework, we assume there is some target clustering (much like a $k$-class
target function in the multi-class learning setting) and we say that an algorithm correctly clusters
data satisfying property $P$ if on any data set having property $P$, the algorithm produces a tree such
that the target is some pruning of the tree. For example if all points are more similar to points
in their own target cluster than to points in any other cluster (this is called the strict separation
property), then any of the  standard single linkage, complete linkage, and average linkage  agglomerative algorithms will succeed\footnote{We note however that the Ward's minimum variance method, another classic linkage based procedure, might fail under the strict separation property in the presence of unbalanced clusters. We provide a concrete example in Appendix~\ref{app:ward}.}.
See Figure~\ref{cluster-tree1} for an example. However,
with just tiny bit of noise, for example if each point has even just one point from a different cluster
that it is similar to, then these standard algorithms will all fail (we elaborate on this in Section~\ref{sec-examples}). See Figure~\ref{cluster-tree2} for an example. This brings up the question: is it possible to design an agglomerative algorithm that is
robust to these types of situations and more generally can tolerate a substantial degree of noise?
The contribution of our paper is to provide a  positive answer to this question; we develop a robust,
linkage based algorithm that will succeed in many interesting cases where standard agglomerative
algorithms will fail.
At a high level, our new algorithm is robust to noise in two different and
important ways.  First, it uses more global information for determining the similarities between clusters;
second, it uses a robust linkage procedure involving a median test for linking the clusters, eliminating the influence of the noisy similarities.

\subsection{Our Results}

In particular, in Section~\ref{sec:RMNL} we show that if the data satisfies a natural good neighborhood property, then our
algorithm can be used to cluster well in the tree model, that is, to output a hierarchy such that the
target clustering is (close to) a pruning of that hierarchy.
The good neighborhood property relaxes the strict separation property,
and only requires that after a small number of bad points (which could be extremely malicious) have been removed,
for the remaining good points in the data set, in the neighborhood of their target cluster's size, most of their nearest neighbors are from their target cluster.
We show that our algorithm produces a hierarchy with a pruning that assigns all good points correctly.
In Section~\ref{sec:weak}, we further generalize this to allow for a good fraction of ``boundary''
points that do not fully satisfy the good neighborhood property.
Unlike the good points, these points may have many nearest neighbors outside their target cluster in the neighborhood of their target cluster's size;
but also unlike the bad points, they have additional structure: they fall into a sufficiently large subset of their target cluster,
such that all points in this subset have most of their nearest neighbors from this subset.
As long as the fraction of boundary points in such subsets is not too large,
our algorithm can produce a hierarchy with a pruning that assigns all good and boundary points correctly.

We further show how to adapt our algorithm to the inductive setting with formal correctness guarantees in Section~\ref{sec:inductive}.
In this setting, the clustering algorithm only uses a small random sample over the data set and generates a hierarchy over this sample,
which also implicitly represents a hierarchy over the entire data set.
This is especially useful when the amount of data is enormous such as in astrophysics and biology.
We prove that our algorithm requires only a small random sample whose size is independent of
that of the entire data set and depends only on the noise and the confidence parameters.

We then perform experimental evaluations of our algorithm on synthetic data and real-world data sets.
In controlled experiments on synthetic data presented in Section~\ref{sec:exp:syn}, our algorithm achieves results consistent with our theoretical analysis,
outperforming several other hierarchical algorithms.
We also show in Section~\ref{sec:exp:real} that our algorithm performs consistently better than other hierarchical algorithms in experiments on  several real-world data.
These experimental results suggest that the properties and the algorithm we propose can handle noise in real-world data as well. To obtain good performance, however, our algorithm requires  tuning the noise parameters which roughly speaking quantify the extent to which the good neighborhood property is satisfied.

\subsection{Related Work}
In agglomerative hierarchical clustering~\citep{hierarchysanjoy05,dudahart2000,Jain81dubes.algorithms,jain1999data}, the goal is not to find a single partitioning of the
data, but a hierarchy (generally represented by a tree) of partitionings which may reveal interesting
structure in the data at multiple levels of granularity. Traditionally, only clusterings at a certain
level are considered, but as we argue in Section~\ref{sec:model} it is more desirable to consider all the prunings of
the tree, since this way we can then handle much more general situations.

As mentioned above, it is
well known that standard agglomerative hierarchical clustering techniques are not tolerant to noise~\citep{Nagy:1968,qcluster2005}.
Several algorithms have been proposed to make the hierarchical clustering techniques more robust to noise,
such as Wishart's method~\citep{Wishart69modeanalysis}, and CURE~\citep{Guha_cure:an98}.
Ward's minimum variance method~\citep{Ward1963} is also more preferable in the presence of noise.
However, these algorithms have no theoretical guarantees for their robustness.
Also, our empirical study demonstrates that our algorithm has better tolerance to noise.

On the theoretical side,~\citet{BBV08} analyzed the $\nu$-strict separation property,
a generalization of the simple strict separation property discussed above, requiring
that after a small number of outliers have been removed all points are strictly more similar to points
in their own cluster than to points in other clusters.
They provided an algorithm for producing a hierarchy such that the target clustering is close to some pruning of the tree,
but via a much more computationally expensive (non-agglomerative) algorithm.
Our algorithm is simpler and substantially faster.
As discussed in Section 2.1, the good neighborhood property is much broader than the $\nu$-strict separation
property, so our algorithm is much more generally applicable compared to their algorithm specifically designed for $\nu$-strict separation.

In a different statistical model, \citet{Sanjoy2010} proposed a generalization of Wishart's method.
They proved that given a sample from a density function,
the method constructs a tree that is consistent with the cluster tree of the density.
Although not directly targeting at robustness, the analysis shows the method successfully identifies salient clusters
separated by low density regions, which suggests the method can be robust to the noise represented by the low density regions.

For general clustering beyond hierarchical clustering, there are also works proposing robust algorithms and analyzing robustness of the algorithms; see~\citep{garcia2010review} for a review. In particular, the trimmed $k$-means algorithm~\citep{garcia1999robustness}, a variant of the classical $k$-means algorithm, updates the centers after trimming points that are far away and thus are likely to be noise. \citep{gallegos2002maximum,gallegos2005robust} introduced an interesting mathematical probabilistic framework for clustering in the presence of outliers, and used maximum likelihood approach to estimate the underlying parameters. An algorithm combining the above two approaches is then proposed in~\citep{garcia2008general}. \citep{hennig2008dissolution,ackerman2013clustering} studied the robustness of the classical algorithms such as $k$-means from the perspective of how the clusters are changed after adding some additional points.

\subsection{Structure of the Paper}
The rest of the paper is organized as follows. In Section~\ref{sec:model}, we formalize our model
and  define the good neighborhood property.
We describe our algorithm and prove it succeeds under the good neighborhood property in Section~\ref{sec:RMNL}.
We then prove that it also succeeds under a generalization of the good neighborhood property in Section~\ref{sec:weak}.
In Section~\ref{sec:inductive}, we show how to adapt our algorithm to the inductive setting with formal correctness guarantees.
We provide the experimental results in Section~\ref{sec:exp}, and conclude the paper in Section~\ref{sec:dis}.


\section{Definitions. A Formal Setup} \label{sec:model}
We consider a clustering problem $(\fset,\ell)$ specified as follows. Assume we have a data set $\fset$ of $n$ objects.
Each $\x \in \fset$ has some (unknown) ``ground-truth'' label $\ell(\x)$ in $\lspace=\{1, \ldots , k\}$, where we will think of $k$ as much smaller than $n$.
Let $\cluster_i = \{x \in \fset: \ell(x)=i\}$ denote the set of points of label $i$ (which could be empty), and denote the target clustering as $\clusterings = \{\cluster_1, \ldots, \cluster_k\}$. Let $C(x)$ be a shorthand of $C_{l(x)}$, and $n_{C}$ denote the size of a cluster $C$.

Given another proposed clustering $h$,  $h: \fset \rightarrow \lspace$, we define the error
of $h$ with respect to the target clustering to be
\begin{equation}
\mathbf{err}(h) = \min_{\sigma \in \simgroup_k}{\left[\Pr_{\x \in \fset}{\left[\sigma(h(\x)) \neq \ell(\x)\right]}\right]}
\end{equation}
where $\simgroup_k$ is the set of all permutations on
$\{1,\ldots,k\}$.  Equivalently, the error of a clustering
$\clusterings' = \{\cluster_1', \ldots, \cluster_k'\}$ is
$\min_{\sigma \in \simgroup_k} \frac{1}{n} \sum_i |\cluster_i -
\cluster_{\sigma(i)}'|$. This is popularly known as
\emph{Classification
Error}~\citep{meila_hecker01,BBG09,Voevodski:2012:ACB}.

We will be considering clustering algorithms whose only access to their data is via a pairwise similarity function
$\simm(\x,\xprime)$ that given two examples outputs a number in the range $[-1,1]$. We will say that $\simm$ is a symmetric similarity function if $\simm(\x,\xprime)=\simm(\xprime,\x)$ for all $\x,\xprime$. In this paper we assume that the similarity function $\simm$ is symmetric.

\begin{figure}
\centering
\includegraphics[scale = 0.5]{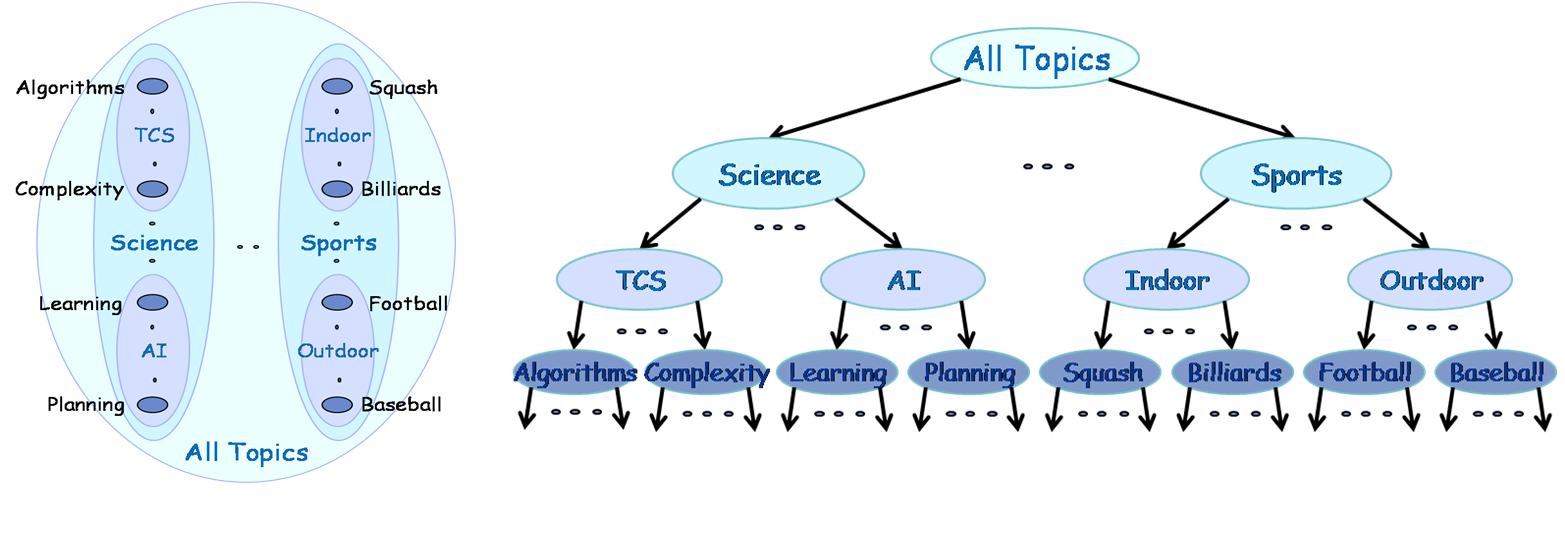}
 \caption{
{\small Consider a document clustering problem. Assume that
data lies in multiple regions $\Algorithms$, $\Complexity$,
$\Learning$, $\Planning$, $\Squash$, $\Billiards$, $\Football$,
$\Baseball$. Suppose that the similarity $\simm(x,y) = 0.999$
if $x$ and $y$ belong to the same inner region; $\simm(x,y) =
3/4$ if $x \in \Algorithms$ and $y \in \Complexity$, or if $x
\in \Learning$ and $y \in \Planning$, or if $x \in \Squash$ and
$y \in \Billiards$, or if $x \in \Football$ and $y \in
\Baseball$; $\simm(x,y) = 1/2$ if $x$ is in ($\Algorithms$ or
$\Complexity$)  and $y$ is in ($\Learning$ or $\Planning$), or
if $x$ is in ($\Squash$ or $\Billiards$) and $y$ is in
($\Football$ or $\Baseball$); define $\simm(x,y) = 0$
otherwise.  Both clusterings
 $\{\Algorithms \cup \Complexity \cup \Learning \cup \Planning, \Squash \cup \Billiards, \Football \cup \Baseball \}$ and
 $\{\Algorithms \cup \Complexity,  \Learning \cup \Planning, \Squash \cup \Billiards \cup \Football \cup \Baseball \}$ satisfy the strict separation
 property.
 }
} \label{cluster-tree1}
\end{figure}

Our goal is to produce a  hierarchical clustering that contains a pruning that is close to the target clustering.
Formally, the goal of the algorithm is to produce a hierarchical clustering: that is, a tree on subsets such
that the root is the set $S$, and the children of any node $S'$ in the tree form a partition of $S'$.  The requirement is that there must exist a {\em pruning} $h$ of the tree (not necessarily using nodes all at the same level) that has error at most $\epsilon$.
\citet{BBV08} have shown that this type of output is necessary in order to be able to analyze
non-trivial properties of the similarity function. For example, even if the similarity function satisfies
the requirement that all points are more similar to all points in their own cluster than to any point in any
other cluster (this is called the strict separation property) and even if we are told the number of clusters, there can  still be multiple  different clusterings that satisfy the property. In particular, one can show examples of similarity functions and two significantly different clusterings of the data satisfying the strict separation property. See Figure~\ref{cluster-tree1} for an example.  However, under the strict separation property, there is a single
hierarchical decomposition such that any consistent clustering is a
pruning of this tree. This motivates clustering in the tree model, which is the model we consider in this work as well.

Given a similarity function satisfying the \strictorder\ property (see Figure~\ref{cluster-tree1} for an example),
we can efficiently construct a tree such that the ground-truth clustering is a pruning of this tree~\citep{BBV08}. Moreover,  the standard linkage  single linkage, average linkage, and complete linkage algorithms would work well under this property. However, one can show that  if the similarity function slightly deviates from the  \strictorder\ condition, then all these standard agglomerative algorithms will fail (we elaborate on this in section~\ref{sec-examples}). In this context, the main question we address in this work is: Can we develop other more robust, linkage based algorithms that will succeed under more realistic and yet natural conditions on the similarity function?

\paragraph{Note}
The strict separation  property does not guarantee that all the cutoffs for different points $x$ are the same, so single linkage would not necessarily have the right clustering if it just stopped once it has $k$ clusters; however the target clustering will provably be a pruning of the final single linkage tree; this is why we define success  based on prunings.

\subsection{Properties of the Similarity Function}\label{subsec:properties}
We describe here some natural properties of the similarity functions that we analyze in this paper.
We start with a noisy version of the  simple strict separation property (mentioned above) which was introduced in~\citep{BBV08} and we then define an interesting and natural generalization of it.

\begin{property}
\label{prop:nu-strict} The similarity function $\simm$ satisfies
{\bf $\nu$-\strictorder} for the clustering problem
$(S,\ell)$ if for some $S' \subseteq S$ of size $(1-\nu)n$, $\simm$
satisfies \strictorder\ for $(S',\ell)$. That is, for all
$x,x',x'' \in S'$ with $x' \in C(x)$ and $x'' \not\in C(x)$ we have
$\simm(x,x') > \simm(x,x'')$.
\end{property}

So, in other words we require that the strict separation is satisfied after a number of bad points have been removed.
A somewhat different condition is to allow each  point to have some bad immediate neighbors  as long as most of its immediate neighbors are good. Formally:

\begin{property}
\label{prop:neigh} The similarity function $\simm$ satisfies
{\bf $\alpha$-\neighbprop\ } property for the clustering problem
$(S,\ell)$ if for all points $x$ we have that all but $\alpha n$ out of their
$n_{C(x)}$ nearest neighbors belong to the cluster $C(x)$.
\end{property}

Note that the $\alpha$-\neighbprop\  property is different from the $\nu$-\strictorder\ property.
For the $\nu$-\strictorder\ property we can have up to $\nu n$ bad points that can misbehave; in particular, these
$\nu n$ bad points  can have similarity $1$ to {\em all} the points in $S$; however, once we remove these
points the remaining points are more similar to  points in their own cluster than to  points in other cluster.
On the other hand, for the $\alpha$-\neighbprop\  property we require that for
 all points $x$ all but $\alpha n$ out of their
$n_{C(x)}$ nearest neighbors belong to the cluster $C(x)$. (So we cannot have a point that has similarity $1$
to all the points in $S$.)  Note however that different points might  misbehave on different $\alpha n$ neighbors.
We can also consider a property that generalizes both the $\nu$-\strictorder\ and the $\alpha$-\neighbprop\ property.
Specifically:

\begin{property}
\label{prop:nu-neigh} The similarity function $\simm$ satisfies
{\bf $(\alpha, \nu)$-\neighbprop\ }property for the clustering
problem $(S,\ell)$ if  for some $S' \subseteq S$ of size
$(1-\nu)n$, $\simm$ satisfies $\alpha$-\neighbprop\ for
$(S',\ell)$. That is, for all points $x \in S'$ we have that
all but $\alpha n$ out of their $n_{C(x) \cap S'}$ nearest
neighbors in $S'$ belong to the cluster $C(x)$.
\end{property}

Clearly, the $(\alpha, \nu)$-good neighborhood property is a generalization of both the $\nu$-strict separation and $\alpha$-good neighborhood property. Formally,

\begin{fact}
If  the similarity function $\simm$ satisfies the  $\alpha$-\neighbprop\ property for the clustering problem
$(S,\ell)$, then  $\simm$ also satisfies the  $(\alpha, 0)$-\neighbprop\ property for the clustering problem $(S,\ell)$.
\end{fact}

\begin{fact}
If  the similarity function $\simm$ satisfies the $\nu$-\strictorder\ property for the clustering problem
$(S,\ell)$, then  $\simm$ also satisfies the  $(0, \nu)$-\neighbprop\ property for the clustering problem $(S,\ell)$.
\end{fact}

\citet{BBV08} have shown that if $\simm$ satisfies the
$\nu$-strict separation property with respect to the target
clustering, then as long as the smallest target cluster has
size $5\nu n$, one can  in polynomial time construct a
hierarchy  such that the ground-truth is $\nu$-close to a
pruning  of the hierarchy.
Unfortunately the algorithm presented in~\citep{BBV08} is computationally very expensive: it first generates a large list of $\Omega(n^2)$ candidate clusters and repeatedly runs pairwise tests in order to laminarize these clusters; its running time is a large unspecified polynomial.  The new robust  linkage algorithm we present in Section~\ref{sec:RMNL} can be used to get a simpler and  much faster algorithm for clustering accurately under the $\nu$-strict separation and the more general $(\alpha, \nu)$-good neighborhood property.

\paragraph{Generalizations} Our algorithm succeeds under an even more general property called weak good neighborhood,
which allows a good fraction of points to only have nice structure in their small local neighborhoods.
The relations between these properties are described in Section~\ref{sec:relate}, and the analysis under the weak good neighborhood is presented in Section~\ref{sec:analysisWeak}.

\subsection{Standard Linkage Based Algorithms Are Not Robust}
\label{sec-examples}

\begin{figure}
\centering
\includegraphics[scale = 0.5]{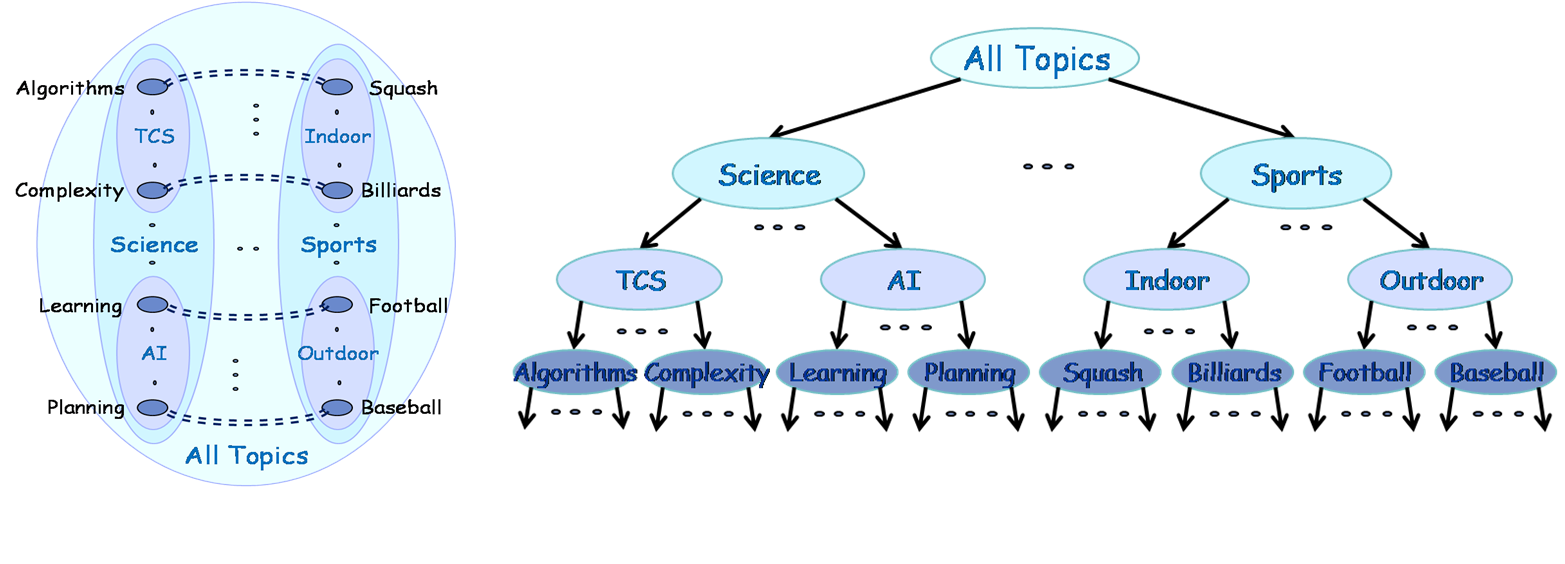}
 \caption{
 {\small
\label{fig2} Same as Figure \ref{cluster-tree1} except that let
us match each point in $\Algorithms$ with a point in $\Squash$,
each point in $\Complexity$ with a point in $\Billiards$, each
point in $\Learning$ with a point in $\Football$, and each
point in $\Planning$ with a point in region $\Baseball$. Define
the similarities to be the same as in
Figure~\ref{cluster-tree1} except that we let $\simm(x,y) = 1$
if $x$ and $y$ are matched.  Note that for $\alpha = 1/n$ the
similarity function satisfies the $\alpha$-\neighbprop\ with
respect to any of the prunings of the tree above.  However,
single linkage, average linkage, and complete linkage would
initially link the matched pairs and produce clusters with very
high error with respect to any such clustering.
}
}
\label{cluster-tree2}
\end{figure}

As we show below, even if the data satisfies the \neighbprop\
property, the standard  single linkage, average linkage, and complete
linkage algorithms might fail. The contribution of our work is to develop a robust,
linkage based algorithm that will succeed under these natural
conditions. More specifically, we can show an example where the standard single linkage, average linkage, and complete
linkage algorithms would perform very badly, but
where our algorithm would work well. In particular, let us
slightly modify the example in Figure~\ref{cluster-tree1}, by
adding a little bit of noise, to form links of high similarity
between points in different inner blobs\footnote{Since, usually, the similarity function between
pairs of objects is constructed based on heuristics, this could
happen in practice; for example we could have a similarity
measure that puts a lot of weight on features such as date or
names, and so we could easily have a document about Learning
being more similar to a document about Football than to other
documents about Learning. While this example seems a little bit
contrived, in Figure~\ref{fig:weakProperty} in
Section~\ref{sec:weak} we will give a naturally occurring
example where  the standard  single linkage, average linkage, and complete
linkage algorithms still fail but our
algorithm succeeds because it satisfies a generalization of the
good neighborhood property that we will discuss in
Section~\ref{sec:weak}.}. See Figure~\ref{cluster-tree2} for a
precise description of the similarity.
In this example all the  single linkage, average linkage, and complete
linkage algorithms would in the first $n/2$ stages merge the matched pairs of points.
From that moment on, no matter how they perform,
none of the natural and desired clusterings will even be $1/2$ close to any of the prunings of the hierarchy produced.
Notice however, that $\simm$ satisfies the $\alpha$-\neighbprop\ with respect to any of the desired clusterings (for $\alpha=1/n$), and that our algorithm will be successful on this instance. The $\nu$-\strictorder\ is not satisfied in this example either, for any constant $\nu$.



\section{Robust Median Neighborhood Linkage}\label{sec:RMNL}
In this section, we propose a new algorithm, Robust Median Neighborhood Linkage,
and show that it succeeds for instances satisfying the $(\alpha, \nu)$-good neighborhood property.

Informally, the algorithm maintains a threshold $t$ and a list $\mathcal{C}'_{t}$ of subsets of points of $S$; these subsets are called blobs for convenience.
We first initialize the threshold to a value $t$ that is not too large and not too small ($t=6(\alpha+\nu)n + 1$),
and initialize $\mathcal{C}'_{t-1}$ to contain $|S|$ \sets, one for each point in $S$.
For each $t$, the algorithm builds $\mathcal{C}'_{t}$ from $\mathcal{C}'_{t-1}$ by merging two or more \sets\ as follows.
It first builds a graph $F_t$, whose 

\newcommand{\bw}{\quad}
\newcommand{\sbw}{\qquad\ \ \ \ \ \quad}
\newcommand{\tw}{\qquad}
\begin{algorithm}[H]
\caption{\label{alg:RMNL} Robust Median Neighborhood Linkage}
\begin{algorithmic}
\STATE{\textbf{Input:} Similarity function $\simm$ on a set of points $S$, $n = |S|$, noise parameters $\alpha > 0, \nu >0$.}
\STATE \textbf{Step 1} \bw Initialize $t = \tinit$. \\
       \sbw Initialize $\mathcal{C}'_{t-1}$ to be a list of blobs so that each point is in its own \set.
\STATE{\sbw \textbf{while} $|\mathcal{C}'_{t-1}|>1$ \textbf{do} }
\STATE{\textbf{Step 2} \bw \tw $\mybullet$ {\tt Build a graph $F_t$ whose vertices are points in $S$ and \\
\sbw \tw \tw  whose edges are specified as follows.}}
\STATE{\sbw \tw Let $N_t(x)$ denote the $t$ nearest neighbors of $x$.}
\STATE{\sbw \tw \textbf{for} any $x,y \in S$ that satisfy $|N_t(x) \cap N_t(y)| \geq t - \tFt$ \textbf{do}}
\STATE{\sbw \tw \tw  Connect $x,y$ in $F_t$.}
\STATE{\sbw \tw \textbf{end for}}
\STATE{\textbf{Step 3} \bw \tw $\mybullet$ {\tt Build a graph $H_t$ whose vertices are \sets\ in $\mathcal{C}'_{t-1}$ and \\
\sbw \tw \tw whose edges are specified as follows. }}
\STATE{\sbw \tw Let $N_F(x)$ denote the neighbors of $x$ in $F_t$.}
\STATE{\sbw \tw \textbf{for} any $C_u, C_v \in \mathcal{C}'_{t-1}$ \textbf{do}}
\STATE{\sbw \tw \tw \textbf{if} $C_u, C_v$ are singleton \sets, i.e.\ $C_u =\{x\}, C_v =\{y\}$ \textbf{then}}
\STATE{\sbw \tw \tw \tw Connect $C_u, C_v$ in $H_t$, if $|N_F(x)\cap N_F(y)| > \tHt$.}
\STATE{\sbw \tw \tw \textbf{else}}
\STATE{\sbw \tw \tw \tw Set $S_t(x,y) = |N_F(x) \cap N_F(y) \cap (C_u \cup C_v)|$, i.e.\ the number of}
\STATE{\sbw \tw \tw \tw \tw points in $C_u \cup C_v$ that are common neighbors of $x,y$ in $F_t$.}
\STATE{\sbw \tw \tw \tw Connect $C_u, C_v$ in $H_t$, if $\median_{x \in C_u, y \in C_v} S_t(x, y) > \frac{|C_u| + |C_v|}{4}$.}
\STATE{\sbw \tw \tw \textbf{end if}}
\STATE{\sbw \tw \textbf{end for}}
\STATE{\textbf{Step 4} \bw \tw $\mybullet$ {\tt Merge \sets\ in $\mathcal{C}'_{t-1}$ to get $\mathcal{C}'_{t}$}}
\STATE{\sbw \tw Set $\mathcal{C}'_{t} = \mathcal{C}'_{t-1}$.}
\STATE{\sbw \tw \textbf{for} any connected component $V$ in $H_t$ with $|\bigcup_{C\in V}C|\geq\tmerge$ \textbf{do}}
\STATE{\sbw \tw \tw Update $\mathcal{C}'_t$ by merging all \sets\ in $V$ into one \set.}
\STATE{\sbw \tw \textbf{end for}}
\STATE{\textbf{Step 5} \bw \tw $\mybullet$ {\tt Increase threshold}}
\STATE{\sbw \tw $t=t+1$.}
\STATE{\sbw \textbf{end while}}
\STATE
\STATE{\textbf{Output:} Tree $T$ with single points as leaves and internal nodes corresponding to the merges performed.}
\end{algorithmic}
\end{algorithm}

\noindent
vertices are the data points in $S$
and whose edges are constructed
by connecting any two points that share at least $t-2(\alpha+\nu)n$ points in common
out of their $t$ nearest neighbors. Then it builds a graph $H_t$ whose vertices correspond to blobs in $\mathcal{C}'_{t-1}$ and whose edges are specified
in the following way.
Two singleton \sets\ $C_u=\{x\}$ and $C_v=\{y\}$ are connected in $H_t$ if the points $x,y$ have more than $(\alpha+\nu) n$ common neighbors in $F_t$.
For \sets\ $C_u$ and $C_v$ that are not both singleton, the algorithm performs a median test.
In this test, for each pair of points $x\in C_u, y\in C_v$, it computes the number $S_t(x,y)$ of points $z\in C_u \cup C_v$ that are the common neighbors of $x$ and $y$ in $F_t$.
It then connects $C_u$ and $C_v$ in $H_t$ if $\median_{x\in C_u, y\in C_v}S_t(x,y)$ is larger than $1/4$ fraction of $|C_u| + |C_v|$.
Once $H_t$ is built, we analyze its connected components in order to create $\mathcal{C}'_{t}$.
For each connected component $V$ of $H_t$, if $V$ contains sufficiently many points from $S$ in its \sets\,
we merge all its \sets\ into one \set\ in $\mathcal{C}'_{t}$.
After building $\mathcal{C}'_{t}$, the threshold is increased and the above steps are repeated until
only one \set\ is left.
The algorithm finally outputs the tree with single points as leaves and internal nodes corresponding to the merges performed.
The full details of our algorithm are described in Algorithm~\ref{alg:RMNL}.
Our main result in this section is the following:

\begin{theorem}\label{thm:RMNL}
Let $\simm$ be a symmetric similarity function satisfying the $(\alpha, \nu)$-good neighborhood for the clustering problem $(S,\ell)$.
As long as the smallest target cluster has size greater than $6( \nu + \alpha) n$,
Algorithm~\ref{alg:RMNL} outputs a hierarchy
such that a pruning of the hierarchy is $\nu$-close to the target clustering in time $O(n^{\omega + 1})$,
where $O(n^{\omega})$ is the state of the art for matrix multiplication.
\end{theorem}

In the rest of this section,
we will first describe the intuition behind the algorithm in Section~\ref{intuition}
and then prove Theorem~\ref{thm:RMNL} in Section~\ref{sec:rmnlproof}.

\subsection{Intuition of the Algorithm under the Good Neighborhood Property}
\label{intuition}
We begin with some convenient terminology and a simple fact about the good neighborhood property.
In the definition of the $(\alpha,\nu)$-good neighborhood property (see Property~\ref{prop:nu-neigh}), we call the points
in $S'$ good points and the points in $B = S \setminus S'$ bad points. Let $G_i = C_i \cap S'$ be the good set of label
$i$. Let $G = \cup_i G_i$ denote the whole set of good points; so $G = S'$. Clearly $|G| \geq n - \nu n$.
Recall that $n_{C_i}$ is the number of points in the cluster $C_i$.
Note that the following is a useful consequence of the $(\alpha,\nu)$-good neighborhood property.

\begin{fact}\label{fact:nn}
Suppose the similarity function $\simm$ satisfies the $(\alpha,\nu)$-good neighborhood property
for the clustering problem $(S,\ell)$. As long as $t$ is smaller than $n_{C_i}$, for any good point $x \in C_i$, all but
at most $(\alpha+\nu)n$ out of its $t$ nearest neighbors lie in its good set $G_i$.
\end{fact}
\begin{proof}
Let $x \in G_i$.
By definition, out of its $|G_i|$ nearest neighbors in $G$, there are at least $|G_i| - \alpha n$
points from $G_i$. These points must be among its $|G_i| + \nu n$ nearest neighbors in $S$, since there are at most $\nu n$ bad
points in $S \setminus G$. This means that at most $(\alpha + \nu) n$ out of its $|G_i| + \nu n$ nearest neighbors are
outside $G_i$. Notice $|G_i| + \nu n \geq n_{C_i}$, we have that at most $(\alpha+\nu)n$ out of its $n_{C_i}$ nearest neighbors
are outside $G_i$, as desired.
\end{proof}

\begin{figure}[!t]
\begin{center}
\centering
    \subfloat[$F$]{\includegraphics[width=0.5\textwidth]{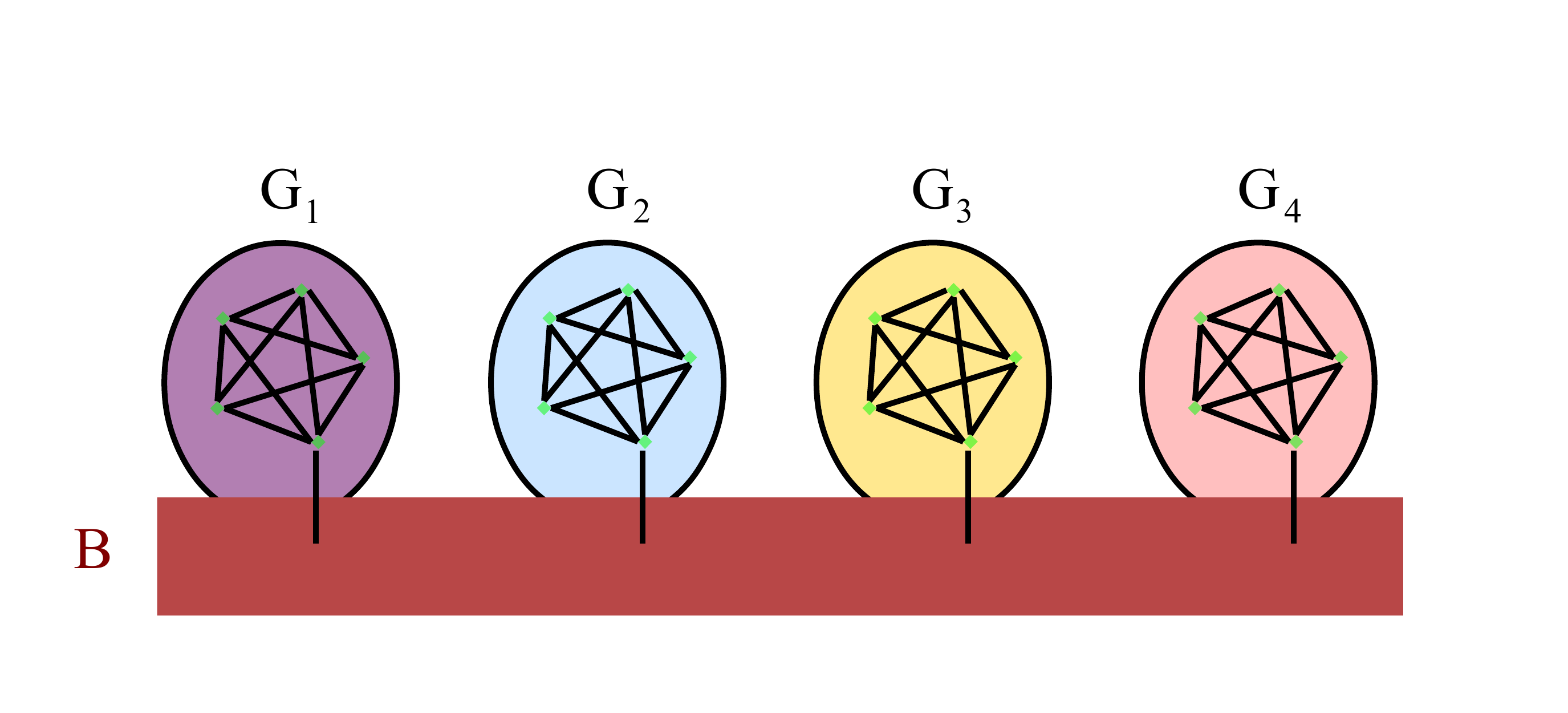}}
    \hspace*{.2in}
    \subfloat[$H$]{\includegraphics[width=0.5\textwidth]{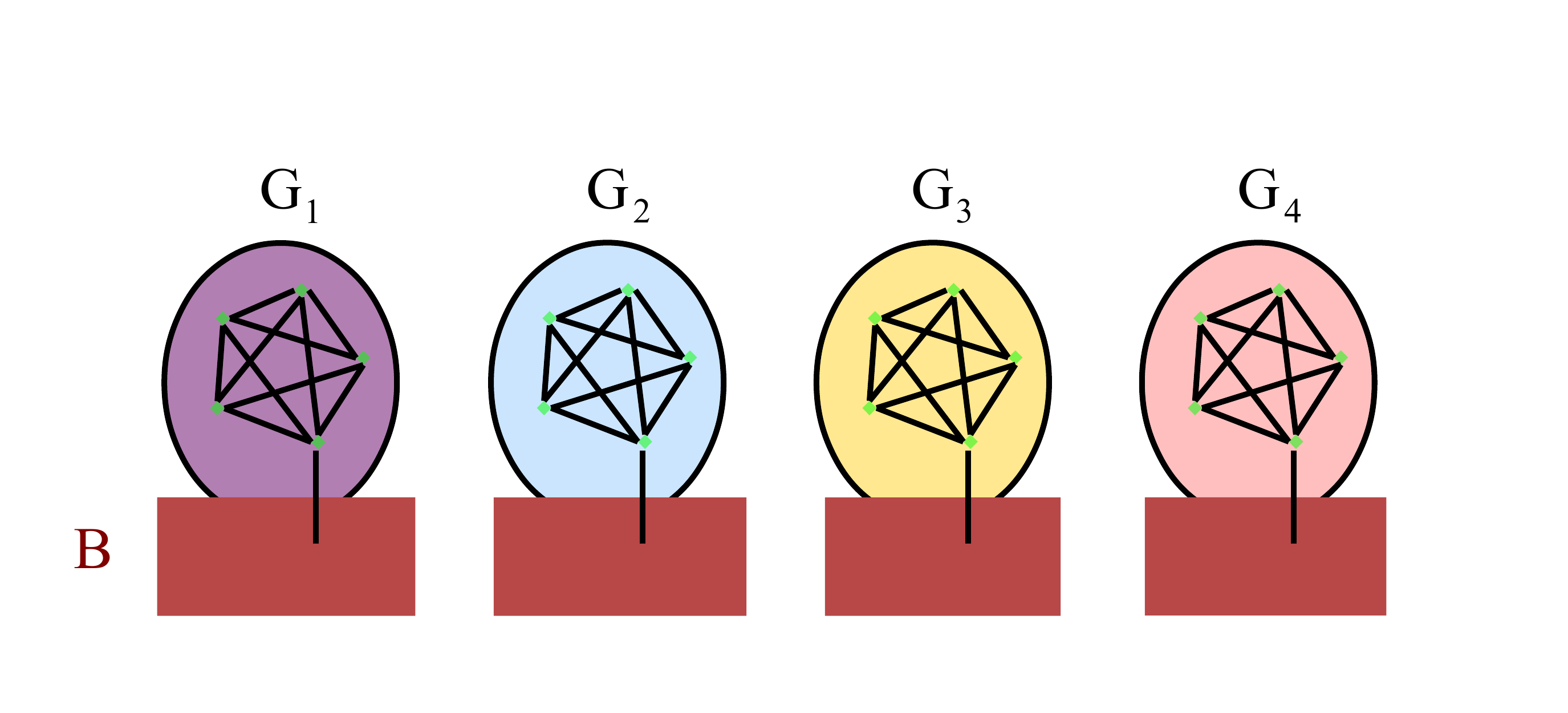}}
\end{center}
\vspace{-.2in}
\caption{
{\small
Graph $F$ and $H$ when all target clusters are of the same size $n_C$, which is known. In $F$, no two good points in different target clusters can be connected,
and all good points in the same target cluster will be connected. In $H$, bad points connected to good points from different target clusters are disconnected.
}
}\label{fig:FH}
\vspace{-.2in}
\end{figure}

\noindent
{\bf Intuition} We first assume for simplicity that all the target clusters have the same size $n_C$ and that we know $n_C$. In this case it is quite easy to recover the target clusters as follows.
We first construct a graph $F$ whose vertices are points in $S$; we connect two points in $F$ if they share at least $n_C - 2(\nu + \alpha )n$ points in common among their $n_C$ nearest neighbors.
By Fact~\ref{fact:nn}, if the target clusters are not too small (namely $n_C > 6(\nu + \alpha)n$), we are guaranteed that
no two good points in different target clusters will be connected in $F$,
and that all good points in the same target cluster will be connected in $F$.
If there are no bad points ($\nu=0$), then each connected component of $F$ corresponds to a target cluster, and we could simply output them.
Alternatively, if there are bad points ($\nu >0$), we can still cluster well as follows. We construct a new graph $H$ on points in $S$ by connecting points that share more than $(\alpha+\nu) n$ neighbors in the graph $F$.
The key point is that in $F$ a bad point can be connected to good points from only one single target cluster.
This then ensures that no good points from different target clusters are in the same connected component in $H$.
So, if we output the largest $k$ components of $H$, we will obtain a clustering with error at most $\nu n$.
See Figure~\ref{fig:FH} for an illustration.

If we do not know $n_C$, we can still use a pretty simple procedure. Specifically, we start with a threshold $t\leq n_C$ that is not too small and not too large (say $6(\nu + \alpha)n < t \leq n_C $), and build a graph $F_t$ on $S$ by connecting two points if they share at least $t - 2(\nu + \alpha )n$ points in common out of their $t$ nearest neighbors.
We then build another graph $H_t$ on $S$ by connecting points if they share more than $(\alpha+\nu) n$ neighbors in the graph $F_t$.
The key idea is that when $t\leq n_C$,
good points from different target clusters share less than $t - 2(\nu + \alpha )n$ neighbors, and thus are not connected in $F_t$ and $H_t$.
If the $k$ largest connected components of $H_t$ all have sizes greater than $(\alpha+\nu)n$ and they cover at least
a $(1-\nu)$ fraction of the whole set of points $S$, then these components must correspond to the target clusters and we can output them.
Otherwise, we increase the critical threshold and repeat.
By the time we reach $n_C$,
all good points in the same target clusters will get connected in $F_t$ and $H_t$,
so we can identify the $k$ largest components as the target clusters.

\begin{figure}[!t]
\begin{center}
\centering
    \subfloat[$F_t$]{\includegraphics[width=0.45\textwidth]{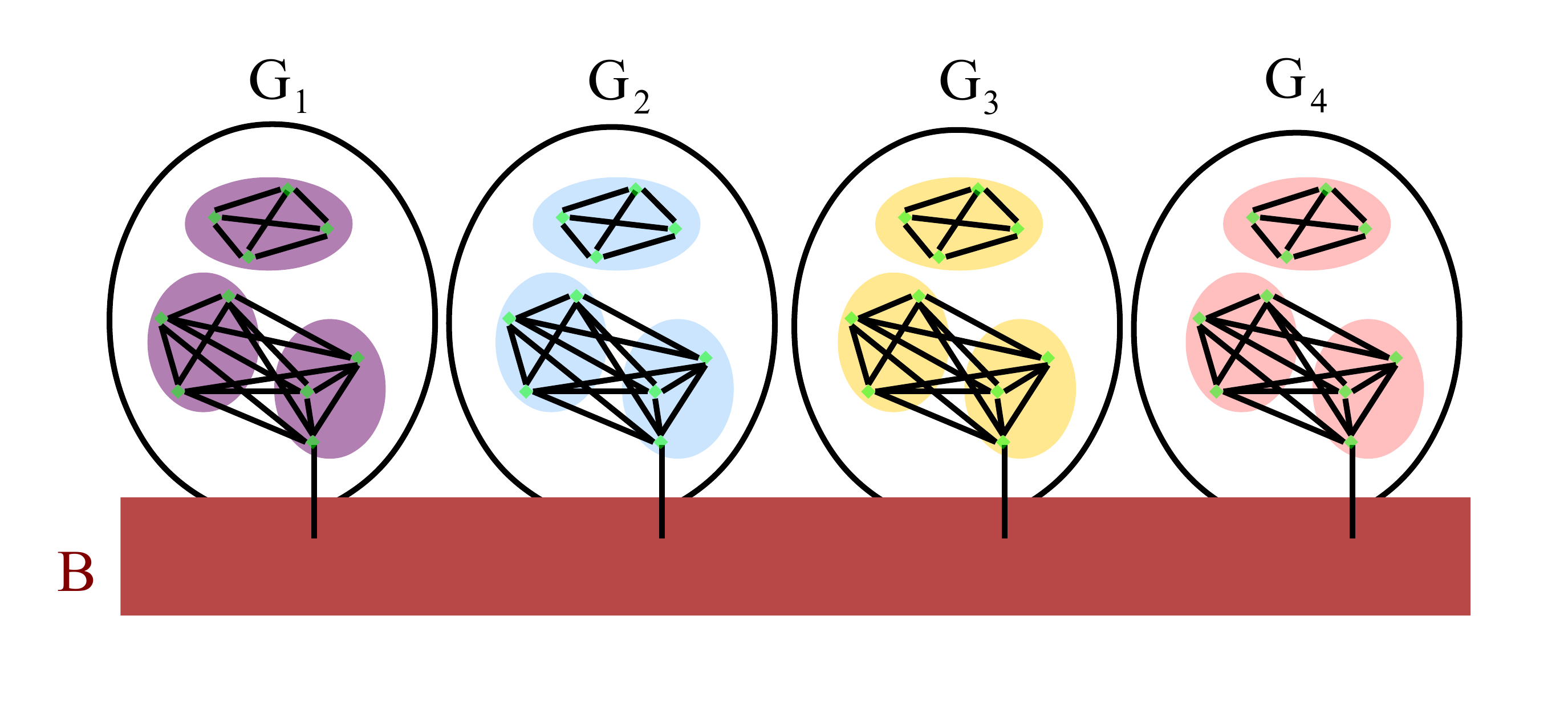}}
    \hspace*{.1in}
    \subfloat[$H_t$]{\includegraphics[width=0.45\textwidth]{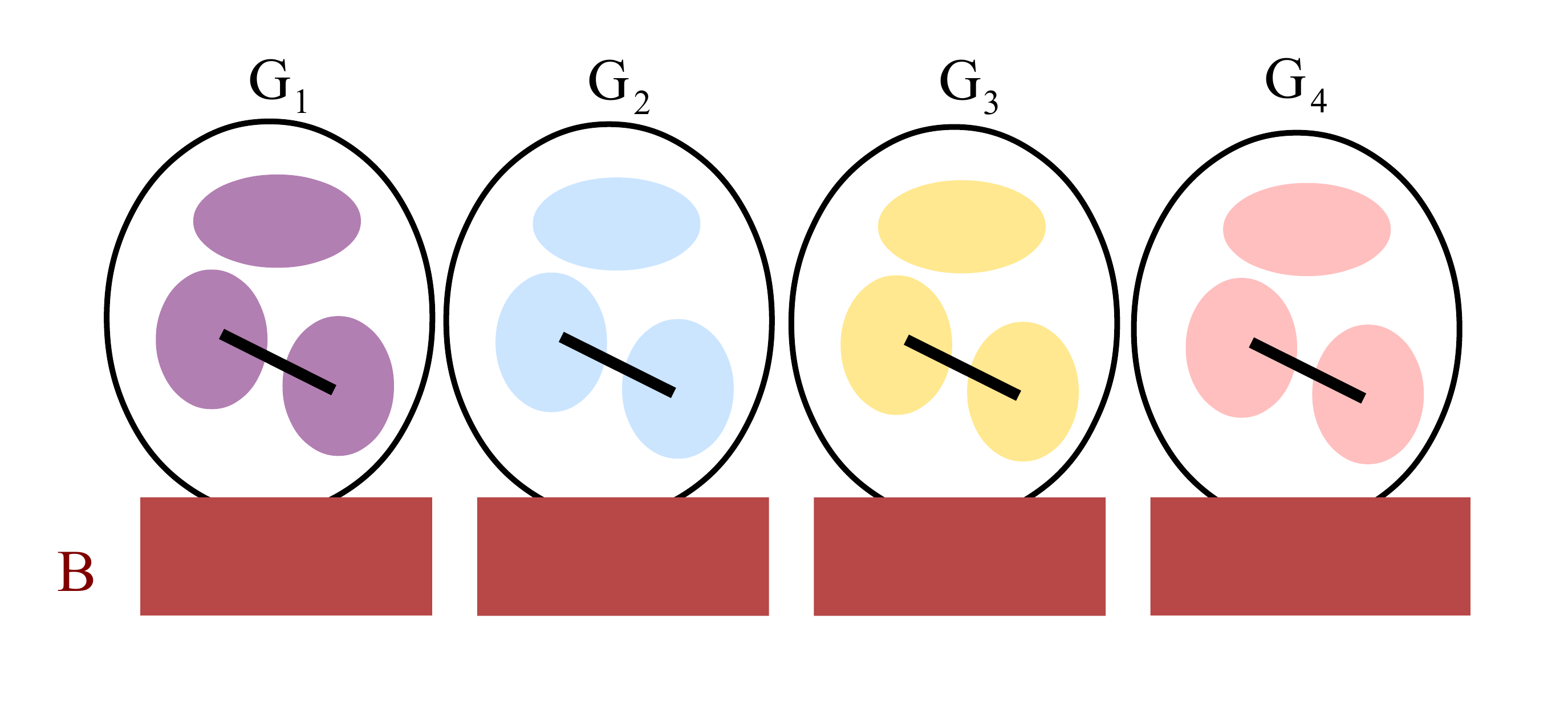}}
\end{center}
\vspace{-.2in}
\caption{
{\small Graph $F_t$ and $H_t$ when target clusters have the same size $n_C$ but we do not know $n_C$. The figure shows the case when $t<n_{C}$.
In $F_t$, no good points are connected with good points outside their target clusters;
in $H_t$, \sets\ containing good points in different target clusters are disconnected.
}
}\label{fig:FH_te}
\vspace{-.2in}
\end{figure}

Note that as mentioned above, when $t \leq n_C$, each connected component in $H_t$ can contain good points from only one target cluster.
An alternative procedure is to reuse this information in later thresholds,
so that we do not need to build the graph $H_t$ from scratch as described in the above paragraph.
Specifically, we maintain a list $\mathcal{C}'_{t}$ of subsets of points;
these subsets are called \sets\ for convenience.
We start with a list where each blob contains a single point.
At each threshold $t$, we build $F_t$ on the points in $S$ as before,
but build $H_t$ on the blobs in $\mathcal{C}'_{t-1}$ (instead of on the points in $S$).
When building $H_t$, for two singleton \sets, it is safe to connect them if their points share enough neighbors in $F_t$.
For non-singleton \sets, it turns out that we can use a median test to outvote the noise\footnote{The median test is quite robust and as we show, it allows some points in these \sets\ to have weaker properties than the good neighborhood. See Section~\ref{sec:weak} for examples of such points and a theoretical analysis of the robustness.}.
In particular, for two \sets\ $C_u$ and $C_v$ that are not both singleton,
we compute for all $x \in C_u$ and $y\in C_v$ the quantity $S_t(x,y)$,
which is the number of points in $C_u \cup C_v$ that are the common neighbors of $x$ and $y$ in $F_t$.
We then connect the two \sets\ in $H_t$ if $\median_{x\in C_u, y\in C_v} S_t(x,y)$ is sufficiently large.
See Figure~\ref{fig:FH_te} for an illustration and see Step 3 in Algorithm~\ref{alg:RMNL} for the details.

\begin{figure}[!t]
\begin{center}
\centering
    \subfloat[$F_t$]{\includegraphics[width=0.45\textwidth]{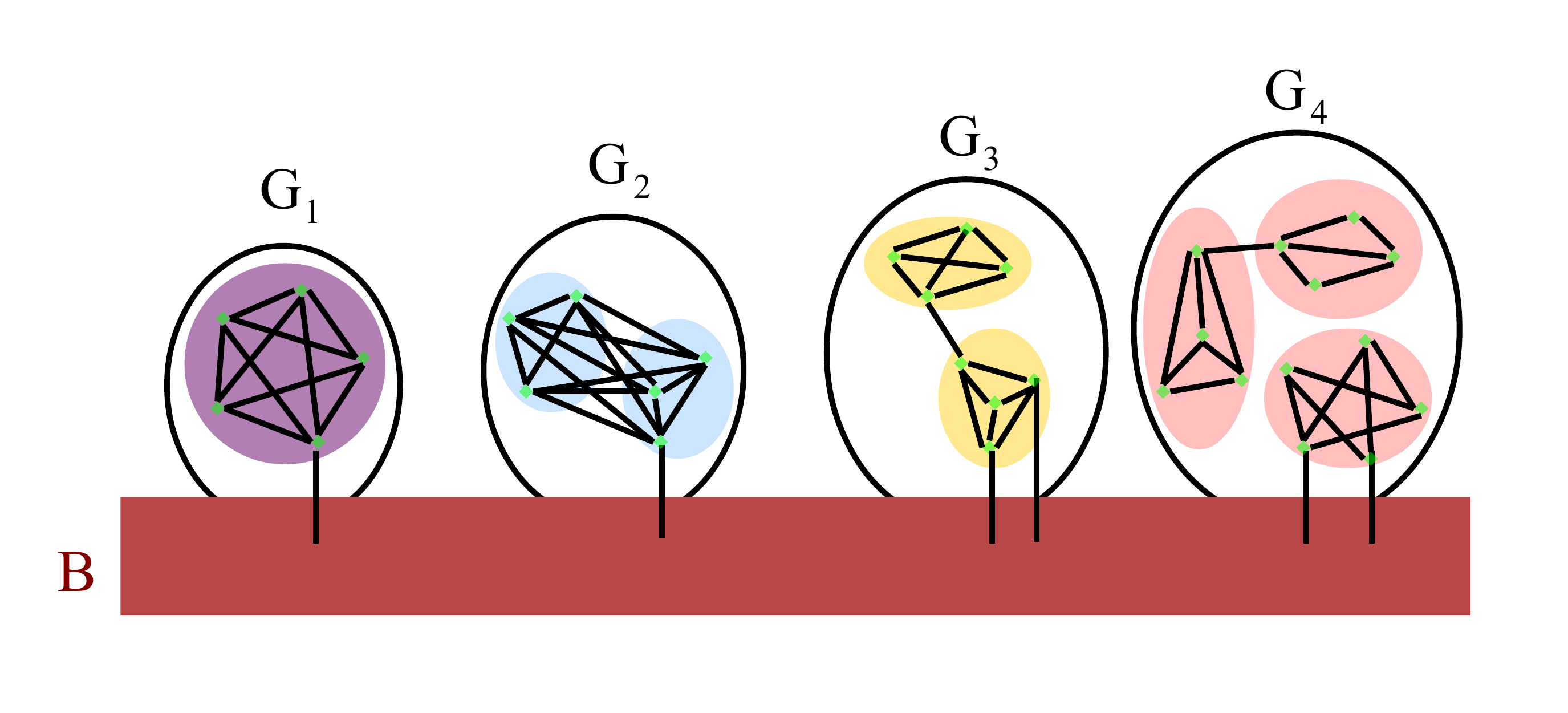}}
    \hspace*{.1in}
    \subfloat[$H_t$]{\includegraphics[width=0.45\textwidth]{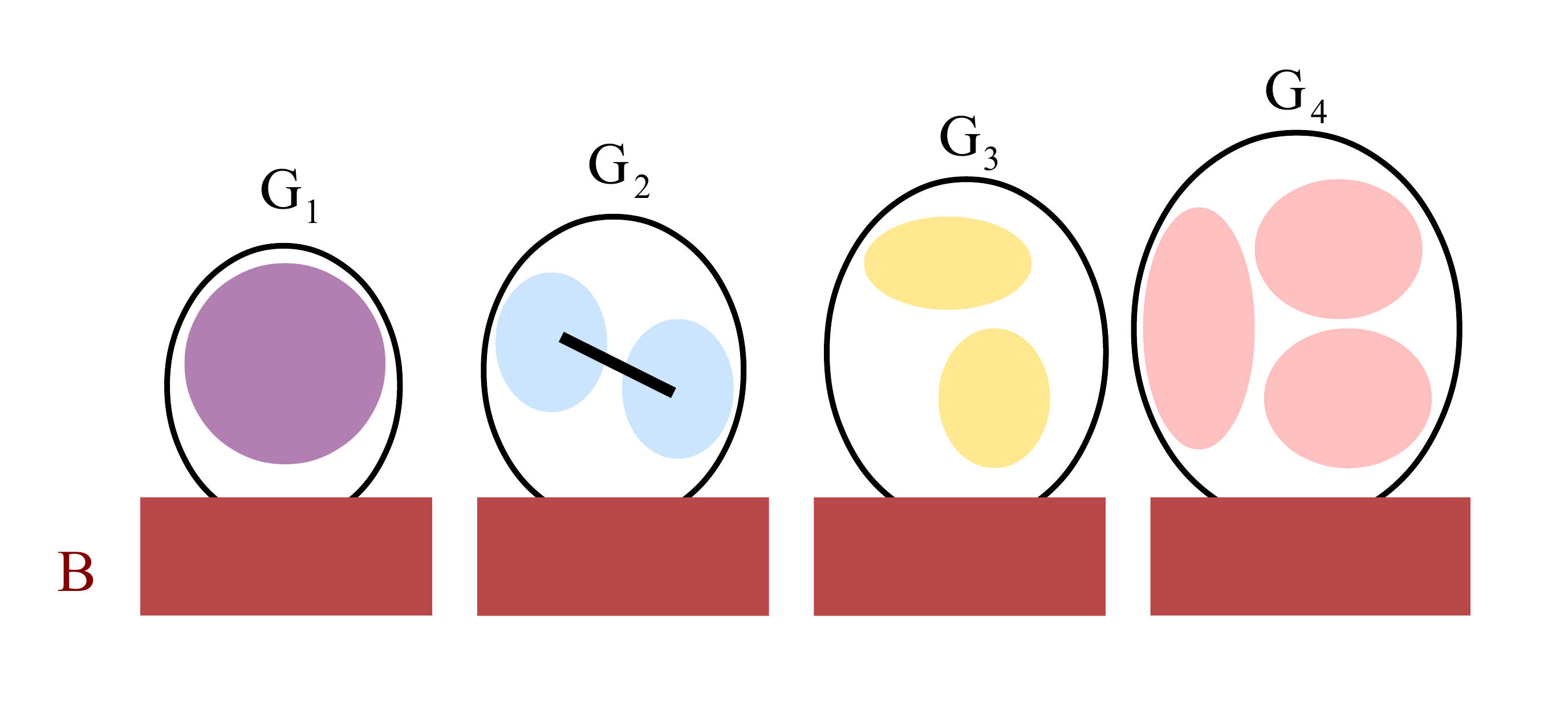}}
\end{center}
\vspace{-.2in}
\caption{
{\small Graph $F_t$ and $H_t$ when target clusters are of different sizes. The figure shows the case when $t=n_{C_2}$.
In $F_t$, no good points are connected with good points outside their target clusters;
good points in $C_2$ form a clique since $t=n_{C_2}$.
In $H_t$, \sets\ containing good points in different target clusters are disconnected;
\sets\ containing good points in $C_2$ are all connected.
}
}\label{fig:FH_t}
\end{figure}

In the general case where the sizes of the target clusters are different,
similar ideas can be applied.
The key point is that when $t \leq n_{C_i}$,
good points from $C_i$ share less than $t - 2(\nu + \alpha )n$ neighbors with good points outside, and thus are not connected to them in $F_t$.
Then in $H_t$, we can make sure that no \sets\ containing good points in $C_i$ will be connected with \sets\ containing good points outside $C_i$.
When $t=n_{C_i}$, good points in $C_i$ form a clique in $F_t$, then all the \sets\ containing good points in $C_i$ are connected in $H_t$, and thus are merged.
See Figure~\ref{fig:FH_t} for an illustration.
Full details are presented in Algorithm~\ref{alg:RMNL} and the proof of Theorem~\ref{thm:RMNL} in the following subsection.

\subsection{Correctness under the Good Neighborhood Property}\label{sec:rmnlproof}

In this subsection, we prove Theorem~\ref{thm:RMNL} for our algorithm.
The correctness follows from Lemma~\ref{lem:blob} and the running time follows from Lemma~\ref{lem:sec:rmnl:runtime}.
Before proving these lemmas, we begin with a useful fact which follows immediately from the design of the algorithm.

\begin{fact}\label{fact:bad}
In Algorithm~\ref{alg:RMNL}, for any $t$, if a \set\ in $\mathcal{C}_t$ contains at least one good point,
then at least $3/4$ fraction of the points in that \set\ are good points.
\end{fact}
\begin{proof}
This is clearly true when the \set\ is singleton.
When it is non-singleton, it must be formed in Step 4 in Algorithm~\ref{alg:RMNL},
so it contains at least $\tmerge$ points.
Then the claim follows since there are at most $\nu n$ bad points.
\end{proof}

\begin{lemma} \label{lem:blob}
The following claims are true in Algorithm~\ref{alg:RMNL}:\\
(1) For any $C_i$ such that $t \leq |C_i|$, any \set\ in $\mathcal{C}'_t$ containing good points from $C_i$ will not contain good points outside $C_i$.\\
(2) For any $C_i$ such that $t = |C_i|$, all good points in $C_i$ belong to one \set\ in $\mathcal{C}'_t$.
\end{lemma}

\begin{proof}
Before proving the claims, we first show that the graph $F_t$ constructed in Step 2 has the following useful properties.
Recall that $F_t$ is constructed on points in $S$ by connecting any two points that share at least $t-2(\alpha+\nu)n$ points in common
out of their $t$ nearest neighbors.
For any $C_i$ such that $t \leq |C_i|$, we have:

\begin{itemize}
\item[(a)] If $x$ is a good point in $C_i$ and $y$ is a good point outside $C_i$, then $x$ and $y$ are not connected in $F_t$. \medskip \\
\hspace*{4mm} To see this, first note that by Fact~\ref{fact:nn},
$x$ has at most $(\alpha + \nu) n$ neighbors outside $C_i$ out of the $t$ nearest neighbors.
For $y \in G_j$, if $n_{C_j} \geq t$, then $y$ has at most $(\alpha + \nu) n$ neighbors in $C_i$;
if $n_{C_j} < t$, $y$ has at most $(\alpha + \nu) n + t - n_{C_j}$ neighbors in $C_i$.
In both cases, $y$ has at most $(\alpha + \nu) n + \max(0, t- n_{C_j}) < t - 5(\alpha+\nu)n$ neighbors in $C_i$, since $n_{C_j} > 6(\alpha+\nu)n$
and $t > 6(\alpha+\nu)n$.
Then $x$ and $y$ have at most $t - 4(\alpha+\nu)n$ common neighbors, so they are not connected in $F_t$.
\item[(b)] If $x$ is a good point in $C_i$, $y$  is a good point outside $C_i$, and $z$ is a bad point, then $z$ cannot be connected to both $x$ and $y$ in $F_t$.\medskip \\ \hspace*{4mm}
    To prove this, we will show that if $z$ is connected to $x$, then $z$ cannot be connected to $y$.
    First, by the same argument as above, out of the $t$ nearest neighbors, $y$ has less than $t - 5(\alpha + \nu) n$ neighbors in $C_i$.
    Second, by Fact~\ref{fact:nn}, $x$ has at most $(\alpha + \nu) n$ neighbors outside $C_i$.
    If $z$ has less than $t - 3(\alpha + \nu) n$ neighbors in $C_i$, then $z$ and $x$ share less than $t - 3(\alpha + \nu) n + (\alpha + \nu) n = t -2(\alpha+\nu) n$ neighbors and will not be connected.
    So $z$ must have at least $t - 3(\alpha + \nu) n$ neighbors in $C_i$, and thus cannot have more than $3(\alpha + \nu) n$ neighbors outside $C_i$. The two statements show that $y$ and $z$ share less than $t - 5(\alpha + \nu) n$ neighbors in $C_i$,
    and at most $3(\alpha + \nu) n$ neighbors outside $C_i$.
    So they share less than $t - 2(\alpha + \nu)n + 3(\alpha+\nu)n = t - 2(\alpha + \nu)n$ neighbors and thus are not connected in $F_t$.
\end{itemize}

Now we prove Claim (1) in the lemma by induction on $t$.
The claim is clearly true initially.
Assume for induction that the claim is true for the threshold $t-1 < |C_i|$, that is,
for any $C_i$ such that $t-1 < |C_i|$, any \set\ in $\mathcal{C}'_{t-1}$ containing good points from $C_i$ will not contain good points outside $C_i$.
We now prove that the graph $H_t$ constructed in Step 3 has the following properties, which can be used to show that the claim is still true for the
threshold $t$.

\begin{figure}[!t]
\begin{center}
\includegraphics[width=0.4\textwidth]{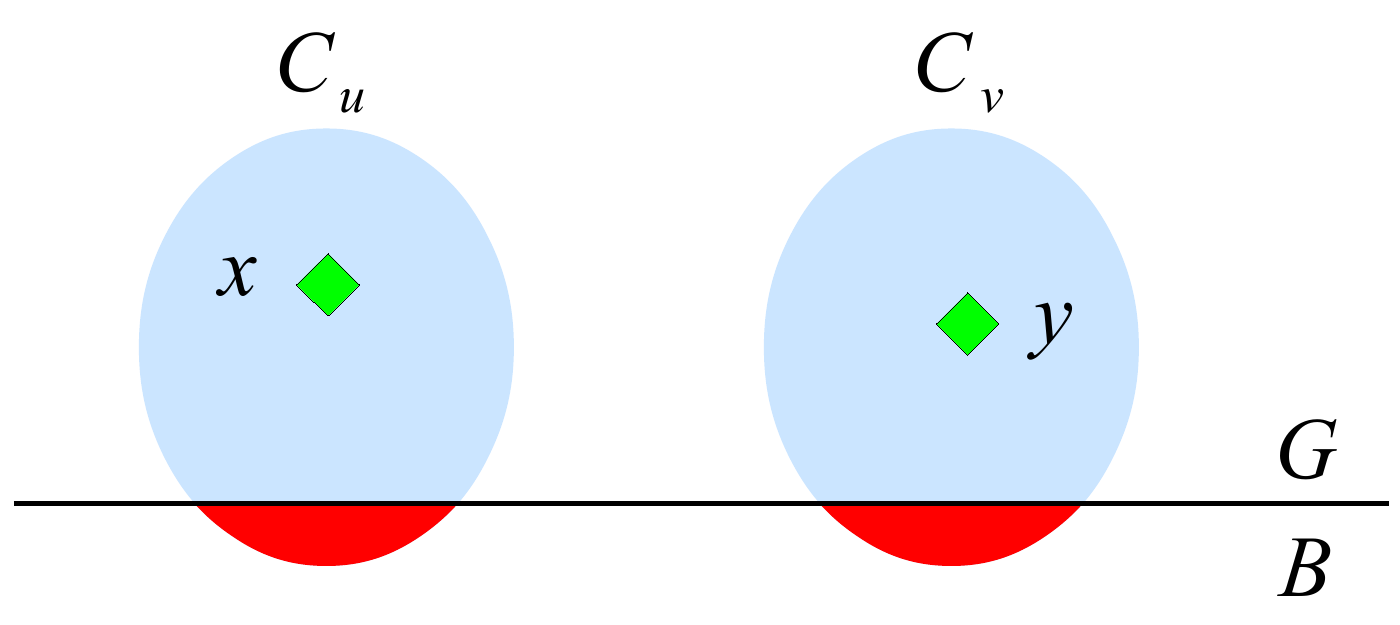}
\end{center}
\caption{
Illustration of the median test on $C_u$ and $C_v$.
At least $3/4$ fraction of the points in $C_u$ and $C_v$ are good points, and thus more than half
of the pairs $(x,y)$ with $x \in C_u$ and $y \in C_v$ are pairs of good points.
}\label{fig:medianTest}
\end{figure}

\begin{itemize}
\item If $C_u \in \mathcal{C}'_{t-1}$ contains good points from $C_i$ and $C_v \in \mathcal{C}'_{t-1}$ contains good points outside $C_i$, then they cannot be connected in $H_t$. \medskip \\
\hspace*{4mm} If both $C_u$ and $C_v$ are singleton \sets, say $C_u=\{x\},C_v=\{y\}$, then by Property~(a) of $F_t$, the common neighbors of $x$ and $y$ can only be bad points, and thus $C_u$ and $C_v$ cannot be connected.\\
If one of the two \sets\ (say $C_u$) is a singleton \set\ and the other is not,
then $C_u$ contains only one good point, and by Fact~\ref{fact:bad}, at least $3/4$ fraction of the points in $C_v$ are good points.
If both $C_u$ and $C_v$ are non-singleton blobs,
then by Fact~\ref{fact:bad}, at least $3/4$ fraction of the points in $C_u$ and $C_v$ are good points.
Therefore, in both cases, the number of pairs $(x,y)$ with good points $x \in C_u$ and $y \in C_v$ is at least $\frac{3}{4}|C_u| \times \frac{3}{4}|C_v| > \frac{|C_u||C_v|}{2} $.
That is, more than half of the pairs $(x,y)$ with $x \in C_u$ and $y \in C_v$ are pairs of good points;
see Figure~\ref{fig:medianTest} for an illustration.
This means there exist good points $x^* \in C_u, y^* \in C_v$ such that $S_t(x^*, y^*)$ is
no less than $\median_{x \in C_u, y \in C_v} S_t(x, y)$.
By the induction assumption, $x^*$ is a good point in $C_i$ and $y^*$ is a good point outside $C_i$.
Then by Property (a)(b) of $F_t$, $x^*$ and $y^*$ have no common neighbors in $F_t$, and thus $\median_{x \in C_u, y \in C_v} S_t(x, y)=0$.
Therefore, $C_u$ and $C_v$ are not connected in $H_t$.
\item If $C_u \in \mathcal{C}'_{t-1}$ contains good points from $C_i$, $C_v \in \mathcal{C}'_{t-1}$ contains good points outside $C_i$, and $C_w \in \mathcal{C}'_{t-1}$ contains only bad points, then $C_w$ cannot be connected to both $C_u$ and $C_v$ in $H_t$. \medskip\\
\hspace*{4mm} To prove this, assume for contradiction that $C_w$ is connected to both $C_u$ and $C_v$.
First, note the following fact about $C_w$.
Since any non-singleton \set\ must be formed in Step 4 in the algorithm and contain at least $\tmerge$ points
and thus cannot contain only bad points, $C_w$ must be a singleton \set, containing only a bad point $z$.\\
Next, we show that if $C_w=\{z\}$ is connected to $C_u$, then $z$ must be connected to some good point in $C_i$ in $F_t$.
If $C_u$ is a singleton \set, say $C_u=\{x\}$, then by Step 4 in the algorithm, $z$ and $x$ share more than $(\alpha+\nu) n$ common neighbors in $F_t$.
By Property (a)(b) of $F_t$, the common neighbors of $x$ and $z$ in $F_t$ can only be good points in $C_i$ or bad points.
Since there are at most $\nu n$ bad points, $z$ must be connected to some good point in $C_i$ in $F_t$.
If $C_u$ is not a singleton \set, then by Step 4 in the algorithm, $\median_{x\in C_u}S_t(x,z) > (|C_u| + |C_w|)/4 $.
By Fact~\ref{fact:bad}, at least $3/4$ fraction of the points in $C_u$ are good points.
So there exists a good point $x^* \in C_u$ such that $S_t(x^*, z) \geq \median_{x\in C_u}S_t(x,z)$,
which leads to $S_t(x^*, z) > (|C_u| + |C_w|)/4 > \nu n$.
By the induction assumption, $x^*$ is a good point in $C_i$.
Then by Property (a) of $F_t$, $x^*$ is only connected to good points from $C_i$ and bad points.
Since $S_t(x^*, z) > \nu n$, $z$ and $x^*$ must share some common neighbors
from $C_i$. Therefore, $z$ is connected to some good point in $C_i$ in $F_t$.\\
Similarly, if $C_w=\{z\}$ is connected to $C_v$, $z$ must be connected to some good point outside $C_i$ in $F_t$.
But then $z$ is connected to both a good point in $C_i$ and a good point outside $F_t$, which contradicts Property (b) of $F_t$.
\end{itemize}
By the properties of $H_t$, no connected component contains both good points in $C_i$ and good points outside $C_i$.
So Claim (1) is still true for the threshold $t$.
By induction, it is true for all thresholds.

Finally, we prove Claim (2).
First, at the threshold $t=|C_i|$, all good points in $C_i$ are connected in $F_t$.
This is because any good point in $C_i$ has at most $(\alpha + \nu) n$
neighbors outside $C_i$, so when $t = |C_i|$, any two good points in $C_i$ share at least $t-2(\alpha + \nu) n$ common neighbors and thus are connected in $F_t$.

Second, all \sets\ in $\mathcal{C}'_{t-1}$ containing good points in $C_i$ are connected in $H_t$. There are two cases.
\begin{itemize}
\item If no good points in $C_i$ have been merged, then all singleton \sets\ containing good points in $C_i$ will be connected in $H_t$. \medskip\\
\hspace*{4mm} This is because all good points in $C_i$ are connected in $F_t$, and thus they share at least $|G_i|\geq 6(\alpha+\nu)n -\nu n$
points as common neighbors in $F_t$.
\item If some good points in $C_i$ have already been merged into non-singleton \sets\ in $\mathcal{C}'_{t-1}$, we can show that in $H_t$ these non-singleton \sets\ will be connected to each other and connected to singleton \sets\ containing good points from $C_i$. \medskip\\
\hspace*{4mm} Consider two non-singleton \sets\ $C_u$ and $C_v$ that contain good points from $C_i$.
By Fact~\ref{fact:bad}, at least $3/4$ fraction of the points in $C_u$ and $C_v$ are good points.
So there exist good points $x^* \in C_u$ and $y^* \in C_v$ such that $S_t(x^*, y^*) \leq \median_{x \in C_u, y \in C_v} S_t(x, y)$.
By Claim (1), $x^*$ and $y^*$ must be good points in $C_i$.
Then they are connected to all good points in $C_i$ in $F_t$,
and thus $S_t(x^*, y^*)$ is
no less than the number of good points in $C_u$ and $C_v$, which is at least $3(|C_u| + |C_v|)/4$.
Now we have $\median_{x \in C_u, y \in C_v} S_t(x, y) \geq S_t(x^*, y^*) \geq 3(|C_u| + |C_v|)/4 > (|C_u| + |C_v|)/4$,
and thus $C_u, C_v$ are connected in $H_t$.\\
Consider a non-singleton \set\ $C_u$ and a singleton \set\ $C_v$ that contain good points from $C_i$.
The above argument also holds, so $C_u, C_v$ are connected in $H_t$.
\end{itemize}
Therefore, in both cases, all \sets\ in $\mathcal{C}'_{t-1}$ containing good points in $C_i$ are connected in $H_t$.
Then in Step 4, all good points in $C_i$ are merged into a \set\ in $\mathcal{C}'_t$.
\end{proof}

\begin{lemma} \label{lem:sec:rmnl:runtime}
Algorithm~\ref{alg:RMNL} has a running time of $O(n^{\omega+1})$.
\end{lemma}
\begin{proof}
The initializations in Step 1 take $O(n)$ time.
To compute $F_t$ in Step 2, for any $x \in S$, let $I_t(x,y) = 1$ if $y$ is within the $t$ nearest neighbors of $x$, and let $I_t(x,y) = 0$ otherwise.
Initializing $I_t$ takes $O(n^2)$ time.
Next we compute $N_t(x, y)$, the number of common neighbors between $x$ and $y$.
Notice that $N_t(x,y) = \sum_{z \in S} I_t(x,z) I_t(y,z) $,
so $N_t = I_t I_t^T$.
Then we can compute the adjacency matrix $F_t$ (overloading notation for the graph $F_t$) from $N_t$.
These steps take $O(n^{\omega})$ time.

To compute the graph $H_t$ in Step 3, first define $N\!S_t = F_t (F_t)^T$. Then for two points $x$ and $y$,
$N\!S_t(x,y)$ is the number of their common neighbors in $F_t$.
Further define a matrix $FC_t$ as follows:
if $x$ and $y$ are connected in $F_t$ and are in the same \set\ in $\mathcal{C}'_{t-1}$, then let $FC_t(x,y)=1$; otherwise, let $FC_t(x,y) = 0$.
As a reminder, for two points $x$ that belongs to $C_u \in \mathcal{C}'_{t-1}$ and $y$ that belongs to $C_v \in \mathcal{C}'_{t-1}$,
$S_t(x,y)$ is the number of points in $C_u \cup C_v$ they share as neighbors in common in $F_t$.
$FC_t$ is useful for computing $S_t$:
since for $x \in C_u$ and $y \in C_v$,
\begin{eqnarray*}
S_t(x,y) & = & \sum_{z \in C_v} F_t(x,z) F_t(y,z) + \sum_{z \in C_u} F_t(x,z) F_t(y,z) \\
& = & \sum_{z \in S} F_t(x,z) FC_t(y,z) + \sum_{z \in S} FC_t(x,z) F_t(y,z),
\end{eqnarray*}
we have $S_t = F_t (FC_t)^T + FC_t (F_t)^T$.
Based on $N\!S_t$ and $S_t$, we can then build the graph $H_t$.
All these steps take $O(n^{\omega})$ time.

When we perform merges in Step 4 or increase the threshold in Step 5, we need to recompute the above data structures, which takes $O(n^{\omega})$ time.
Since there are $O(n)$ merges and $O(n)$ thresholds, Algorithm~\ref{alg:RMNL} takes time $O(n^{\omega+1})$ in total.
\end{proof}

\section{A More General Property: Weak Good Neighborhood}\label{sec:weak}

\begin{figure}[h!]
\centering
    \subfloat[]{\includegraphics[width=0.33\textwidth]{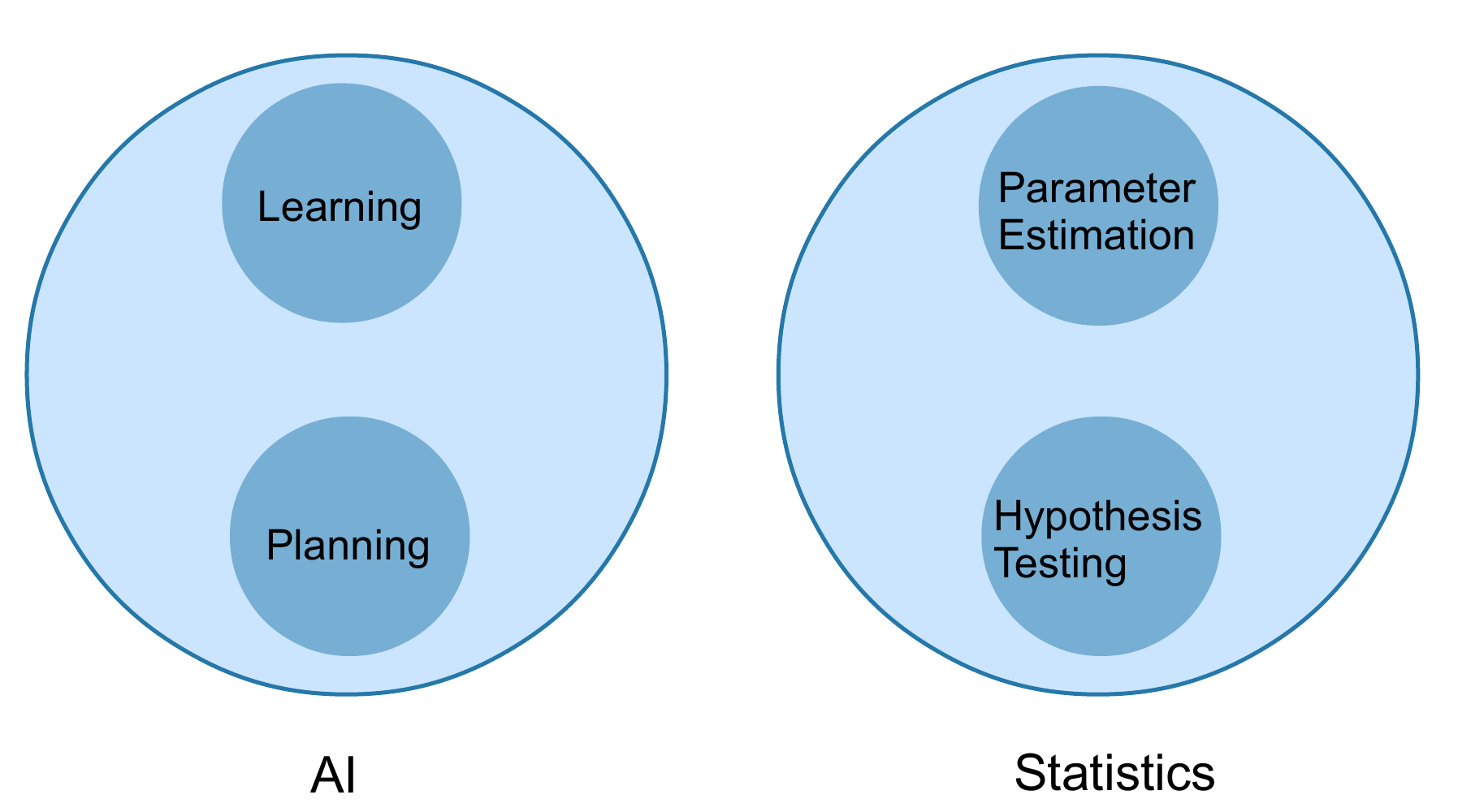}}
    \hspace*{.1in}
    \subfloat[]{\includegraphics[width=0.33\textwidth]{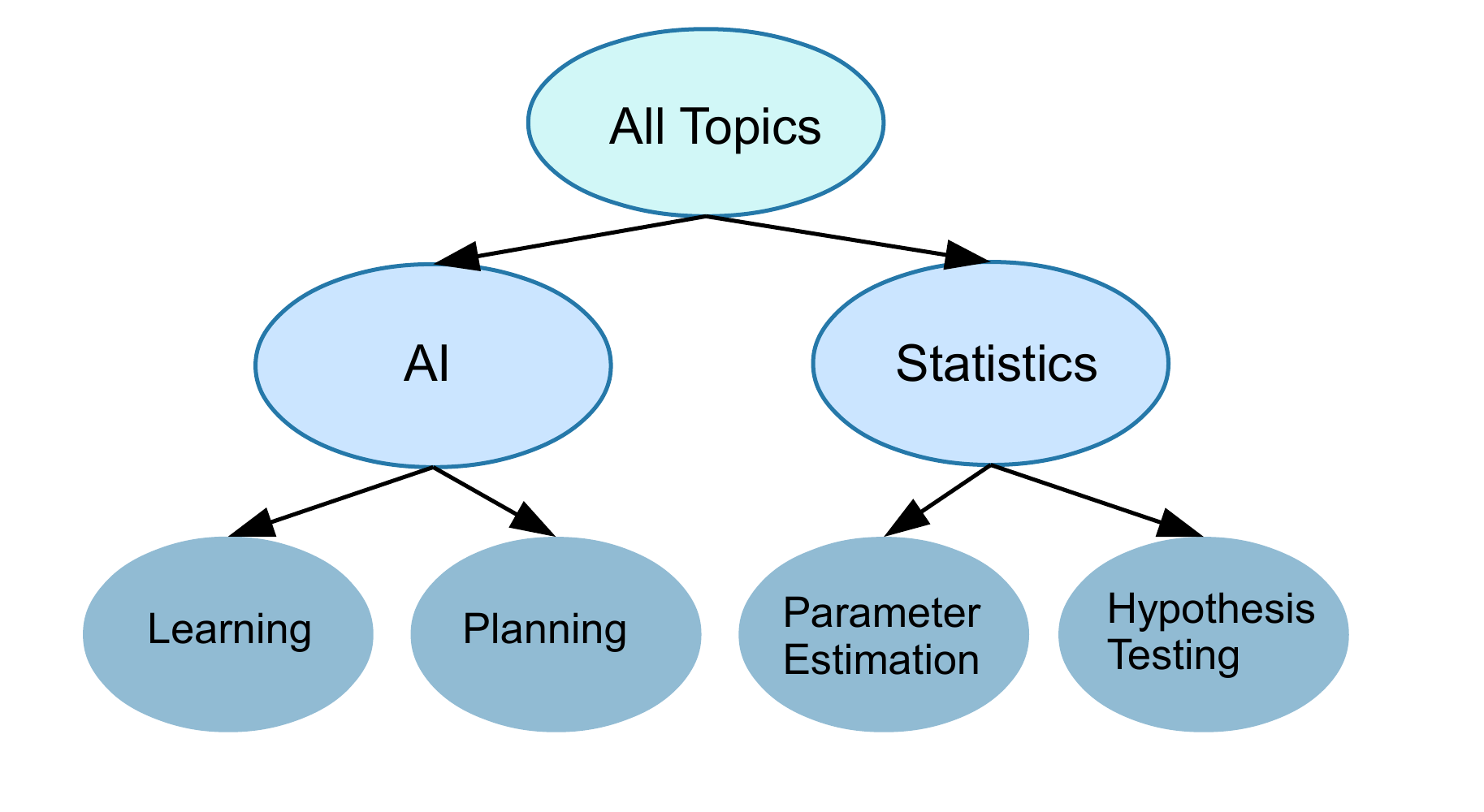}}
    \hspace*{.1in}
    \subfloat[]{\includegraphics[width=0.33\textwidth]{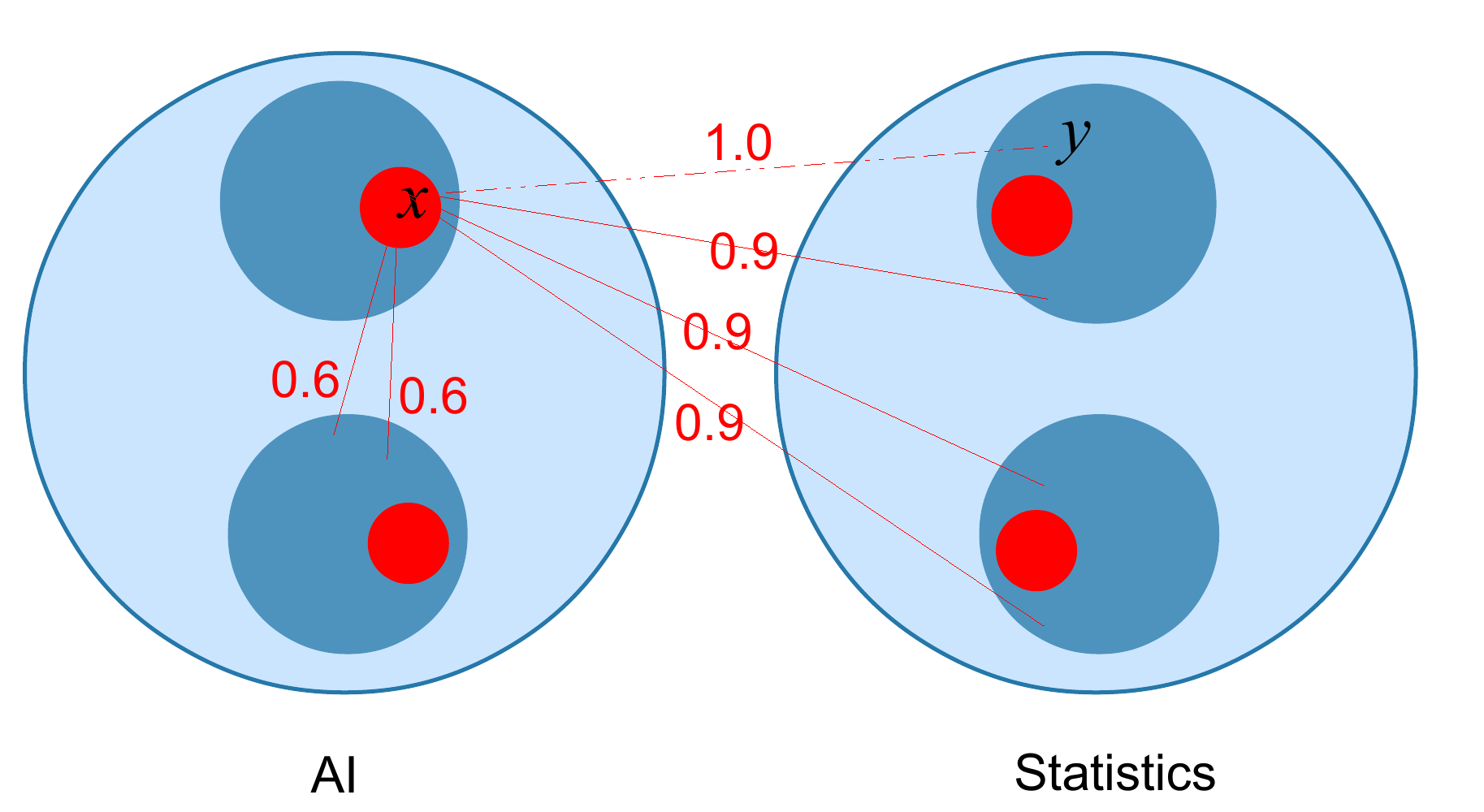}}
\caption{
{\small
Consider a document clustering problem. Assume that there are $n/4$ documents in each of the four areas: Learning, Planning, ParameterEstimation and HypothesisTesting. The first two belong to the field AI, and the last two belong to the field Statistics.
The similarities are specified as follows.
(1) $\simm(x,y) = 0.99$ if $x, y$ belong to the same area; (2) $\simm(x,y) = 0.8$ if $x, y$  belong to different areas in the same field; (3) $\simm(x,y) = 0.5$ if $x, y$ belong to different fields.
As shown in (b), there are four prunings: $\{\textrm{Learning}, \textrm{Planning}, \textrm{ParameterEstimation}, \textrm{HypothesisTesting}\}$, $\{\textrm{AI}, \textrm{ParameterEstimation}, \textrm{HypothesisTesting}\}$, $\{\textrm{Learning}, \textrm{Planning}, \textrm{Statistics}\}$, and $\{\textrm{AI}, \textrm{Statistics}\}$.
All these four prunings satisfy the strict separation property, and consequently satisfy the $\alpha$-good neighborhood property for $\alpha=0$.
However, this is no longer true if we take into account noise that naturally arises in practice.
As shown in (c), in each area, $1/8$ fraction of the documents lie close to the boundary between the two fields.
More precisely, the similarities for these boundary documents are defined as follows.
(1) These documents are very similar to some document in the other field:
for each boundary document $x$, we randomly pick one document $y$ in the other field and set $\simm(x,y) = \simm(y,x) = 1.0$;
(2) These documents are also closely related to the other documents in the other field: $\simm(x,y) = \simm(y,x) = 0.9$ when $x$ is a boundary document and $y$ belongs to the other field;
(3) These documents are not close to those in the other area in the same field:
$\simm(x,y) = \simm(y,x) = 0.6$ when $x$ is a boundary document and $y$ belongs to the other area in the same field.
Then the clustering $\{\textrm{AI}, \textrm{Statistics}\}$ satisfies the $(\alpha,\nu)$-good neighbor property only for $\alpha\geq 1/4$ or $\nu \geq 1/8$.
Similarly, $\{\textrm{AI}, \textrm{ParameterEstimation}, \textrm{HypothesisTesting}\}$ and $\{\textrm{Learning}, \textrm{Planning}, \textrm{Statistics}\}$ satisfy the property only for $\alpha\geq 1/4$ or $\nu \geq 1/16$.
See the text for more details, and see Section~\ref{sec:exp:syn} for simulations of this example and its variants.
}
}
\label{fig:weakProperty}
\end{figure}

In this section we introduce a weaker notion of good neighborhood property and prove that our algorithm also succeeds for
data satisfying this weaker property.

To motivate the property,
consider a point $x$ with the following neighborhood structure.
In the neighborhood of size $n_{C(x)}$, $x$ has a significant amount of its neighbors from other target clusters.
However, in a smaller, more local neighborhood, $x$ has most of its nearest neighbors from its target clusters $C(x)$.
In practice, points close to the boundaries between different target clusters typically have such neighborhood structure;
for this reason, points with such neighborhood are called boundary points.

We present an example in Figure~\ref{fig:weakProperty}.
A document close to the boundary between the two fields AI and Statistics has the following neighborhood structure:
out of its $n/4$ nearest neighbors, it has all its neighbors from its own field;
but out of its $n/2$ nearest neighbors, it has $n/4$ neighbors outside its field.
With $1/8$ fraction of such boundary points,
the clustering $\{\textrm{AI}, \textrm{Statistics}\}$ satisfies the $(\alpha,\nu)$-good neighbor property only for $\alpha\geq 1/4$ or $\nu \geq 1/8$.
This is because either we view all the boundary points as bad points in the $(\alpha, \nu)$-good neighborhood property which leads to $\nu \geq 1/8$,
or we need $\alpha\geq 1/4$ since a boundary point has $n/4$ neighbors outside its target cluster.
Similarly, $\{\textrm{AI}, \textrm{ParameterEstimation}, \textrm{HypothesisTesting}\}$ and
$\{\textrm{Learning}, \textrm{Planning}, \textrm{Statistics}\}$ satisfy the property only for $\alpha\geq 1/4$ or $\nu \geq 1/16$.

For this example, either $\alpha$ is too large so that Theorem~\ref{thm:RMNL}
is not applicable, or $\nu$ is too large so that the guarantee in Theorem~\ref{thm:RMNL} leads to constant error rate.
However, it turns out that our algorithm can still successfully produce a hierarchy as in Figure~\ref{fig:weakProperty}(b)
where the desired clusterings ($\{\textrm{AI}, \textrm{Statistics}\}$, \\$\{\textrm{Learning}, \textrm{Planning}, \textrm{ParameterEstimation}, \textrm{HypothesisTesting}\}$, $\{\textrm{Learning}, \textrm{Planning}, \textrm{Statistics}\}$, and \\ $\{\textrm{AI}, \textrm{ParameterEstimation}, \textrm{HypothesisTesting}\}$) are prunings of the hierarchy.
As we show, the reason is that each of these prunings satisfies a generalization of the good neighborhood property which takes into account the boundary points, and for which our algorithm still succeeds.
Note that the standard linkage algorithms fail on this example~\footnote{
For any fixed non-boundary point $y$ and fixed boundary point $x$ in the other field, the probability that $y$ has similarity $1.0$ only with $x$
is $\frac{2}{n}(1-\frac{2}{n})^{n/16-1} \approx \frac{2}{n} e^{-1/8}$.
Since there are $n/16$ such boundary points $x$ and $7n/8$ such non-boundary points $y$, when $n$ is sufficiently large, with high probability $n/12$ non-boundary points have similarity $1.0$ with one single boundary point.
Then the standard linkage algorithms (in particular, single linkage, average linkage, and complete
linkage) would first merge these pairs of points with similarity $1.0$.
From then on, no matter how they perform, any pruning of the hierarchy produced will have error higher than $1/12$.
}.
In the following, we first formalize this property and discuss how it relates to the properties of the similarity function
described in the paper so far.
We then prove that our algorithm succeeds under this property, correctly clustering all points that are not adversarially bad.

For clarity, we first relax the $\alpha$-good neighborhood to the weak $(\alpha, \beta)$-good neighborhood defined as follows.

\begin{property}\label{prop:weak}
A similarity function $\simm$ satisfies {\bf weak $(\alpha,\beta)$-good neighborhood} property for the clustering problem $(S,\ell)$,
if for each $p \in S$, there exists $A_p \subseteq C(p)$ of size greater than $6\alpha n$ such that $p\in A_p$ and
\begin{itemize}
\item any point in $A_p$ has at most $\alpha n$ neighbors outside $A_p$ out of the $|A_p|$ nearest neighbors;
\item for any such subset $A_p \subseteq C(p)$,
at least $\beta$ fraction of points in $A_p$ have all but at most $\alpha n$ nearest neighbors from $C(p)$ out of their $n_{C(p)}$ nearest neighbors.
\end{itemize}
\end{property}

Informally, the first condition implies that every point falls into a sufficiently large subset of its target cluster, and points in the subset
are close to each other in the sense that most of their nearest neighbors are in the subset.
This condition is about the local neighborhood structure of the points.
It shows that each point has a local neighborhood in which points closely relate to each other.
Note that the local neighborhood should be large enough so that the membership of the point is clearly established:
it should have size comparable to the number of connections to points outside ($\alpha n$).
Here we choose a minimum size of greater than $6 \alpha n$ mainly because it guarantees that our algorithm can still succeed in the worst case.
The second condition implies that for points in these large enough subsets, a majority of them have
most of their nearest neighbors from their target cluster.
This condition is about more global neighborhood structure. It shows how the subsets are closely related to those in the same target cluster in the neighborhood of size equal to the target cluster size.
Note that in this more global neighborhood, we do not require all points in these subsets have most nearest neighbors from their target clusters;
we allow the presence of $(1-\beta)$ fraction of points that may have a significant number of nearest neighbors outside their target clusters.

Naturally, as we can relax the $\alpha$-good neighborhood property to the $(\alpha, \nu)$-good neighborhood property,
we can relax the weak $(\alpha, \beta)$-good neighborhood to the weak $(\alpha, \beta, \nu)$-good neighborhood as follows.
Informally, it implies that the target clustering satisfies the weak $(\alpha,\beta)$-good neighborhood property after removing a few bad points.

\begin{property}\label{prop:weak}
A similarity function $\simm$ satisfies {\bf weak $(\alpha,\beta,\nu)$-good neighborhood} property for the clustering problem $(S,\ell)$,
if there exist a subset of points $B$ of size at most $\nu n$,
and for each $p \in S\setminus B$, there exists $A_p \subseteq C(p)\setminus B$ of size greater than $6(\alpha+\nu) n$ such that $p\in A_p$ and
\begin{itemize}
\item any point in $A_p$ has at most $\alpha n$ neighbors outside $A_p$ out of the $|A_p|$ nearest neighbors;
\item for any such subset $A_p \subseteq C_i \setminus B$,
at least $\beta$ fraction of points in $A_p$ have all but at most $\alpha n$ nearest neighbors from $C_i \setminus B$ out of their $|C_i \setminus B|$ nearest neighbors in $S\setminus B$.
\end{itemize}
\end{property}

For convenience, we call points in $B$ bad points.
If a point in $C_i \setminus B$ has all but at most $\alpha n$ nearest neighbors from $C_i \setminus B$ out of its $|C_i \setminus B|$ nearest neighbors in $S\setminus B$,
we call it a good point. Then the second condition in the definition can be simply stated as: any $A_p$ has at least $\beta$ fraction of good points.
Note that $C_i$ can contain points that are neither bad nor good.
Such points are called boundary points, since in practice such points typically lie close to the boundaries between target clusters.

As a concrete example, consider the clustering $\{\textrm{AI}, \textrm{Statistics}\}$ in Figure~\ref{fig:weakProperty}(c).
It satisfies the weak $(\alpha,\beta,\nu)$-good neighborhood property with probability at least $1-\delta$ when the number of points $n=O(\ln\frac{1}{\delta})$.
To see this, first note that for a fixed point $y$ and a fixed boundary point $x$ in the other field, the probability that $\simm(y,x)=1$ is $2/n$.
Since there are $n/16$ boundary point in the other field, by Hoeffding bound, the probability that $y$ has similarity $1$ with more than $n/32$ points is bounded by $\exp\{-2\cdot n/16 \cdot (1/2)^2\}=\exp\{-n/32\}$. By union bound, with probability at least $1-n\exp\{-n/32\}$,
no point has similarity $1$ with more than $n/32$ points.
Then by setting $A_p$ as the area that $p$ falls in, we can see that the clustering satisfies the weak $(\alpha, \beta, \nu)$-good neighborhood property for $\alpha=1/32, \beta= 7/8$ and $\nu=0$.
Note that there may also be some adversarial bad points.
Then the weak $(\alpha, \beta,\nu)$-good neighborhood property is satisfied when $\alpha=1/32, \beta= 7/8$ and $\nu$ is the fraction of bad points.
See Section~\ref{sec:exp:syn} for simulations of this example and its variants.

\subsection{Relating Different Versions of Good Neighborhood Properties}\label{sec:relate}

\begin{figure}[!h]
\centering
\includegraphics[scale = 0.6]{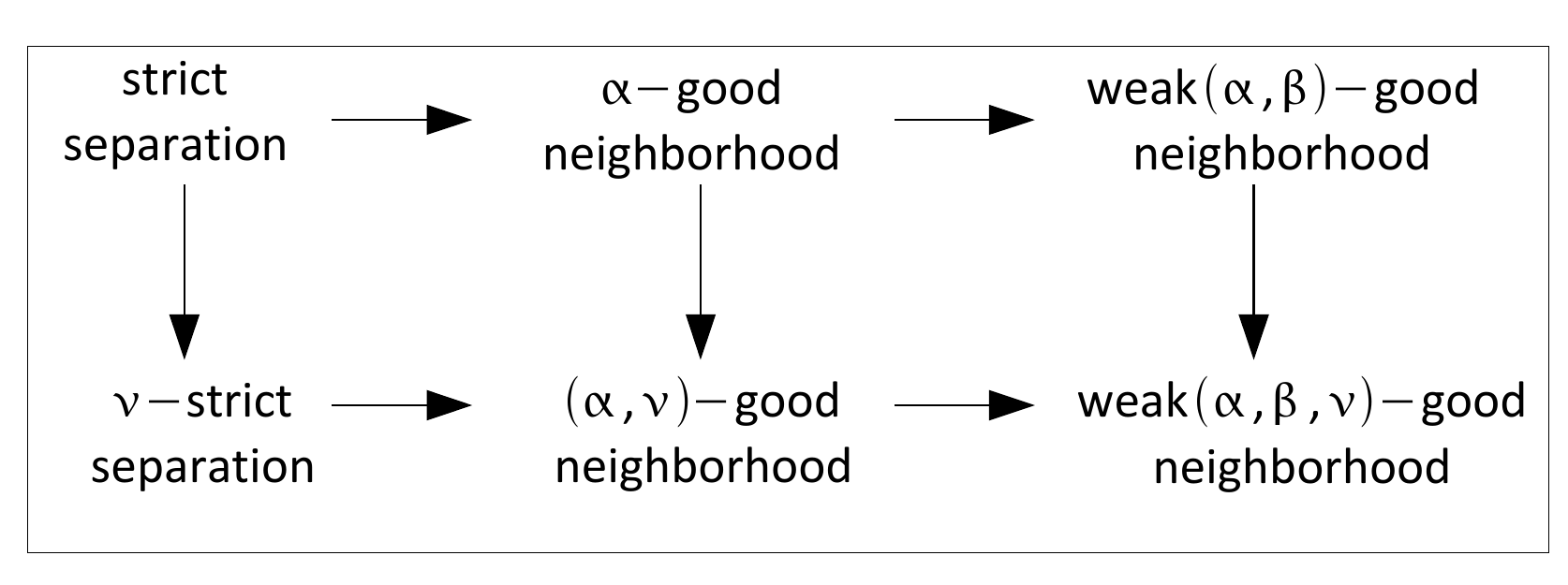}
 \caption{
{\small
Relations between various properties. The arrows represent generalization.
}
}
\label{fig:properties}
\end{figure}

The relations between these properties are illustrated in Figure~\ref{fig:properties}.
The relations between the weak good neighborhood properties and other properties are discussed below, while the other relations in the figure
follow from the facts in Section~\ref{subsec:properties}.

By setting $A_p = C_i$ for $p \in C_i$ in the definition, we can see that the weak $(\alpha,\beta)$-good neighborhood property is a
generalization of the $\alpha$-good neighborhood property when each target cluster has size greater than $6\alpha n$. Formally,

\begin{fact}
If  the similarity function $\simm$ satisfies the $\alpha$-\neighbprop\ property for the clustering problem
$(S,\ell)$ and $\min_i |C_i| > 6\alpha n$, then $\simm$ also satisfies the weak $(\alpha,\beta)$-\neighbprop\ property for the clustering problem $(S,\ell)$
for any $0<\beta\leq 1$.
\end{fact}
\begin{proof}
If $\simm$ satisfies the $\alpha$-\neighbprop\ property and $\min_i |C_i| > 6\alpha n$, then we have:
for any $p \in C_i$, there exists a subset $C_i \subseteq C_i$ of size greater than $6\alpha n$, such that
out of the $n_{C_i}$ nearest neighbors, any point $x \in C_i$ has at most $\alpha n$ neighbors outside $C_i$.
So $\simm$ satisfies both conditions of the weak $(\alpha,\beta)$-\neighbprop\ property.
\end{proof}

By setting $A_p = G_i$ for $p \in G_i$ in the definition, we can see that the weak $(\alpha, \beta, \nu)$-good neighborhood property generalizes the $(\alpha, \nu)$-good neighborhood property when each target cluster has size greater than $7(\alpha + \nu) n$.
Also, by setting $\nu = 0$, we can see that the weak $(\alpha,\beta)$-good neighborhood property is equivalent to the weak $(\alpha, \beta, 0)$-good neighborhood.

\begin{fact}
If the similarity function $\simm$ satisfies the $(\alpha, \nu)$-\neighbprop\ property for the clustering problem
$(S,\ell)$ and $\min_i |C_i| > 7(\alpha + \nu) n$, then $\simm$ also satisfies the weak $(\alpha, \beta, \nu)$-\neighbprop\ property
for the clustering problem $(S,\ell)$ for any $0<\beta\leq 1$.
\end{fact}
\begin{proof}
If $\simm$ satisfies the $(\alpha, \nu)$-\neighbprop\ property and $\min_i |C_i| > 7(\alpha + \nu) n$, then we have:
for any $p \in G_i = C_i \setminus B$, there exists a subset $G_i \subseteq G_i$ of size greater than $6(\alpha + \nu) n$, such that
 out of the $|G_i|$ nearest neighbors in $S\setminus B$,
 any good point $x \in G_i$ has at most $\alpha n$ neighbors outside $G_i$.
So $\simm$ satisfies both conditions of the weak $(\alpha, \beta,\nu)$-\neighbprop\ property.
\end{proof}

\begin{fact}
If  the similarity function $\simm$ satisfies the weak $(\alpha,\beta)$-\neighbprop\ property for the clustering problem
$(S,\ell)$, then  $\simm$ also satisfies the weak $(\alpha, \beta, 0)$-\neighbprop\ property for the clustering problem $(S,\ell)$.
\end{fact}
\begin{proof}
By setting $\nu = 0$ in the definition of the weak $(\alpha, \beta, \nu)$-\neighbprop\ property, we can see that it is
the same as the weak $(\alpha,\beta)$-\neighbprop\ property.
\end{proof}

\subsection{Correctness under the Weak Good Neighborhood Property}\label{sec:analysisWeak}

Now we prove that our algorithm also succeeds under the weak $(\alpha, \beta, \nu)$-good neighborhood property when $\beta \geq \minb$.
Formally,

\begin{theorem}\label{thm:RMNL_weak}
Let $\simm$ be a symmetric similarity function satisfying the weak $(\alpha, \beta, \nu)$-good neighborhood property for the clustering problem $(S,\ell)$
with $\beta \geq \minb$.
Then Algorithm~\ref{alg:RMNL} outputs a hierarchy
such that a pruning of the hierarchy is $\nu$-close to the target clustering in time $O(n^{\omega + 1})$,
where $O(n^{\omega})$ is the state of the art for matrix multiplication.
\end{theorem}

Theorem~\ref{thm:RMNL_weak} is a generalization of Theorem~\ref{thm:RMNL}, and the proof follows a similar reasoning.
The proof of correctness is from Lemma~\ref{lem:main_RMNL_weak} stated and proved below and the running time follows from Lemma~\ref{lem:sec:rmnl:runtime}.
The intuition is as follows.
First, by similar arguments as for the good neighborhood property,
each point $p$ in $S\setminus B$ will only be merged with other points in $A_p$ at $t \leq |A_p|$,
and all points in $A_p$ will belong to one \set\ at $t=|A_p|$ (Lemma~\ref{lem:blob_weak}), since in the local neighborhood of size $|A_p|$,
the point has most of its nearest neighbor from $A_p$.
Then, we need to show that such \sets\ will be correctly merged.
The key point is to show that even in the presence of boundary points,
the majority of points in such blobs are good points (Lemma~\ref{lem:disjointUnion}).
Then the median test can successfully distinguish blobs containing good points from different target clusters,
and our algorithm can correctly merge \sets\ from the same target clusters together.

To formally prove the correctness, we begin with Lemma~\ref{lem:blob_weak}.
The proof is similar to that for Lemma~\ref{lem:blob}, replacing $C_i$ with $A_p$.

\begin{lemma} \label{lem:blob_weak}
The following claims are true in Algorithm~\ref{alg:RMNL}:\\
(1) For any point $p \in S\setminus B$ and $t$ such that $ t \leq |A_p|$, any \set\ in $\mathcal{C}'_{t}$ containing points from $A_p$ will not contain points in $(S\setminus A_p)\setminus B$.\\
(2) For any point $p \in S\setminus B$ and $t = |A_p|$, all points in $A_p$ belong to one \set\ in $\mathcal{C}'_{t}$.
\end{lemma}

Lemma~\ref{lem:blob_weak} states that for any $p \in S\setminus B$, we will form $A_p$ before merging them with points outside.
Then we only need to make sure that these $A_p$ formed will be correctly merged.
More precisely, we need to consider the \sets\ that are ``fully formed'' in the following sense:

\begin{definition}
A \set\ $C_u \in \mathcal{C}'_t$ in Algorithm~\ref{alg:RMNL} is said to be fully formed
if for any point $p \in C_u \setminus B$, $A_p \subseteq C_u$.
\end{definition}

\begin{figure}[!t]
\begin{center}
\includegraphics[width=0.5\textwidth]{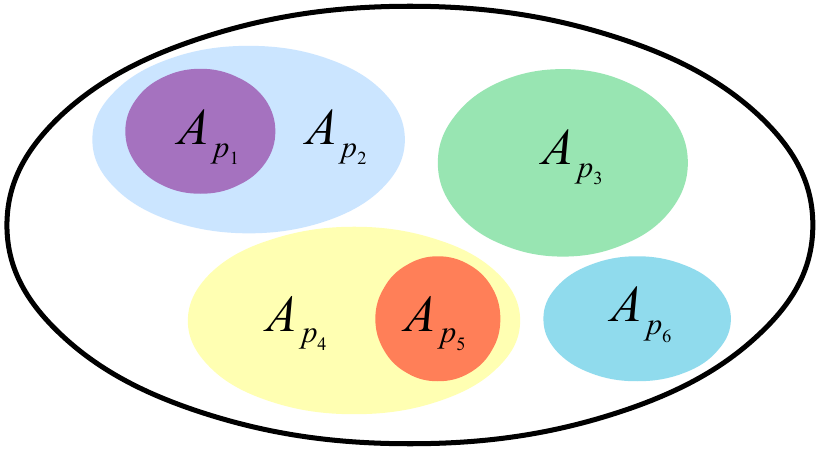}
\end{center}
\caption{
Illustration of a fully formed blob $C_u$: for any point $p \in C_u \setminus B$, $A_p \subseteq C_u$.
Then we can show that sets in $\{A_p: p \in C_u \setminus B\}$ are laminar, that is, for any $p, q\in C_u \setminus B$,
either $A_p \cap A_q = \emptyset$ or $A_p \subseteq A_q$ or $A_q \subseteq A_p$.
For example, in the figure we have $A_{p_5} \subseteq A_{p_4}$.
}\label{fig:partition}
\end{figure}

To show that fully formed \sets\ are correctly merged, the key point is to show that
the majority of points in such \sets\ are good points, and thus
the median test in the algorithm can successfully distinguish blobs containing good points from different target clusters.
This key point is in fact a consequence of Lemma~\ref{lem:blob_weak}:

\begin{lemma}\label{lem:disjointUnion}
For any fully formed \set\ $C_u \in \mathcal{C}'_t$ in Algorithm~\ref{alg:RMNL},
at least $\beta$ fraction of points in $C_u \setminus B$ are good points.
\end{lemma}

\begin{proof}
It suffices to show that there exist a set of points $P \subseteq  C_u \setminus B$, such that $\{A_p: p \in P\}$ is a partition of $C_u \setminus B$.
Clearly $C_u \setminus B = \cup_{p \in C_u \setminus B} A_p$. So we only need to show that sets in $\{A_p: p \in C_u \setminus B\}$
are laminar, that is, for any $p, q\in C_u \setminus B$, either $A_p \cap A_q = \emptyset$ or $A_p \subseteq A_q$ or $A_q \subseteq A_p$.
See Figure~\ref{fig:partition} for an illustration.

Assume for contradiction that there exist $A_p$ and $A_q$ such that $A_p \setminus A_q \neq \emptyset,
A_q \setminus A_p \neq \emptyset $ and $ A_p \cap A_q \neq \emptyset$. Without loss of generality, suppose $|A_p| \leq |A_q|$.
Then by the second claim in Lemma~\ref{lem:blob_weak}, when $t = |A_p|$, all points in $A_p$ belong to one \set\ in $\mathcal{C}'_t$.
In other words, this \set\ contains $A_p \cap A_q$ and $A_p \setminus A_q$. So for $t \leq |A_q|$, the \set\ contains points in $A_q$
and also points in $S\setminus B\setminus A_q$, which contradicts the first claim in Lemma~\ref{lem:blob_weak}.
\end{proof}

We are now ready to prove the following lemma that implies Theorem~\ref{thm:RMNL_weak}.

\begin{lemma}\label{lem:main_RMNL_weak}
The following claims are true in Algorithm~\ref{alg:RMNL}:\\
(1) For any $C_i$ such that $t\leq |C_i|$, any \set\ in $\mathcal{C}'_t$ containing points in $C_i\setminus B$ will not contain points in $(S\setminus C_i)\setminus B$.\\
(2) For any $C_i$ such that $t=|C_i|$, all points in $C_i \setminus B$ belong to one \set\ in $\mathcal{C}'_t$.
\end{lemma}

\begin{proof}
Before proving the claims, we first show that the graph $F_t$ constructed in Step 2 has the following useful properties by an argument similar to that in Lemma~\ref{lem:blob}.
Recall that $F_t$ is constructed on points in $S$ by connecting any two points that share at least $t-2(\alpha+\nu)n$ points in common
out of their $t$ nearest neighbors.
For any $C_i$ such that $t \leq |C_i|$, we have:

\begin{itemize}
\item[(a)] If $x$ is a good point in $C_i$ and $y$ is a good point outside $C_i$, then $x$ and $y$ are not connected in $F_t$. \medskip\\
\hspace*{4mm} To see this, first note that by Fact~\ref{fact:nn},
$x$ has at most $(\alpha + \nu) n$ neighbors outside $C_i$ out of the $t$ nearest neighbors.
Suppose $y$ is a good point from $C_j$.
If $n_{C_j} \geq t$, then $y$ has at most $(\alpha + \nu) n$ neighbors in $C_i$;
if $n_{C_j} < t$, $y$ has at most $(\alpha + \nu) n + t - n_{C_j}$ neighbors in $C_i$.
In both cases, $y$ has at most $(\alpha + \nu) n + \max(0, t- n_{C_j}) < t - 5(\alpha+\nu)n$ neighbors in $C_i$, since $n_{C_j} > 6(\alpha+\nu)n$
and $t > 6(\alpha+\nu)n$.
Then $x$ and $y$ have at most $t - 4(\alpha+\nu)n$ common neighbors, so they are not connected in $F_t$.
\item[(b)] If $x$ is a good point in $C_i$, $y$  is a good point outside $C_i$, and $z$ is a bad point, then $z$ cannot be connected to both $x$ and $y$ in $F_t$. \medskip\\
    \hspace*{4mm} To prove this, we will show that if $z$ is connected to $x$, then $z$ cannot be connected to $y$.
    First, by the same argument as above, out of the $t$ nearest neighbors, $y$ has less than $t - 5(\alpha + \nu) n$ neighbors in $C_i$.
    Second, by Fact~\ref{fact:nn}, $x$ has at most $(\alpha + \nu) n$ neighbors outside $C_i$.
    If $z$ has less than $t - 3(\alpha + \nu) n$ neighbors in $C_i$, then $z$ and $x$ share less than $t - 3(\alpha + \nu) n + (\alpha + \nu) n = t -2(\alpha+\nu) n$ neighbors and will not be connected.
    So $z$ must have at least $t - 3(\alpha + \nu) n$ neighbors in $C_i$, and thus cannot have more than $3(\alpha + \nu) n$ neighbors outside $C_i$. The two statements show that $y$ and $z$ share less than $t - 5(\alpha + \nu) n$ neighbors in $C_i$,
    and at most $3(\alpha + \nu) n$ neighbors outside $C_i$.
    So they share less than $t - 2(\alpha + \nu)n + 3(\alpha+\nu)n = t - 2(\alpha + \nu)n$ neighbors and thus are not connected in $F_t$.
\end{itemize}

Now we prove Claim (1) in the lemma by induction on $t$.
The claim is clearly true initially.
Assume for induction that the claim is true for the threshold $t-1$, that is,
for any $C_i$ such that $t-1 \leq |C_i|$, any \set\ in $\mathcal{C}'_{t-1}$ containing points in $C_i\setminus B$ will not contain points in $(S\setminus C_i)\setminus B$.
We now prove that the graph $H_t$ constructed in Step 3 has the following properties, which can be used to show that the claim is still true for the
threshold $t$.

\begin{itemize}
\item If $C_u \in \mathcal{C}'_{t-1}$ contains points from $C_i\setminus B$ and $C_v \in \mathcal{C}'_{t-1}$ contains points from $ (S \setminus C_i) \setminus B $, then they cannot be connected in $H_t$. \medskip\\
    \hspace*{4mm} Suppose one of them (say $C_u$) is not fully formed, that is, there is a point $p \in C_u \setminus B$
    such that $A_p \not \subseteq C_u$. Then by Lemma~\ref{lem:blob_weak},
    the algorithm will not merge $C_u$ with $C_v$ at this threshold.
    More precisely, since not all points in $A_p$ belong to $C_u$, we have $t-1 < |A_p|$ by Claim (2) in Lemma~\ref{lem:blob_weak}.
    Then by Claim (1) in Lemma~\ref{lem:blob_weak}, since $C_v$ contains points in $(S\setminus A_p)\setminus B$,
    $C_u$ and $C_v$ will not be merged in $\mathcal{C}'_t$. So they are not connected in $H_t$.\\
    So we only need to consider the other case
    when $C_u$ and $C_v$ are fully formed blobs.
    By Lemma~\ref{lem:disjointUnion}, the majority of points in the two \sets\ are good points.
    The good points from different target clusters have few common neighbors in $F_t$,
    then by the median test in our algorithm, the two \sets\ will not be connected in $H_t$.
    Formally, we can find two good points $x^* \in C_u, y^* \in C_v$ that satisfy the following two statements.
    \begin{itemize}
    \item $S_t(x^*, y^*) \geq \median_{x \in C_u, y \in C_v} S_t(x,y)$.\\
    By Lemma~\ref{lem:disjointUnion},
    at least $\beta\geq \minb$ fraction of points in $C_u \setminus B$ are good points.
    The fraction of good points in $C_u$ is at least $$\frac{\beta |C_u\setminus B|}{|C_u \setminus B|+|B|} \geq \frac{\minb \times 6(\alpha+\nu)n}{6(\alpha+\nu)n + \nu n} \geq \frac{3}{4}$$ since $|C_u \setminus B| \geq 6(\alpha+\nu)n$ and $|B|\leq \nu n$.
    Similarly, at least $\frac{3}{4}$ fraction of points in $C_v$ are good points.
    Then among all the pairs $(x,y)$ such that $x \in C_u, y\in C_v$, at least $\frac{3}{4} \times \frac{3}{4} > \frac{1}{2}$ fraction are pairs of good points.
    So there exist good points $x^* \in C_u, y^* \in C_v$ such that $S_t(x^*, y^*) \geq \median_{x \in C_u, y \in C_v} S_t(x,y)$.
    \item $S_t(x^*, y^*) \leq (|C_u| + |C_v|)/4$.\\
    The fraction of good points in $C_u\cup C_v$ is at least $\frac{3}{4}$.
    Since in $F_t$, good points in $C_u$ are not connected to good points in $C_v$,
    we have $S_t(x^*, y^*) \leq (|C_u| + |C_v|)/4$.
    \end{itemize}
    Combining the two statements, we have $\median_{x \in C_u, y \in C_v} S_t(x,y) \leq (|C_u| + |C_v|)/4$ and thus $C_u$ and $C_v$ are not connected in $H_t$.
\item If in $\mathcal{C}'_{t-1}$, $C_u$ contains points from $C_i\setminus B$, $C_v$ contains points from $ (S \setminus C_i) \setminus B $, and $C_w$ contains only bad points, then $C_w$ cannot be connected to both $C_u$ and $C_v$. \medskip\\
    \hspace*{4mm} By the same argument as above, we only need to consider the case when $C_u$ and $C_v$ are fully formed blobs.
    To prove the claim in this case, assume for contradiction that $C_w$ is connected to both $C_u$ and $C_v$.
    First, note the following fact about $C_w$. Since any non-singleton \set\ must be formed in Step 4 in the algorithm and contain at least $\tmerge$ points and thus cannot contain only bad points, $C_w$ must be a singleton \set, containing only a bad point $z$.\\
    Next, we show that if $C_w=\{z\}$ are connected to $C_u$ in $H_t$, then $z$ must be connected to at least one good point in $C_u$ in $F_t$.
    We have $\median_{x\in C_u} S_t(x,z) > \frac{|C_u| + |C_w|}{4}$,
    which means $z$ is connected to more than $\frac{|C_u|}{4}$ points in $C_u$ in $F_t$.
    By the same argument as above, at least $3/4$ fraction of points in $C_u$ are good points,
    then $z$ must be connected to at least one good point in $C_u$.\\
    Similarly, if $C_w$ is connected to $C_v$ in $H_t$, then $z$ must be connected to at least one good point in $C_v$ in $F_t$.
    But this contradicts Property (b) of $F_t$, so $C_w$ cannot be connected to both $C_u$ and $C_v$ in $H_t$.
\end{itemize}

By the properties of $H_t$, no connected component contains both points in $C_i \setminus B$ and points in $(S \setminus C_i) \setminus B$.
So Claim (1) is still true for the threshold $t$.
By induction, it is true for all thresholds.

Finally, we prove Claim (2).
By Lemma~\ref{lem:blob_weak}, when $t=|C_i|$, for any point $p \in C_i\setminus B$, $A_p$ belong to the same \set.
So all points in $C_i\setminus B$ are in sufficiently large \sets.
We will show that any two of these \sets\ $C_u, C_v$ are connected in $H_t$, and thus will be merged into one \set.
By Lemma~\ref{lem:disjointUnion}, we know that more than $3/4$ fraction of points in $C_u$ ($C_v$ respectively) are good points,
and thus there exist good points $x^* \in C_u, y^* \in C_v$ such that $S_t(x^*, y^*) \leq \median_{x \in C_u, y \in C_v} S_t(x, y)$.
By Claim (1), all good points in $C_u$ and $C_v$ are from $C_i$, so they share at least $t - 2 (\alpha + \nu) n$ neighbors when $t=|C_i|$,
and thus are connected in $F_t$.
Then $S_t(x^*,y^*)$ is at least the number of good points in $C_u \cup C_v$,
which is at least $3(|C_u| +  |C_v|)/4$.
Then $\median_{x \in C_u, y \in C_v} S_t(x, y) \geq S_t(x^*,y^*) > (|C_u| + |C_v|) /4$.
Therefore, all \sets\ containing points from $C_i\setminus B$ are connected in $H_t$ and thus merged into a \set.
\end{proof}


\renewcommand{\NN}{N\!N}

\section{The Inductive Setting} \label{sec:inductive}
Many clustering applications have recently faced an explosion of data, such as in astrophysics and biology.
For large data sets, it is often resource and time intensive to run an algorithm over the entire data set.
It is thus increasingly important to develop algorithms that can remove the dependence on the actual size of the data and still perform reasonably well.

In this section we consider an inductive model that formalizes this problem.
In this model, the given data is merely a small random subset of points
from a much larger data set.
The algorithm outputs a hierarchy over the sample, which also implicitly represents a hierarchy over the data set.
In the following we describe the inductive version of our algorithm and prove that
when the data satisfies the good neighborhood properties, the algorithm achieves small error on the entire data set,
requiring only a small random sample whose size is independent of that of the entire data set.

\subsection{Formal Definition}
First we describe the formal definition of the inductive model.
In this setting, the given data $\fset$ is merely a small random subset of points from a much larger abstract instance space $\xspace$.
For simplicity, we assume that $\xspace$ is finite and that the underlying distribution is uniform over $\xspace$.
Let $N=|X|$ denote the size of the entire instance space, and let $n = |S|$ denote the size of the sample.

Our goal is to design an algorithm that based on the sample produces a hierarchy of small error
with respect to the whole distribution.
Formally, we assume that each node $u$ in the hierarchy derived
over the sample induces a cluster (a subset of $X$).
For convenience, $u$ is also used to denote the \set\ of sampled points it represents.
The cluster $u$ induces over $X$ is implicitly represented as a function $f_u: \xspace \rightarrow \{0,1\}$,
that is, for each $x \in X$, $f_u(x) = 1$ if $x$ is a point in the cluster and $0$ otherwise.
We say that the hierarchy has error at most $\epsilon$ if it has a pruning $f_{u_1},\dots,f_{u_k}$ of error at most $\epsilon$.

\subsection{Inductive Robust Median Neighborhood Linkage} \label{sec:ind_rmnl}

The inductive version of our algorithm is described in Algorithm~\ref{alg:ind_RMNL}.
To analyze the algorithm, we first present the following lemmas showing that, when the data satisfies the good neighborhood property, a sample of sufficiently large size also satisfies the weak good neighborhood property.

\begin{algorithm}[H]
\caption{ Inductive Robust Median Neighborhood Linkage}\label{alg:ind_RMNL}
\begin{algorithmic}
\STATE \textbf{Input:}  similarity function $\simm$, $n \in \mathbf{Z}^+$,
    parameters $\alpha > 0, \nu>0$.
\STATE
\STATE $\mybullet$ {\tt Get a hierarchy on the sample}
\STATE Sample i.i.d.\ examples $S=\{x_1, \ldots, x_{n}\}$ uniformly at random from $\xspace$.
\STATE Run \algref{alg:RMNL} with parameters $(2\alpha, 2\nu)$ on $S$ and obtain a hierarchy $\tree$.
\STATE
\STATE $\mybullet$ {\tt Get the implicit hierarchy over $X$}
\FOR{any $\x \in \xspace$}
\STATE Let $N_S(x)$ denote the $6(\alpha + \nu) n$ nearest neighbors of $x$ in $S$.
\STATE Initialize $u = \mathrm{root}(T)$ and $f_u(x) = 1$.
\WHILE{$u$ is not a leaf}
\STATE Let $w$ be the child of $u$ that contains the most points in $N_S(x)$.
\STATE Set $u=w$ and $f_u(x) = 1$.
\ENDWHILE
\ENDFOR
\STATE
\STATE \textbf{Output:} Hierarchy $T$ and $\{f_u, u\in T\}$.
\end{algorithmic}
\end{algorithm}

\begin{lemma}
\label{lem:ind_good0}
 Let $\simm$ be a symmetric similarity function satisfying the
$(\alpha, \nu)$-\neighbprop\ for the clustering problem $(X,\ell)$.
Consider any fixed $x \in  X\setminus B$.
If the sample size satisfies $\dimmn= \Theta\left(\frac{1}{\alpha} \ln{\frac{1}{\delta}} \right)$,
then with probability at least $1-\delta$, $x$ has at most $2\alpha n$ neighbors outside $(C(x)\setminus B) \cap S$
out of the $|(C(x)\setminus B) \cap S|$ nearest neighbors in $S \setminus B$.
\end{lemma}

\begin{proof}
Suppose $x \in G_i$.
Let $\NN(x)$ denote its $|G_i|$ nearest  neighbors in $X$.
By assumption we have that
$|\NN(x) \setminus G_i| \leq \alpha N$ and $|G_i\setminus \NN(x)| \leq \alpha N$.
Then by Chernoff bounds, with probability at least $1 - \delta$ at most $2\alpha n$ points in our sample are in $\NN(x) \setminus G_i$ and at most $2\alpha n$ points in our sample are in $G_i \setminus \NN(x)$.

We now argue that at most $2\alpha  n$ of the $|G_i \cap S|$ nearest neighbors of $x$ in $S \setminus B$ can be outside $G_i$.
Let $n_1$ be the number of points in $(\NN(x) \setminus G_i)\cap S$,
$n_2$ be the number of points in $(G_i \setminus \NN(x))\cap S$, and $n_3$ be the number of points in $(G_i \intersect \NN(x))\cap S$.
Then $|G_i \cap S| = n_2 + n_3$ and we know that $n_1 \leq 2\alpha  n$, $n_2 \leq 2\alpha  n$.
We consider the following two cases.
\begin{itemize}
    \item $n_1 \geq n_2$.
    Then $n_1 + n_3 \geq n_2 + n_3 = |G_i \cap S|$.
    This implies that the $|G_i \cap S|$ nearest neighbors of $x$ in the sample all lie inside $\NN(x)$, since by definition all points inside $\NN(x)$ are closer to $x$ than any point outside $\NN(x)$.
    But we are given that at most $n_1 \leq 2\alpha  n$ of them can be outside $G_i$.
    Thus, we get that at most $2\alpha  n$ of the $|G_i \cap S|$ nearest neighbors of $x$ are not from $G_i$.
    \item $n_1 < n_2$.
    This implies that the $|G_i \cap S|$ nearest  neighbors of $x$ in the sample include {\em all} the points in $\NN(x)$ in the sample, and possibly some others too.
    But this implies in particular that it includes all the $n_3$ points in $G_i \intersect \NN(x)$ in the
    sample. So, it can include at most $|G_i \cap S| - n_3 = n_2 \leq 2\alpha n$ points not in $G_i \intersect \NN(x)$.
    Even if all those are not in $G_i$, the $|G_i \cap S|$ nearest neighbors of $x$ still include at most $2\alpha  n$ points not from $G_i$.
\end{itemize}
In both cases, at most $2\alpha  n$ of the $|G_i \cap S|$ nearest neighbors of $x$ in $S \setminus B$ can be outside $G_i$.
\end{proof}

\begin{lemma}
\label{lem:ind_good}
 Let $\simm$ be a symmetric similarity function satisfying the
$(\alpha, \nu)$-\neighbprop\ for the clustering problem $(X,\ell)$.
If the sample size satisfies $\dimmn= \Theta\left(\frac{1}{\min(\alpha,\nu)} \ln{\frac{1}{\delta \min(\alpha,\nu)}} \right)$,
then with probability at least $1-\delta$, $\simm$ satisfies the $(2\alpha, 2\nu)$-\neighbprop\ with respect to the clustering induced
over the sample $S$.
\end{lemma}

\begin{proof}
First, by Chernoff bounds, when $n\geq \frac{3}{\nu}\ln\frac{2}{\delta}$,
we have that with probability at least $1-\delta/2$, at most $2\nu n$ bad points fall into the sample.

Next, by Lemma~\ref{lem:ind_good0} and union bound, when $\dimmn= \Theta\left(\frac{1}{\alpha} \ln{\frac{n}{\delta}} \right)$
we have that with probability at least $1-\delta/2$,
for any $C_i$, any $x \in G_i \cap S$, $x$ has at most $2\alpha n$
points outside $G_i \cap S$ out of its $|G_i \cap S|$ nearest neighbors in $(X\setminus B) \cap S$.
Therefore, if $\dimmn= \Theta\left(\frac{1}{\min(\alpha,\nu)} \ln{\frac{n}{\delta}} \right)$, then with probability at least $1 - \delta$, the similarity function satisfies the $(2\alpha, 2\nu)$-\neighbprop\ property with respect to the clustering induced over the sample $S$.

It now suffices to show $\dimmn$ is large enough so that $\dimmn= \Theta\left(\frac{1}{\min(\alpha,\nu)} \ln{\frac{n}{\delta}} \right)$.
To see this, let $\eta=\min(\alpha,\nu)$.
Since $\ln n \leq t n - \ln t -1$ for any $t, n>0$, we have
$$\frac{c}{\eta}\ln n \leq \frac{c}{\eta}\left(\frac{\eta}{2c} n + \ln\frac{2c}{\eta} -1\right) = \frac{n}{2} + \frac{c}{\eta}\ln\frac{2c}{e \cdot \eta}$$
for any constant $c >0$. Then $\dimmn = \Theta\left(\frac{1}{\eta} \ln{ \frac{1}{\eta}} \right)$ implies $\dimmn = \Theta\left(\frac{1}{\eta} \ln n\right)$, and $\dimmn= \Theta\left(\frac{1}{\eta} \ln{ \frac{1}{\delta\cdot\eta}} \right)$
implies $\dimmn=\Theta\left(\frac{1}{\eta} \ln{\frac{n}{\delta}} \right)$.
\end{proof}

\begin{theorem}  \label{thm:ind_rmnl}
Let $\simm$ be a symmetric similarity function satisfying the
$(\alpha, \nu)$-\neighbprop\ for the clustering problem $(X,\ell)$.
As long as the smallest target cluster
has size greater than $12( \nu + \alpha) N$, then
 \algref{alg:ind_RMNL} with parameters
$\dimmn= \Theta\left(\frac{1}{\min(\alpha,\nu)} \ln{ \frac{1}{\delta\cdot\min(\alpha,\nu)}} \right)$
produces a hierarchy with a pruning that is $(\nu+\delta)$-close to the target clustering with probability $1-\delta$.
\end{theorem}

\begin{proof}
Note that by \lemref{lem:ind_good}, with
probability at least $1-\delta/4$, we have that $\simm$ satisfies the $ (2\alpha, 2\nu)$-\neighbprop\
with respect to the clustering induced over the sample.
Moreover, by Chernoff bounds, with
probability at least $1-\delta/4$, each $G_i$ has at
least $6(\nu + \alpha)n$ points in the sample.
Then by Theorem~\ref{thm:RMNL}, \algref{alg:RMNL} outputs a hierarchy $T$ on the sample $S$ with
a pruning that assigns all good points correctly.
Denote this pruning as $\{u_1,\dots,u_k\}$ such that $u_i \setminus B = (C_i\cap S) \setminus B$.

Now we want to show that $f_{u_1}, \dots, f_{u_k}$ have error at most $\nu+\delta$ with probability at least $1-\delta/2$.
For convenience, let $u(x)$ be a shorthand of $u_{\ell(x)}$.
Then it is sufficient to show that with probability at least $1-\delta/2$, a $(1-\delta)$ fraction of points $x \in X \setminus B$ have $f_{u(x)}(x)=1$.

Fix $C_i$ and a point $x \in C_i\setminus B$.
By \lemref{lem:ind_good0}, with probability at least $1-\delta^2/2$,
out of the $|G_i \cap S|$ nearest neighbors of $x$ in $S\setminus B$,
at most $2\alpha n$ can be outside $G_i$.
Recall that \algref{alg:ind_RMNL} checks $N_S(x)$, the $6(\alpha + \nu) n$ nearest neighbors of $x$ in $S$.
Then out of $N_S(x)$, at most $2(\alpha+\nu)n$ points are outside $G_i\cap S$.
By Lemma~\ref{lem:blob}, $u_i$ contains $G_i\cap S$, so
$u_i$ must contain at least $4(\alpha+\nu)n$ points in $N_S(x)$.
Consequently, any ancestor $w$ of $u_i$, including $u_i$, has more points in $N_S(x)$ than any other sibling of $w$.
Then we must have $f_w(x) =1$ for any ancestor $w$ of $u_i$. In particular, $f_{u_i}(x)=1$.
So, for any point $x\in X\setminus B$, with probability at least $1-\delta^2/2$ over the draw of the random sample,
$f_{u(x)}(x)=1$.

Then by Markov inequality, with probability at least $1-\delta/2$, a $(1-\delta)$ fraction of points $x \in X \setminus B$ have $f_{u(x)}(x)=1$.
More precisely, let $U_x$ denote the uniform distribution over $X\setminus B$, and let $U_S$ denote the distribution of the sample $S$.
Let $I(x,S)$ denote the event that $f_{u(x)}(x) \neq 1$. Then we have
$$\mathbf{E}_{x \sim U_x, S \sim U_S}[I(x,S)] = \mathbf{E}_{S \sim U_S}\biggl[ \mathbf{E}_{x \sim U_x}[I(x,S)|S] \biggr]  \leq \delta^2/2.$$
Then by Markov inequality, we have
$${\Pr}_{S \sim U_S} \biggl[ \mathbf{E}_{x \sim U_x}[I(x,S)|S] \geq \delta \biggr]  \leq \delta/2$$
which means that with probability at least $1-\delta/2$ over the draw of the random sample $S$, a $(1-\delta)$ fraction of points $x \in X \setminus B$ have $f_{u(x)}(x)=1$.
\end{proof}

Similarly, Algorithm~\ref{alg:ind_RMNL} also succeeds for the weak good neighborhood property.
By similar arguments as those in Lemma~\ref{lem:ind_good0} and~\ref{lem:ind_good}, we can prove that $\simm$ satisfies the weak good neighborhood property over a sufficiently large sample (Lemma~\ref{lem:ind_good_weak}),
which then leads to the final guarantee Theorem~\ref{thm:ind_rmnl_weak}.
For clarity, the proofs are provided in Appendix~\ref{app:ind_good_weak}.

\begin{lemma}
\label{lem:ind_good_weak}
Let $\simm$ be a symmetric similarity function satisfying the
weak $(\alpha, \beta, \nu)$-\neighbprop\ for the clustering problem $(X,\ell)$.
Furthermore, it satisfies that for any $p \in X\setminus B$, $|A_p| > 24 (\alpha+\nu)N$.
If the sample size satisfies $\dimmn= \Theta\left(\frac{1}{\min(\alpha,\nu)} \ln{\frac{1}{\delta \min(\alpha,\nu)}} \right)$,
then with probability at least $1-\delta$, $\simm$ satisfies the $(2\alpha, \frac{15}{16}\beta, 2\nu)$-\neighbprop\ with respect to the clustering induced
over the sample $S$.
\end{lemma}

\begin{theorem}  \label{thm:ind_rmnl_weak}
Let $\simm$ be a symmetric similarity function satisfying the
weak $(\alpha, \beta, \nu)$-\neighbprop\ for the clustering problem $(X,\ell)$ with $\beta\geq \frac{14}{15}$.
Furthermore, it satisfies that for any $p \in X\setminus B$, $|A_p| > 24 (\alpha+\nu)N$.
Then \algref{alg:ind_RMNL} with parameters
$\dimmn= \Theta\left(\frac{1}{\min(\alpha,\nu)} \ln{ \frac{1}{\delta\cdot\min(\alpha,\nu)}} \right)$
produces a hierarchy with a pruning that is $(\nu+\delta)$-close to the target clustering with probability $1-\delta$.
\end{theorem}


\newcommand{\stable}{stable}

\section{Experiments}  \label{sec:exp}

In this section, we compare our algorithm (called RMNL for convenience) with popular hierarchical clustering algorithms, including
standard linkage algorithms~\citep{Sneath73numericaltaxonomy,KING67step-wiseclustering,everitt2011cluster}, (Generalized) Wishart's
Method~\citep{Wishart69modeanalysis,Sanjoy2010}, Ward's minimum variance method~\citep{Ward1963}, CURE~\citep{Guha_cure:an98}, and EigenCluster~\citep{CKVW06}.

To evaluate the performance of the algorithms, we use the model discussed in~\secref{sec:model}.
Given a hierarchy output by an algorithm, we generate all possible prunings of size $k$\footnote{Note that we generate all prunings of size $k$  for evaluating the performance of various algorithms only. The hierarchical clustering algorithms do not need to generate these prunings when creating the hierarchies.},
where $k$ is the number of clusters in the target clustering.
Then we compute the \emph{Classification Error} of each pruning with respect to the target clustering,
and report the best error.
The Classification Error of a computed clustering $h$ with respect to the target clustering $\ell$
is the probability that a point chosen at random from the data is labeled incorrectly
\footnote{To compute this error for a computed clustering in polynomial time, we first find its best match to the target clustering using the Hungarian Method~\citep{kuhn1955hungarian} for min-cost bipartite matching in time $O(n^3)$, and then calculate the error as the fraction of points misclassified in matched clusters.}.
Formally,
\begin{equation*}
\mathbf{err}(h) = \min_{\sigma \in \simgroup_k}{\left[\Pr_{\x \in \fset}{\left[\sigma(h(\x)) \neq \ell(\x)\right]}\right]}
\end{equation*}
where $\simgroup_k$ is the set of all permutations on $\{1,\ldots,k\}$.
For reporting results, we follow the methodology in~\citep{Guha_cure:an98}: for all algorithms accepting input parameters
(including (Generalized) Wisharts' Method, CURE, and RMNL),
the experiments are repeated on the same data over a range of input parameter values, and the best results are considered.

\smallskip
\noindent
{\bf Data sets}
To emphasize the effect of noise on different algorithms, we perform controlled experiments
on a synthetic data set AIStat. This data set contains $512$ points. It is an instance of the example discussed in
Section~\ref{sec:weak} and is described in Figure~\ref{fig:weakProperty}.
We further consider the following real-world data sets from UCI Repository~\citep{Frank+Asuncion:2010}:
Wine, Iris, BCW (Breast Cancer Wisconsin), BCWD (Breast Cancer Wisconsin Diagnostic), Spambase, and Mushroom.
We also consider the MNIST data set~\citep{lecun1998gradient} and use two subsets of the test set for our experiments:
Digits0123 that contains the examples of the digits $0, 1, 2, 3$,
and Digits4567 that contains the examples of the digits $4, 5, 6, 7$.

We additionally consider the $10$ data sets (PFAM1 to PFAM10) from~\citep{Voevodski:2012:ACB},
which are created by randomly choosing $8$ families (of size between $1000$ and $10000$)
from the biology database Pfam~\citep{pfam:Punta01012012}, version 24.0, October 2009.
The similarities for the PFAM data sets are generated by biological sequence alignment software BLAST~\citep{blast:altsbl90}.
BLAST performs one versus all queries by aligning a queried sequence to sequences in the data set, and
produces a score for each alignment.
The score is a measure of the alignment quality and thus can be used as similarity.
However, BLAST does not consider alignments with some of the sequences,
in which case we assign similarities $0$ to the corresponding sequences and exclude them from the neighbors of the queried sequence.

The smaller data sets are used in the transductive setting: Wine ($178$ points of dimension $13$), Iris ($150 \times 4$), BCW ($699 \times 10$), and BCWD ($569 \times 32$). The larger ones are used in the inductive setting:  Spambase ($4601 \times 57$), Mushroom ($8124 \times 22$), Digits0123 ($4157 \times 784$), Digits4567 ($3860 \times 784$), and PFAM1 to PFAM10 ($10000 \sim 100000$  sequences each).

\subsection{Synthetic Data}\label{sec:exp:syn}

\newcommand{\synFigScale}{0.32}
\newcommand{\synHSpace}{0.1in}
\newcommand{\synHMargin}{.in}
\newcommand{\synVSpace}{-0.2in}
\begin{figure}[!t]
    \subfloat[$\alpha$]{\includegraphics[scale = \synFigScale]{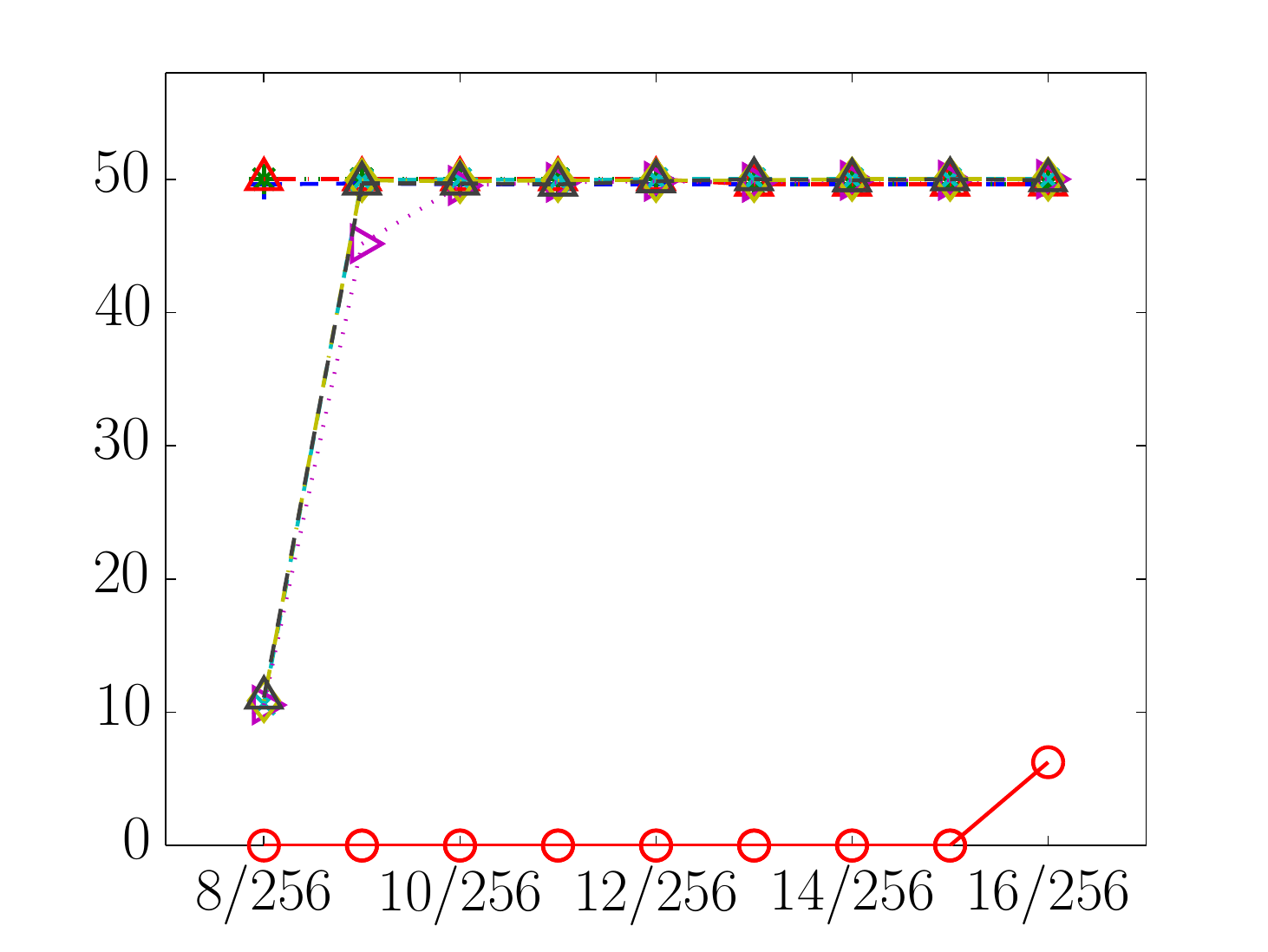}}
    \hspace*{\synHSpace}
    \subfloat[$\nu$]{\includegraphics[scale = \synFigScale]{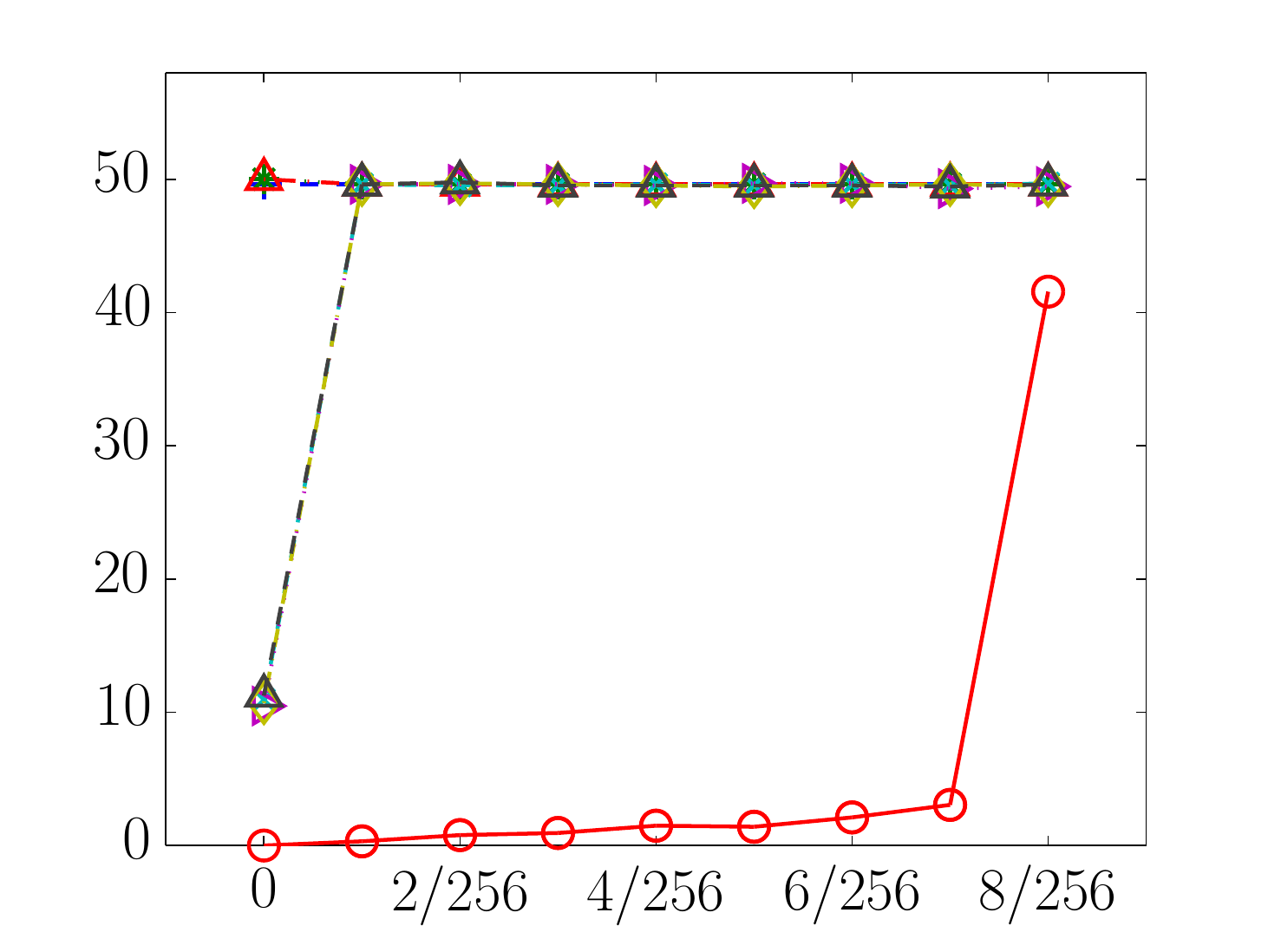}}
    \hspace*{\synHSpace}
    \subfloat[$\alpha+\nu$]{\includegraphics[scale=\synFigScale]{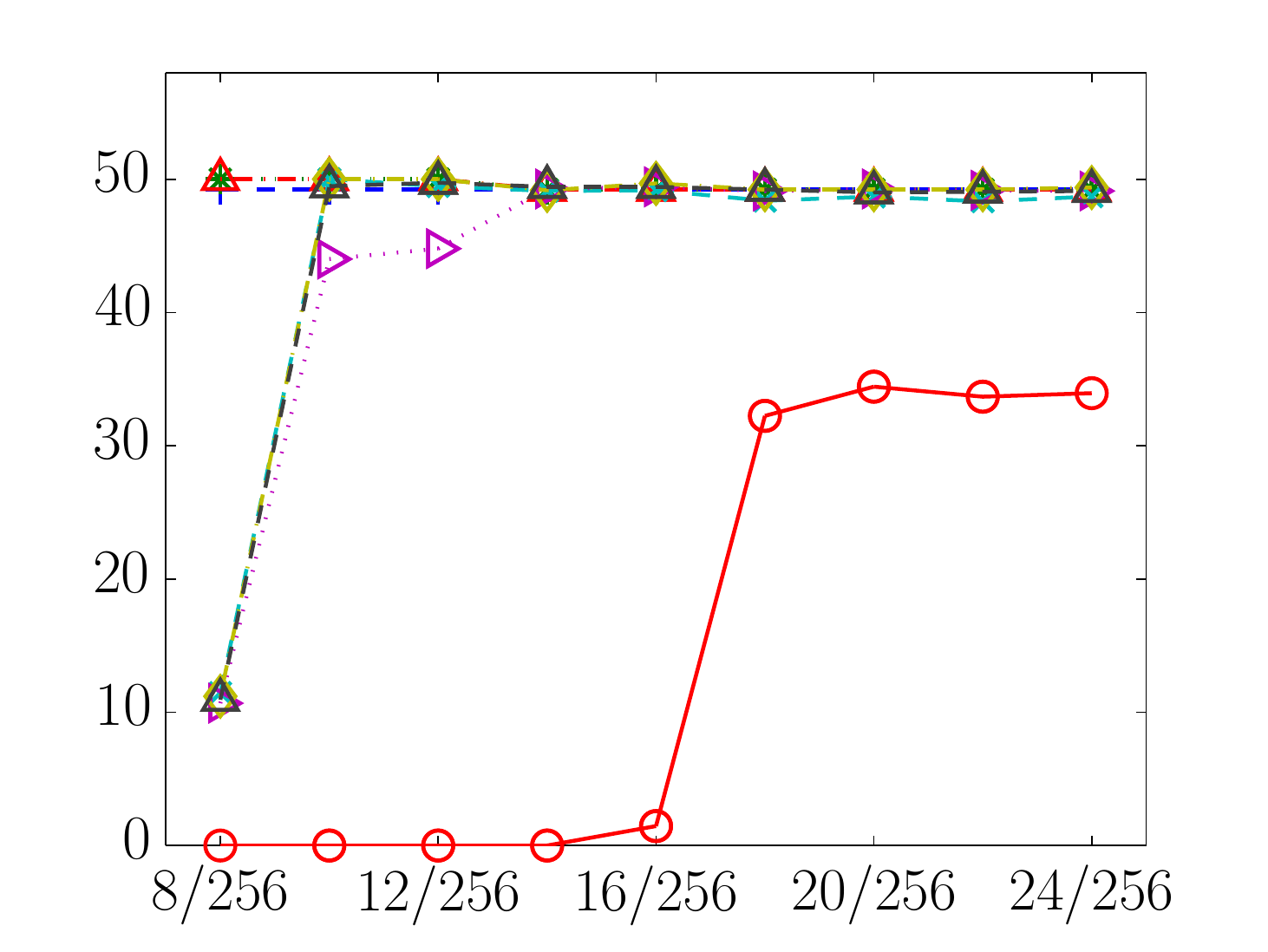}}
\begin{center}
\subfloat{\label{fig:sec:exp:syn:legend}\includegraphics[scale = 0.5]{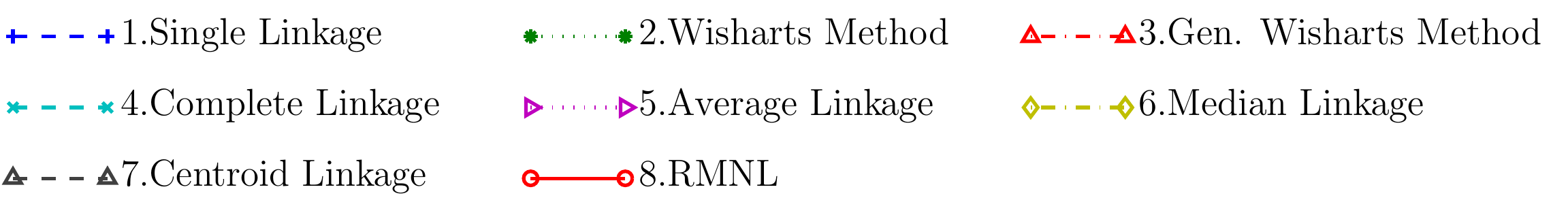}}
\end{center}
\caption{
{\small Classification Error on the synthetic data AIStat. The $y$-axis represents the \% error.
(a) Fix $\nu=0$ and vary $\alpha$ from $1/32$ to $1/16$. The $x$-axis represents the value of $\alpha$;
(b) Fix $\alpha=1/32$ and vary $\nu$ from $0$ to $1/32$. The $x$-axis represents the value of $\nu$;
(c) Vary $\alpha$ from $1/32$ to $1/16$, and vary $\nu$ from $0$ to $1/32$. The $x$-axis represents the value of $\alpha + \nu$.
Note that the instance no longer satisfies the weak good neighborhood property when $\alpha+\nu\geq 1/24 \approx 11/256$.
}
}
\label{fig:exp_syn}
\vspace*{\synVSpace}
\end{figure}

Here we compare the performance of the algorithms on the synthetic data AIStat.
Recall that the clustering $\{\textrm{AI}, \textrm{Statistics}\}$ satisfies the weak $(\alpha,\beta,\nu)$-good neighborhood property for $\alpha=1/32, \beta=7/8,\nu=0$ with high probability (See Figure~\ref{fig:weakProperty} in Section~\ref{sec:weak}). We conduct three sets of experiments, where we vary the values of $\alpha$ and $\nu$
by modifying the similarities between the points.
\begin{itemize}
\item[(a)] For each point $x$, we choose $\Delta\alpha n$ points $y$ from the other field and set the similarities $\simm(x,y)=\simm(y,x)=1$,
so that the value of $\alpha$ is increased to $1/32+\Delta\alpha$. By varying $\Delta\alpha$, we control $\alpha=1/32 + i/256$ for $i=0,\dots,8$ and run the clustering algorithms on the modified data.
\item[(b)] We randomly choose $\nu n$ points $x$, and then set the similarity between $x$ and any other point to be $1$ minus the original similarity.
This introduces $\nu n$ bad points. We thus control $\nu=i/256$ for $i=0,\dots,8$.
\item[(c)] We perform the above two modifications simultaneously, that is, we control $\alpha=1/32 + i/256$ and $\nu=i/256$ for $i=0,\dots,8$.
\end{itemize}
Note that the instance no longer satisfies the weak good neighborhood property when $\alpha+\nu\geq 1/24$.
This is because the weak good neighborhood requires that each point $p \not\in B$ falls into a subset $A_p$ of size greater than $6(\alpha+\nu)n$ with desired properties (see Property~\ref{prop:weak}), and the largest such subsets in AIStat have size $n/4$.

Figure~\ref{fig:exp_syn} shows the results of these experiments, averaged over $10$ runs.
When $\alpha+\nu< 1/24$, the instance satisfies the weak good neighborhood property and our algorithm has error at most $\nu$.
Moreover, even if the instance does not satisfy the weak good neighborhood property when $\alpha+\nu \geq 1/24$,
our algorithm still reports lower error.
All the other algorithms have higher error than our algorithm and fail rapidly as $\alpha+\nu$ increases.
This demonstrates that in cases modeled by the properties we propose, our algorithm will be successful while the
traditional agglomerative algorithms fail.

\subsection{Real-World Data}\label{sec:exp:real}

In this section, we compare the performance of our algorithm with the other algorithms on real-world data sets
and show that our algorithm consistently outperforms the others.

\subsubsection{Transductive Setting} \label{sec:exp:trans}

Here we compare the performance of the algorithms in the transductive setting where the algorithms use all the points in the data set.
Figure~\ref{fig:transductive} shows that our algorithm consistently achieves lowest or close to lowest errors on all the data sets.
Ward's Method is the best among the other algorithms, but still shows larger errors than our algorithms.
All the other algorithms generally show worse performance, and report significantly higher errors on some of the data sets.
The comparison shows the robustness of our algorithm to the noise in the real world data sets.

To further evaluate the robustness of the algorithms, in the following
we show their performance when different types of noise are added to the data.
Since our algorithm requires additional parameters to characterize noise,
we also discuss their robustness to parameter tuning.

\newcommand{\transFigScale}{0.32}
\begin{figure}[!t]
    \hspace*{-.45in}
    \subfloat[Wine]{\label{fig:wine}\includegraphics[scale = \transFigScale]{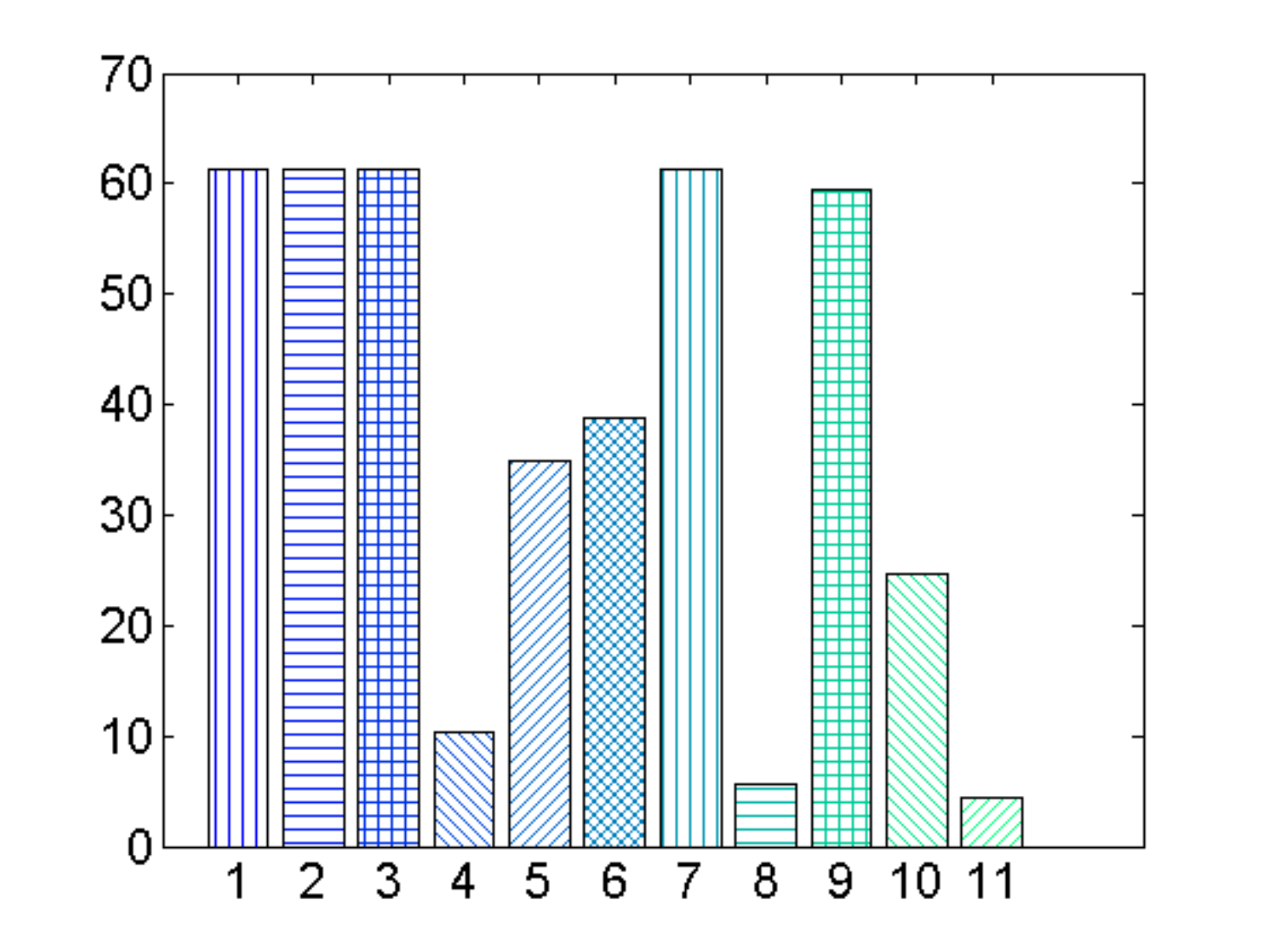} }
    \hspace*{-.25in}
    \subfloat[Iris]{\label{fig:iris}\includegraphics[scale = \transFigScale]{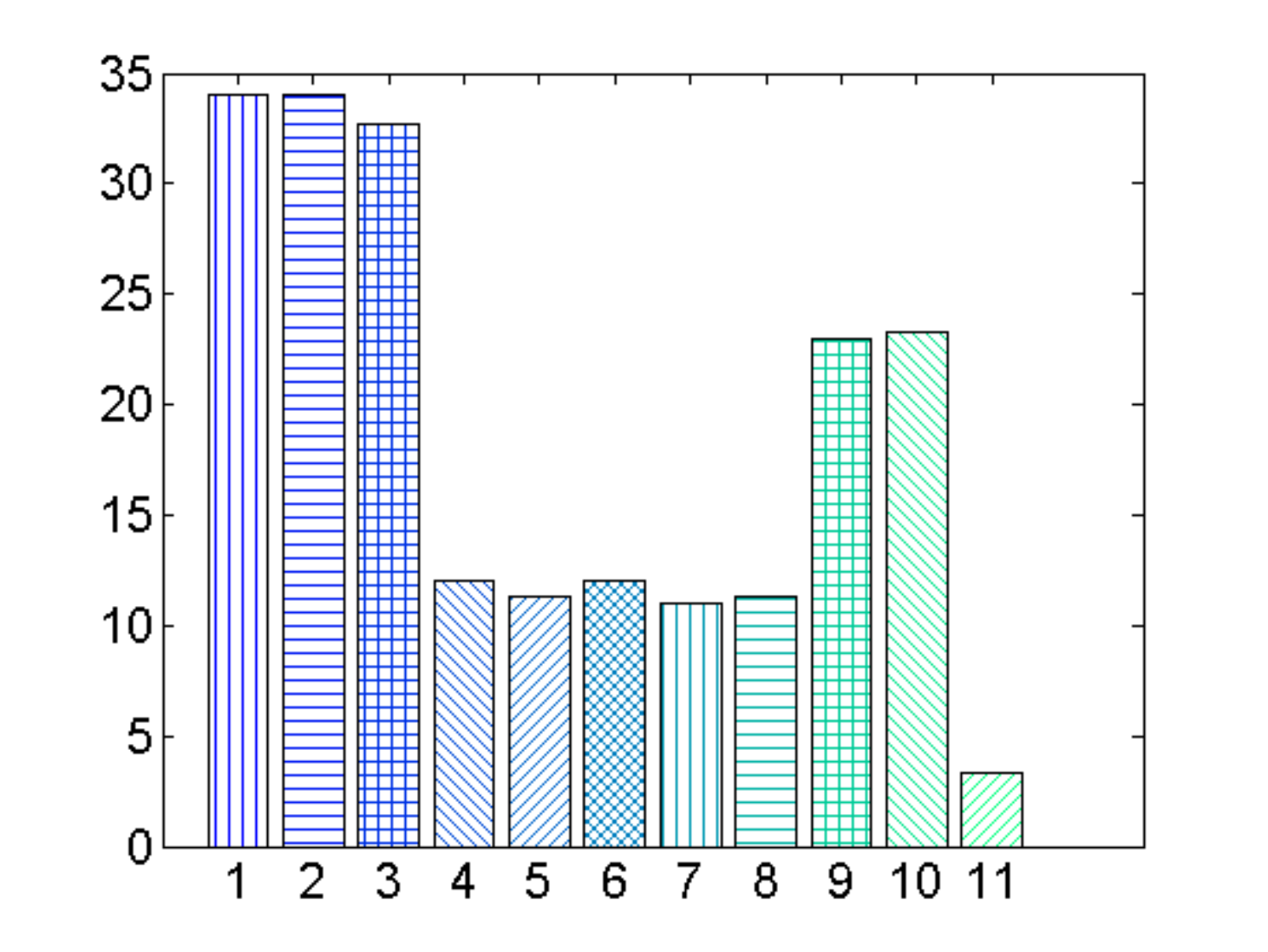}  }
    \hspace*{-.25in}
    \subfloat[BCW]{\label{fig:bcw}\includegraphics[scale=\transFigScale]{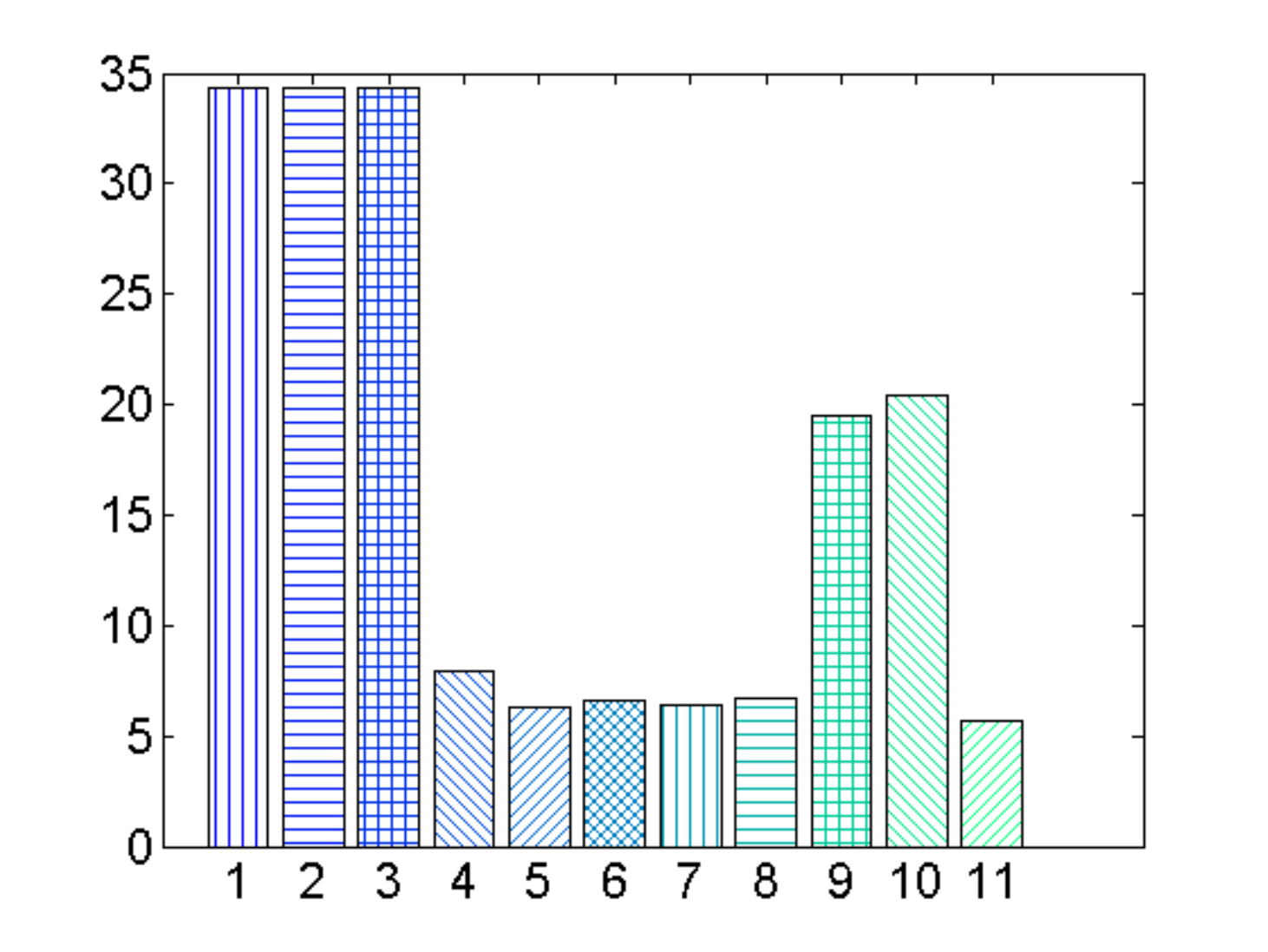}}
    \hspace*{-.2in}
    \subfloat[BCWD]{\label{fig:bcwd}\includegraphics[scale=\transFigScale]{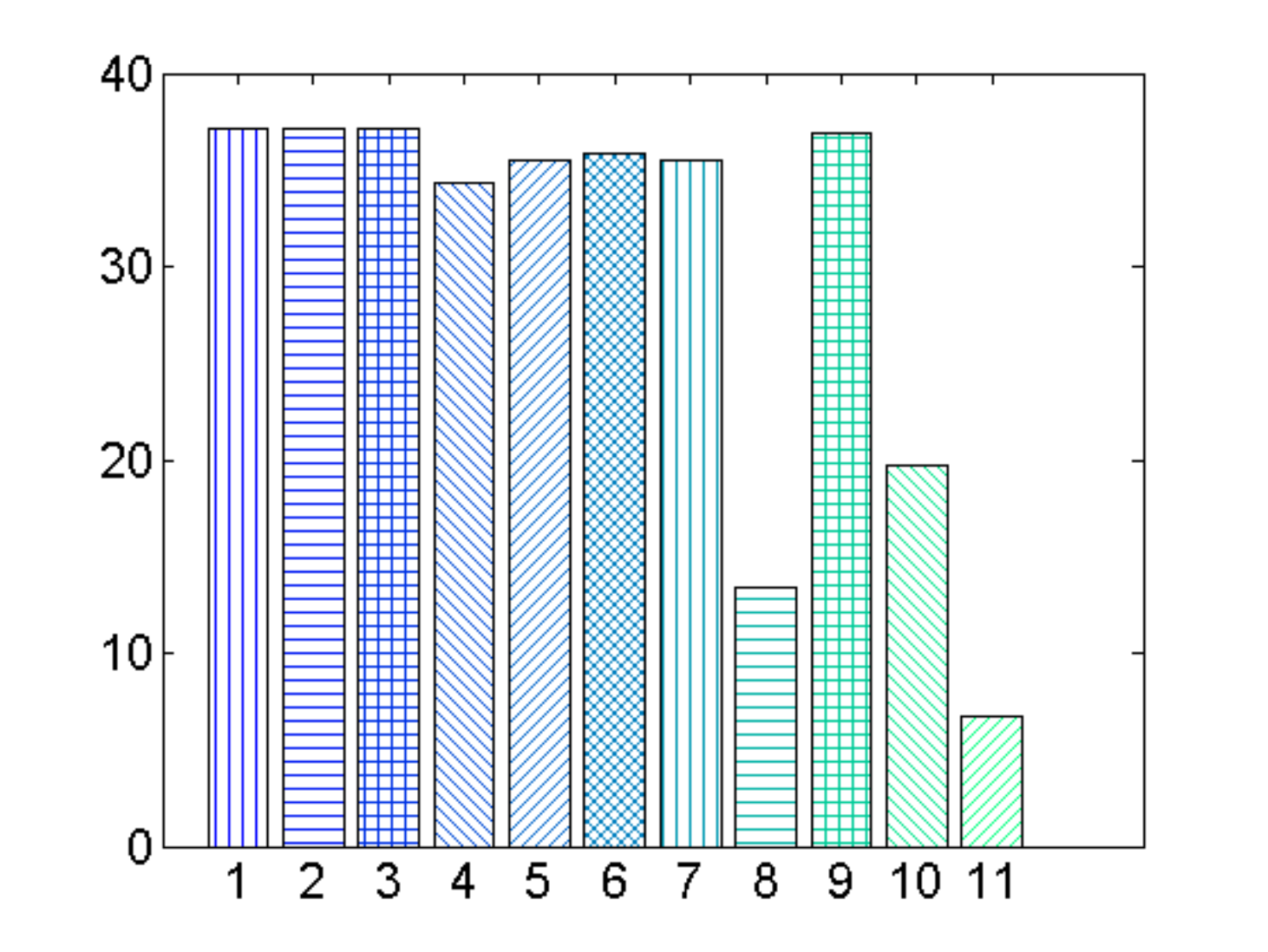}}
\begin{center}
\centering
\subfloat{\label{fig:sec:exp:trans:legend}\includegraphics[scale = 0.45]{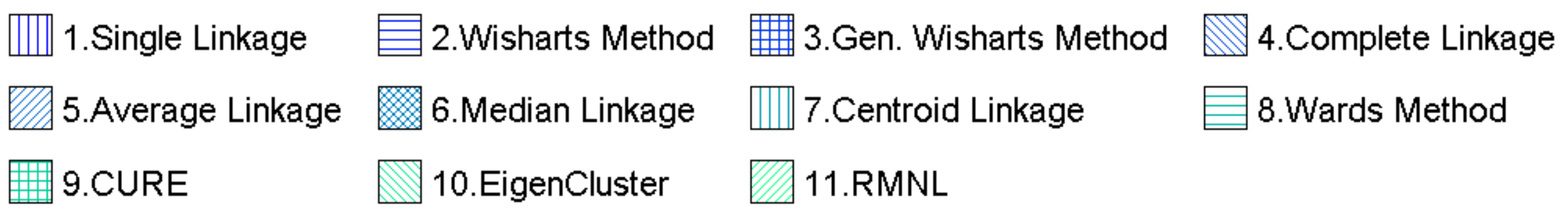}}
\end{center}
\vspace*{-.1in}
\caption{{\small Classification Error in the transductive setting. The $y$-axis represents the \% error.}}
\label{fig:transductive}
\end{figure}

\paragraph{Robustness to Noise}
Here we present the performance of the algorithms when Gaussian noise or corruption noise is added and the level of noise is increased monotonically.
The Gaussian noise model essentially corresponds to additive perturbations to the data entries and it is a very common type of noise studied throughout machine learning.
The corruption noise models data corruption or missing values,
and is also frequently studied in machine learning and coding theory~\citep{blum2007separating,feldman2008learning,wigderson2012population,moitra2013polynomial}.
The experiments on different types of noise then evaluate the robustness of the algorithms to
noise caused by different reasons in real world scenarios.
Note that the instance is not in a metric space after adding noise to the similarities,
so in this case, we only evaluate algorithms that can be run on non-metric instances.

\newcommand{\noiseFigScale}{0.3}
\newcommand{\noiseHSpace}{-0.2in}
\newcommand{\noiseHMargin}{-.5in}
\newcommand{\noiseVSpace}{-0.2in}
\newcommand{\noiseLegendScale}{0.5}

\begin{figure}[!t]
    \hspace*{\noiseHMargin}
    \begin{sideways}
    \small corrupted attributes
    \end{sideways}
    \subfloat{\includegraphics[scale = \noiseFigScale]{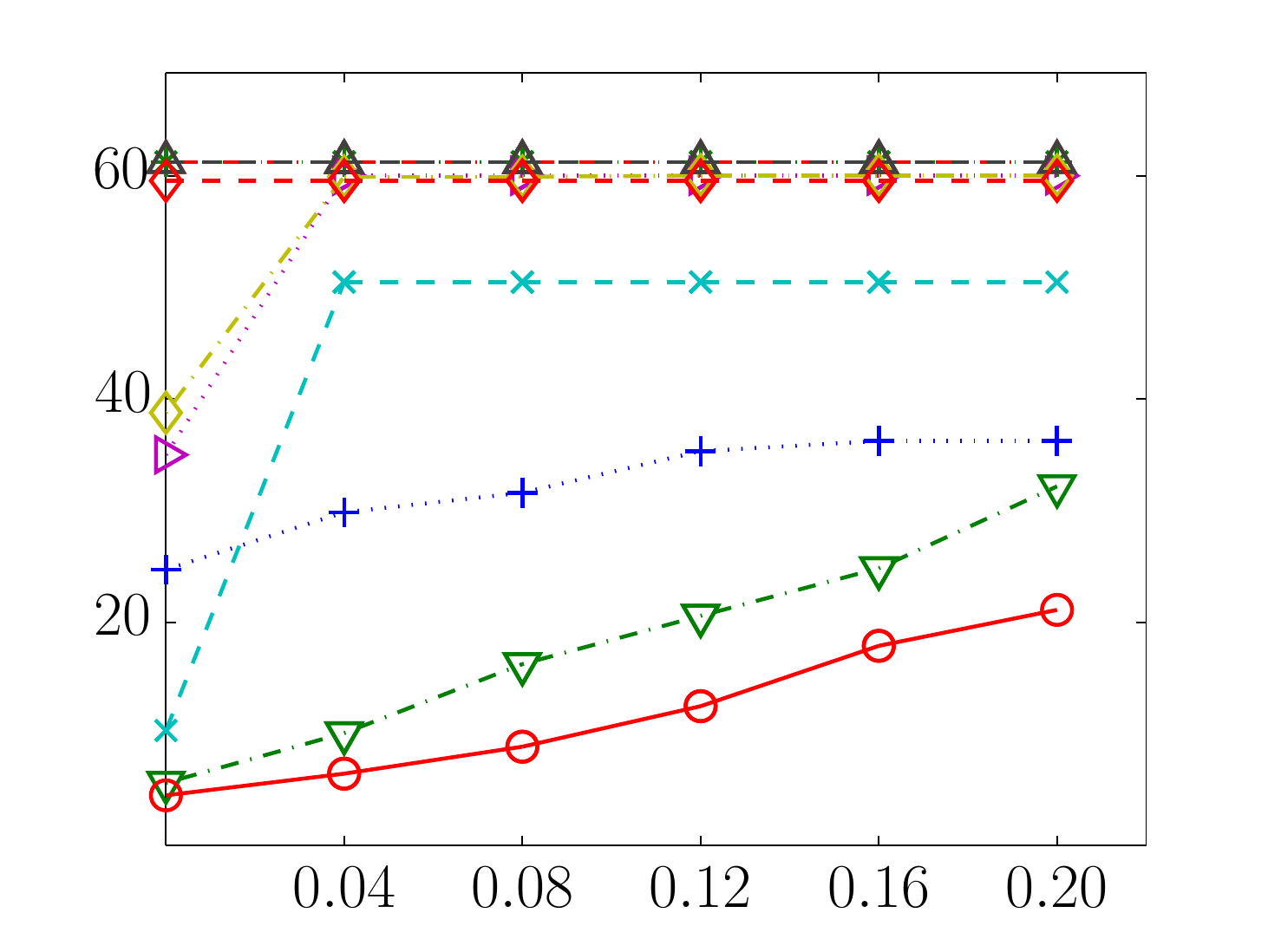} }
    \hspace*{\noiseHSpace}
    \subfloat{\includegraphics[scale = \noiseFigScale]{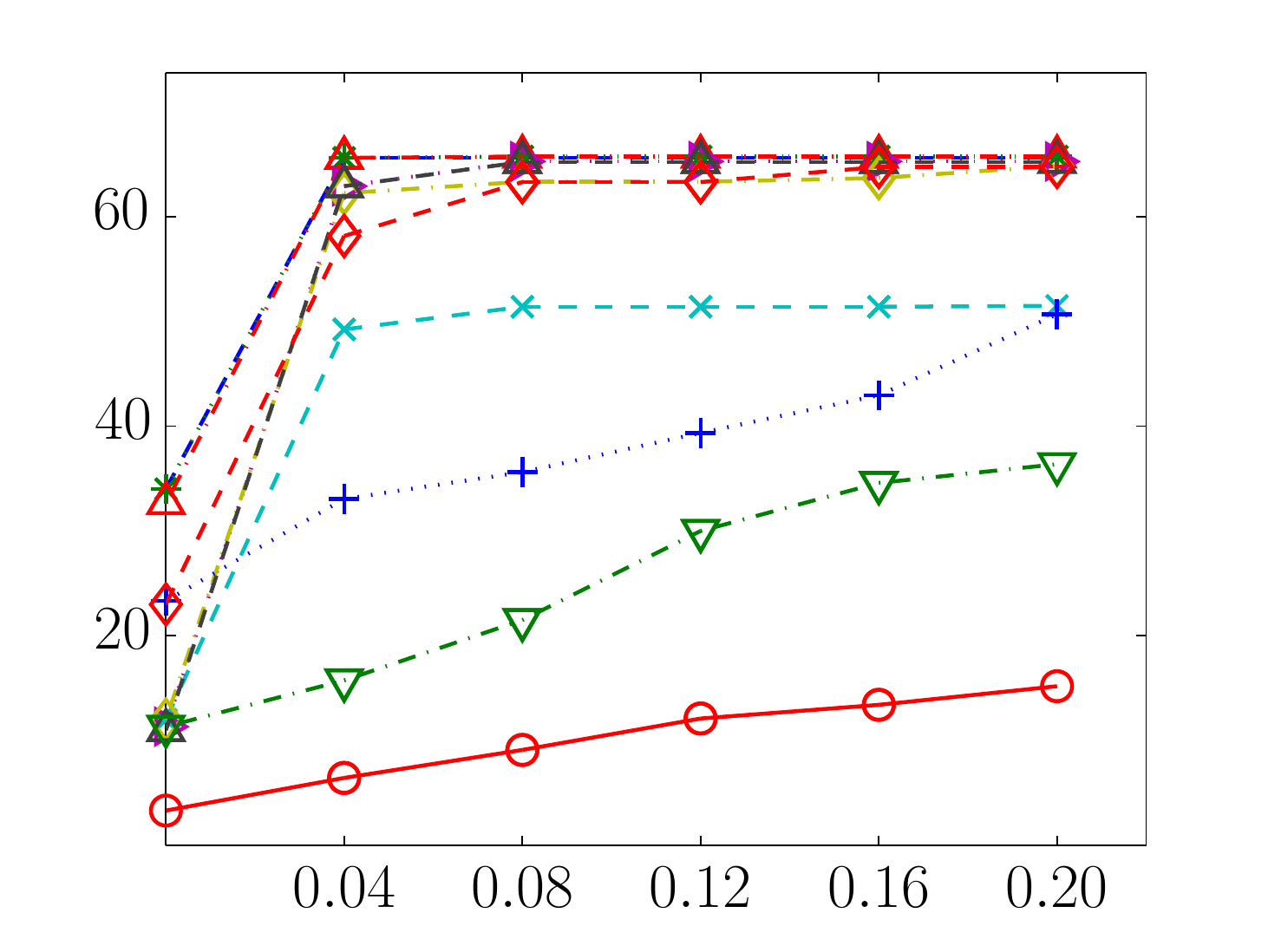}  }
    \hspace*{\noiseHSpace}
    \subfloat{\includegraphics[scale = \noiseFigScale]{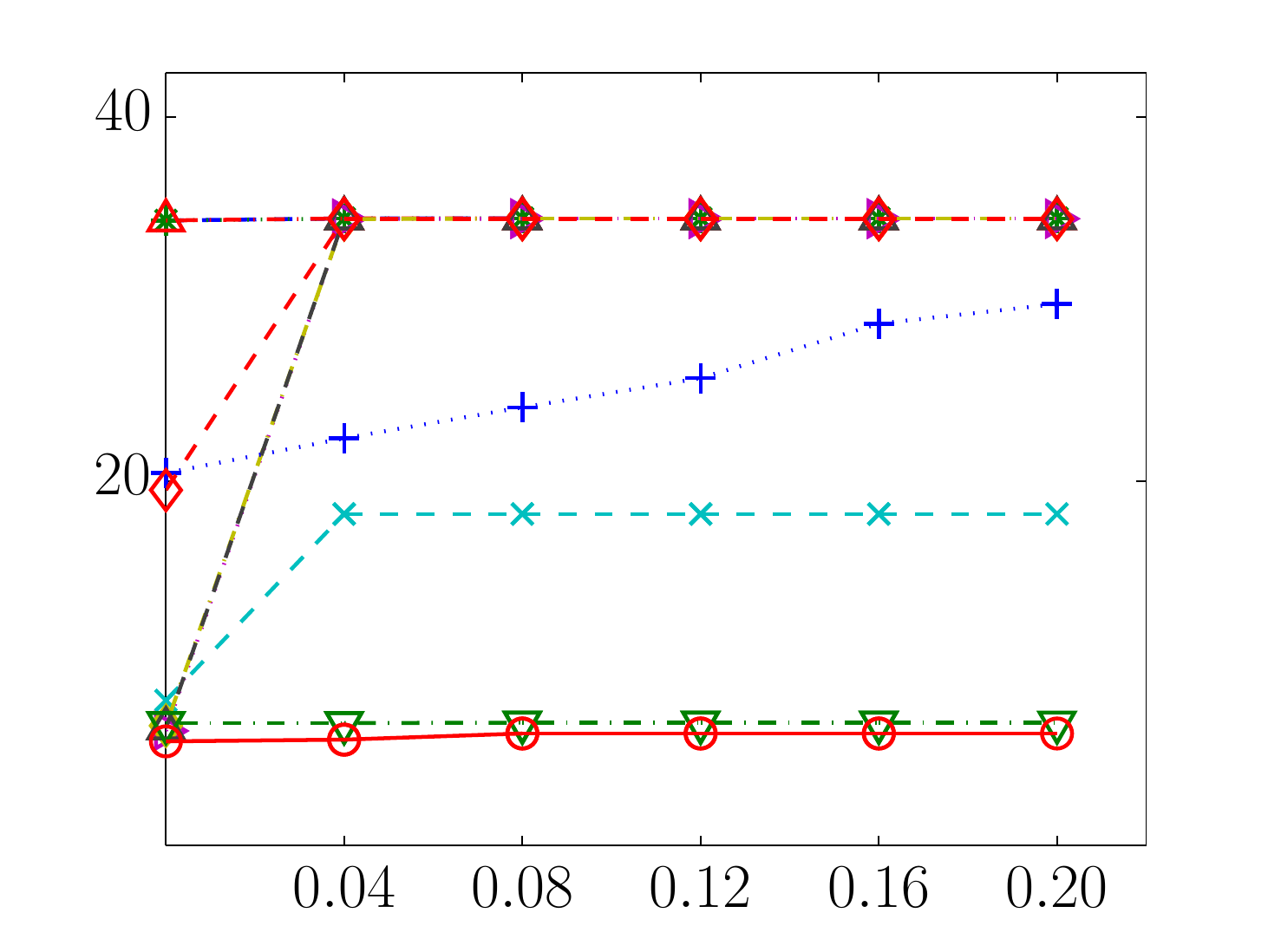} }
    \hspace*{\noiseHSpace}
    \subfloat{\includegraphics[scale = \noiseFigScale]{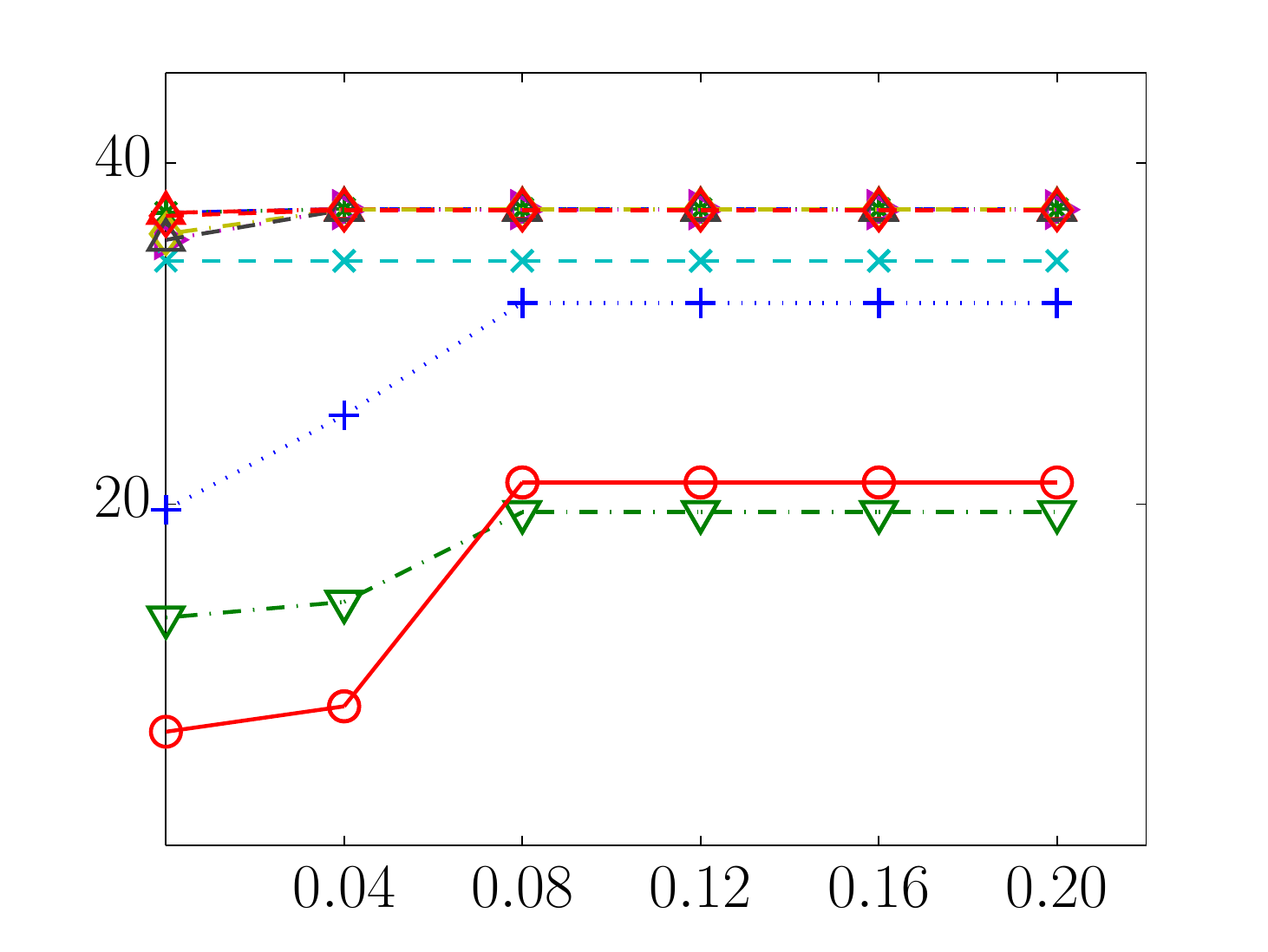}  }

    \hspace*{\noiseHMargin}
    \begin{sideways}
    \small corrupted similarities
    \end{sideways}
    \subfloat{\includegraphics[scale = \noiseFigScale]{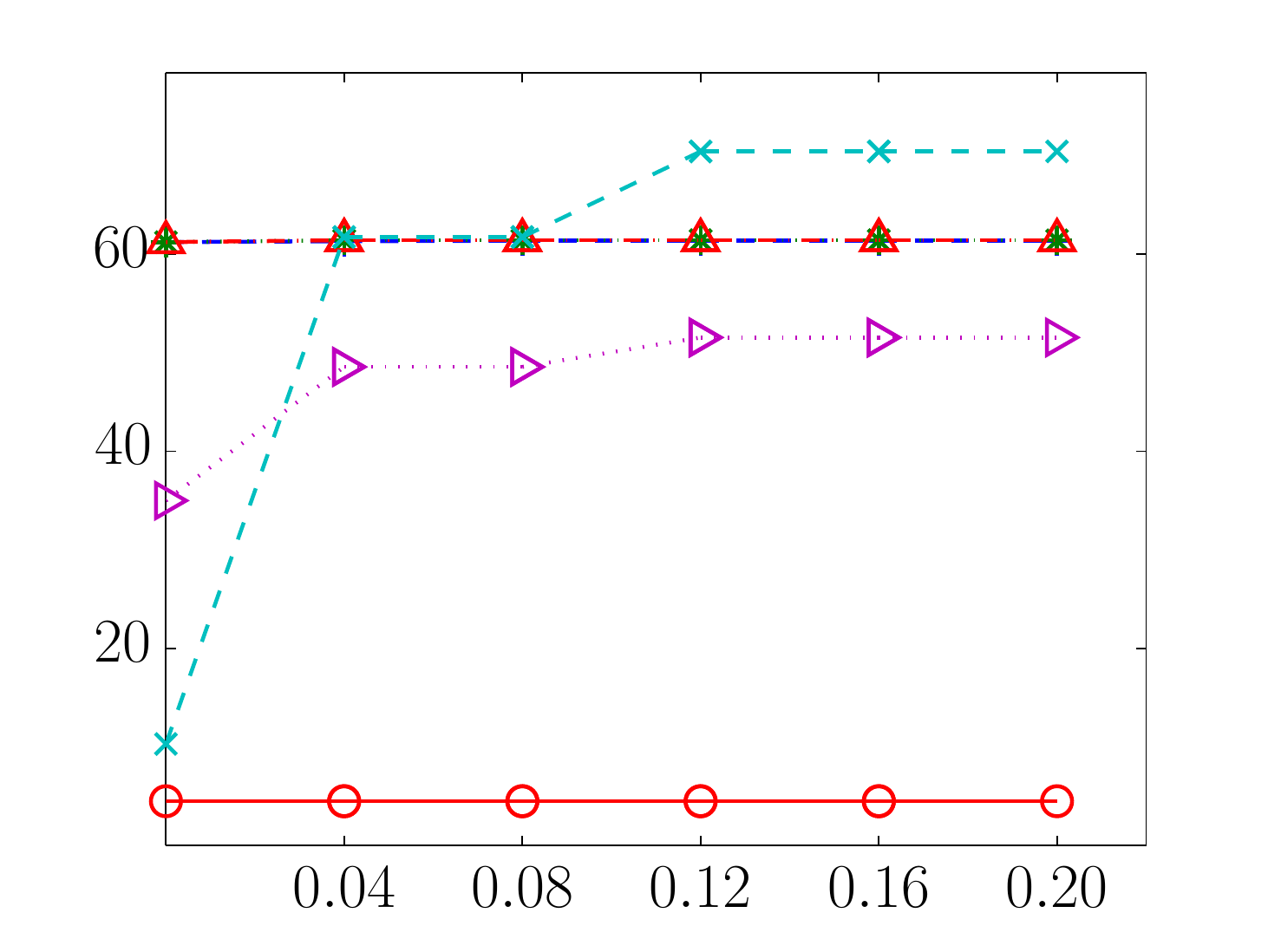} }
    \hspace*{\noiseHSpace}
    \subfloat{\includegraphics[scale = \noiseFigScale]{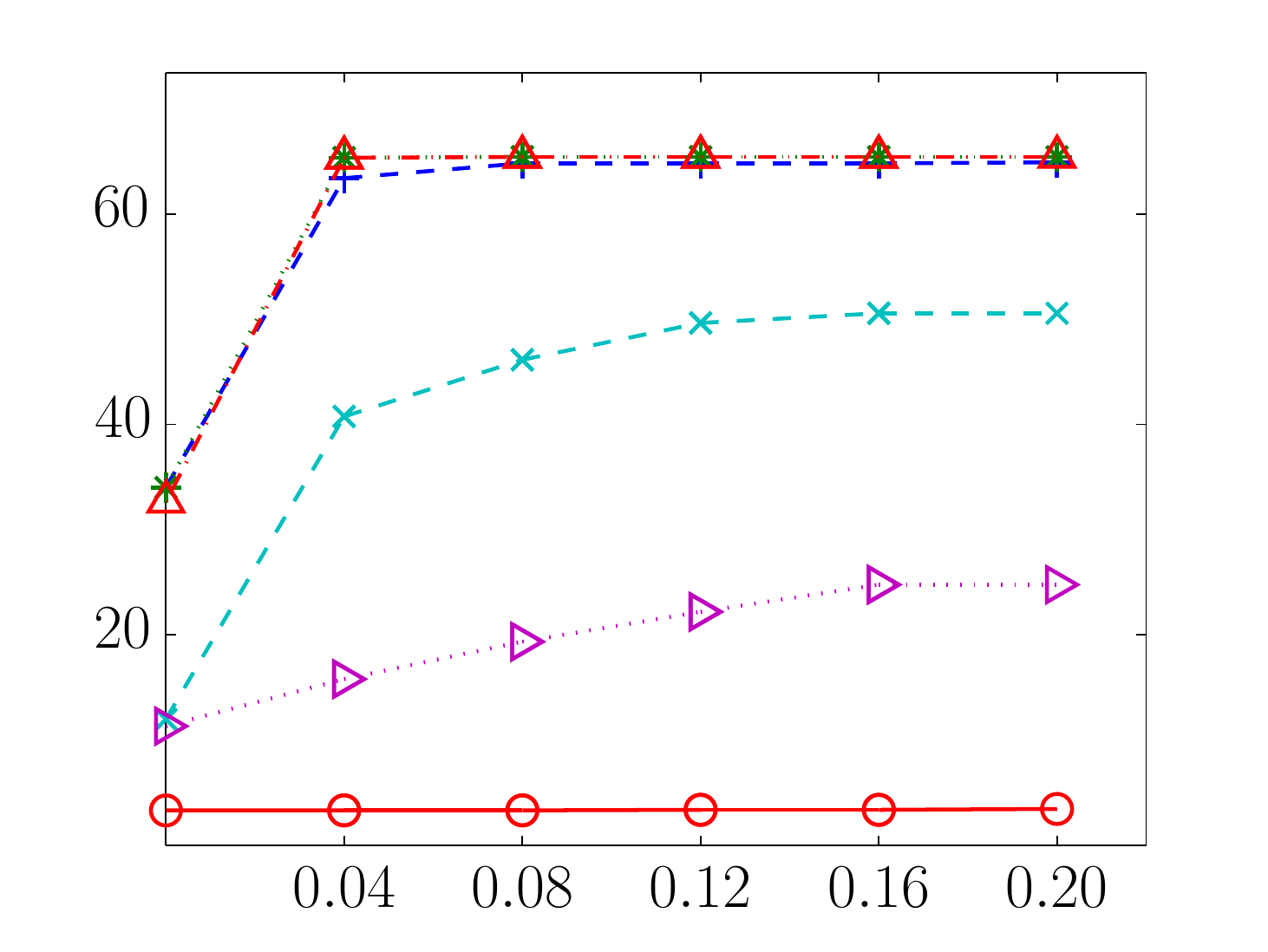}  }
    \hspace*{\noiseHSpace}
    \subfloat{\includegraphics[scale = \noiseFigScale]{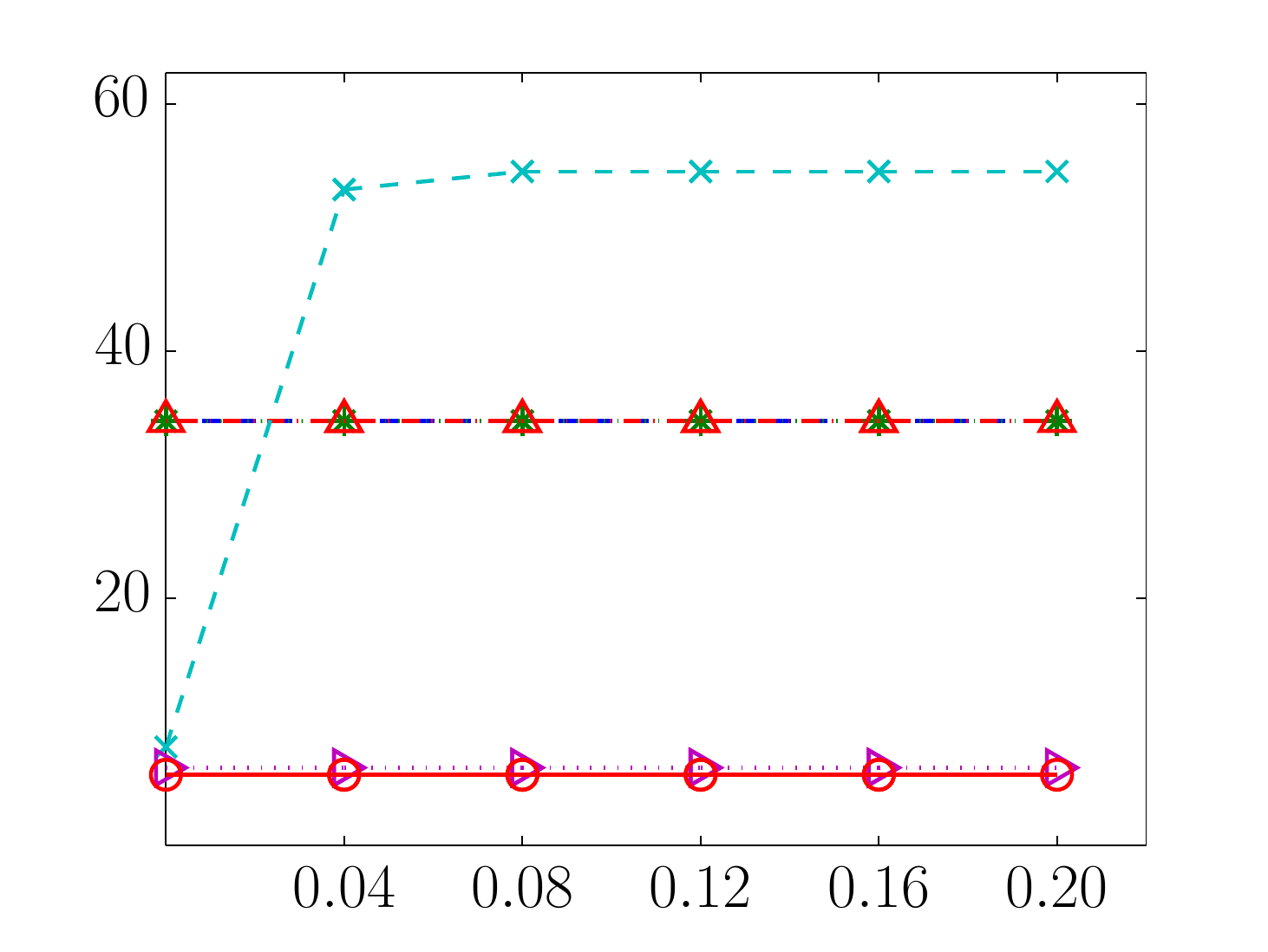} }
    \hspace*{\noiseHSpace}
    \subfloat{\includegraphics[scale = \noiseFigScale]{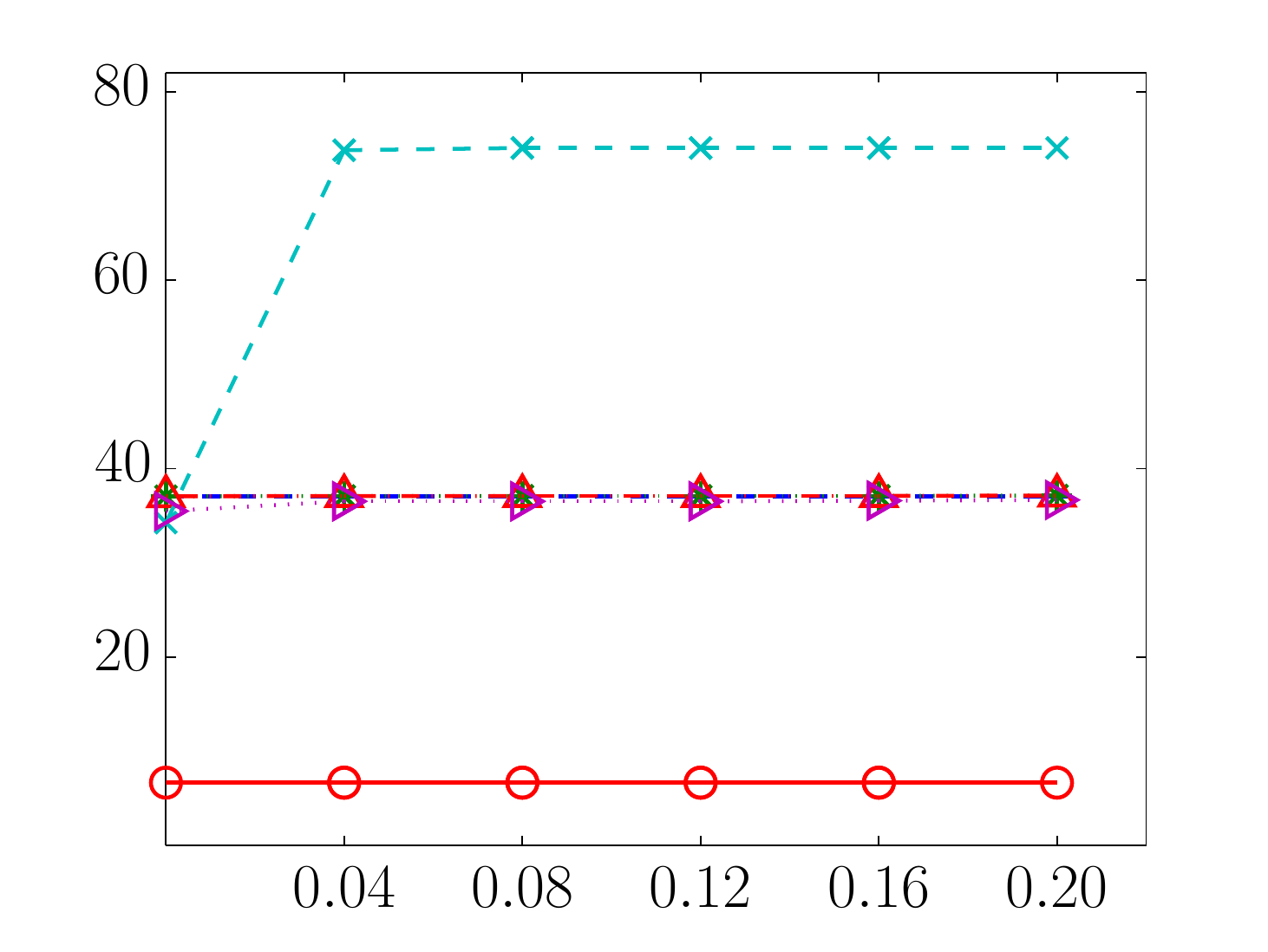}  }

    \hspace*{\noiseHMargin}
    \begin{sideways}
    \small Gaussian attributes
    \end{sideways}
    \subfloat{\includegraphics[scale = \noiseFigScale]{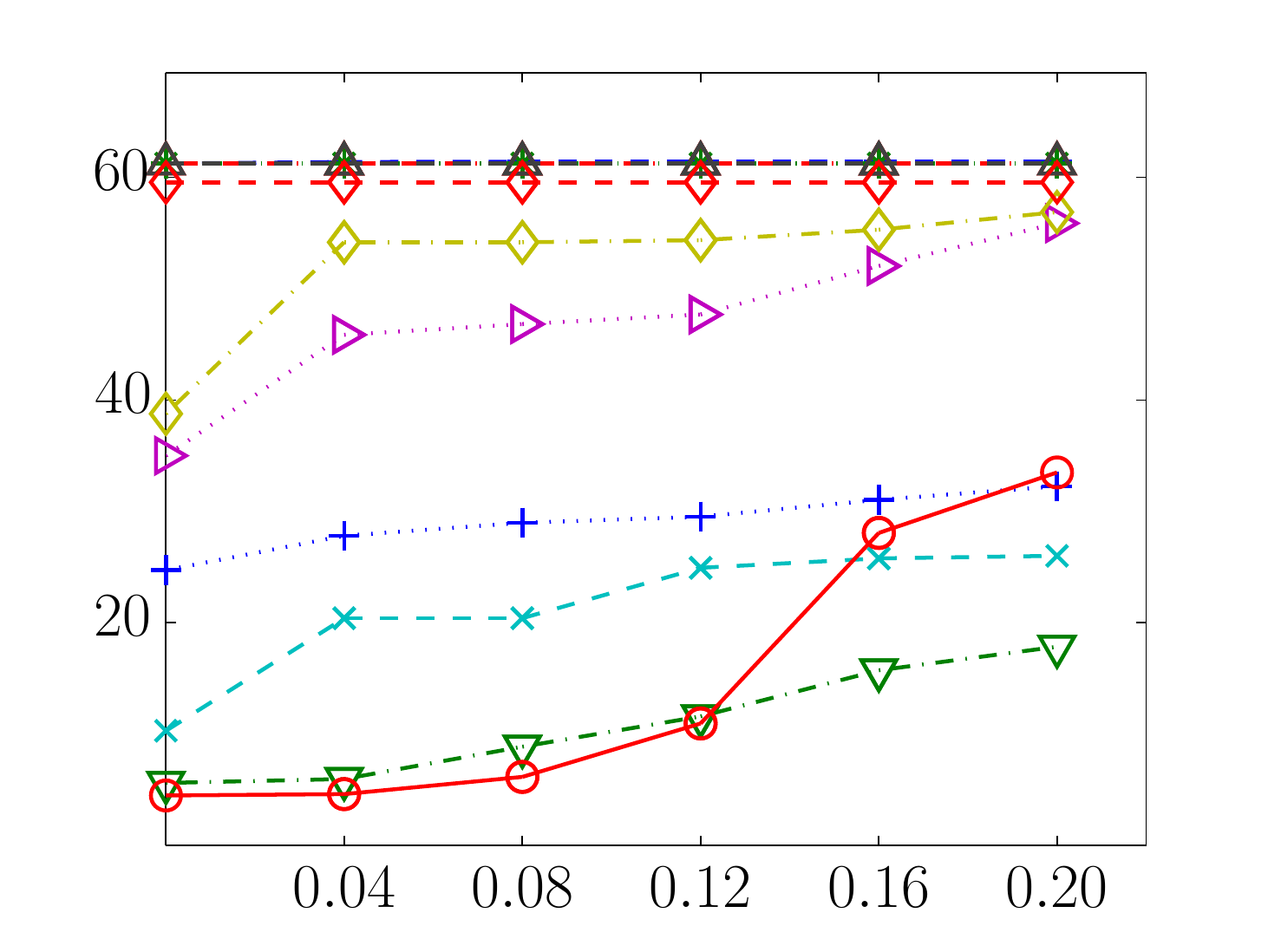} }
    \hspace*{\noiseHSpace}
    \subfloat{\includegraphics[scale = \noiseFigScale]{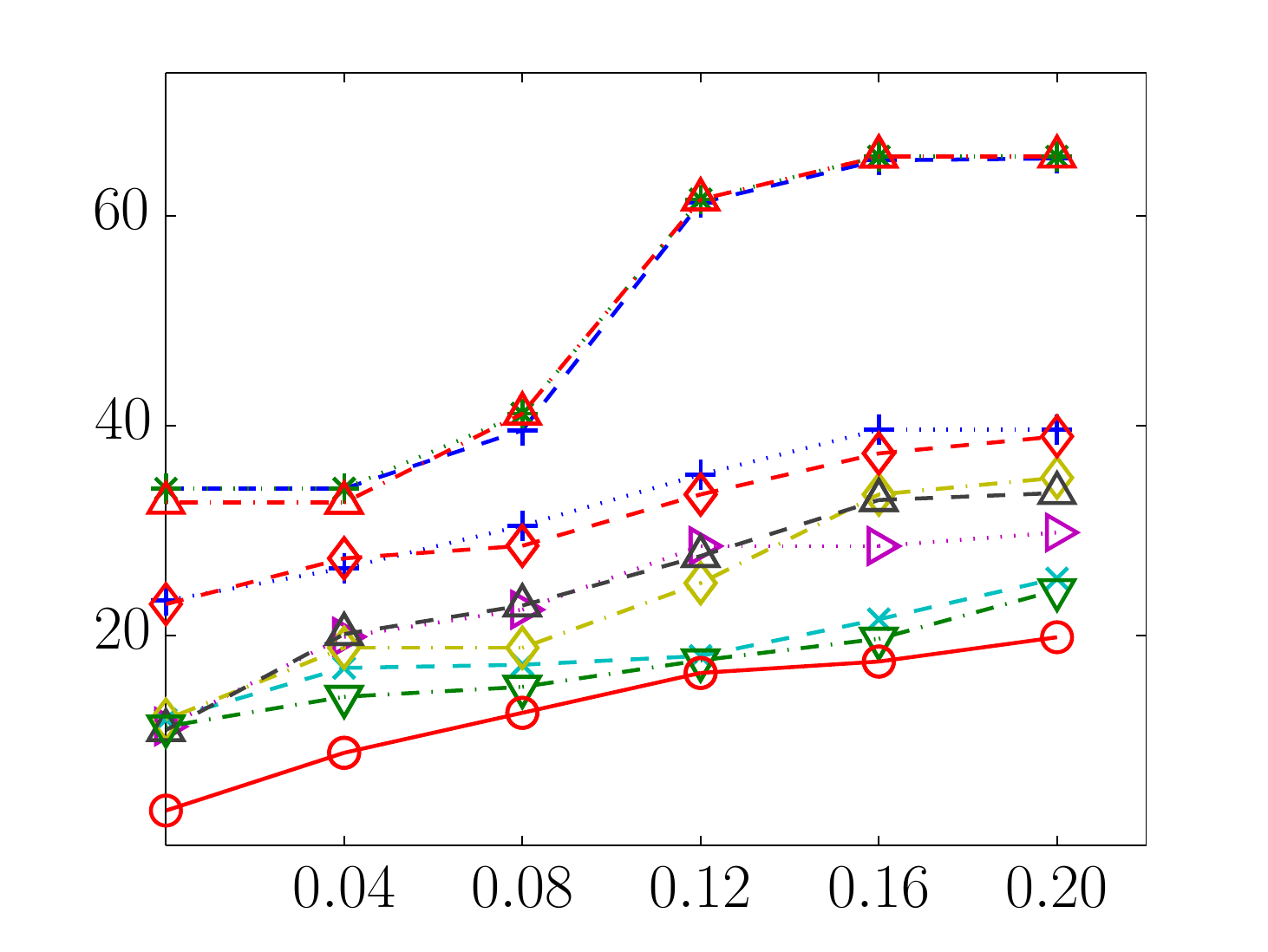}  }
    \hspace*{\noiseHSpace}
    \subfloat{\includegraphics[scale = \noiseFigScale]{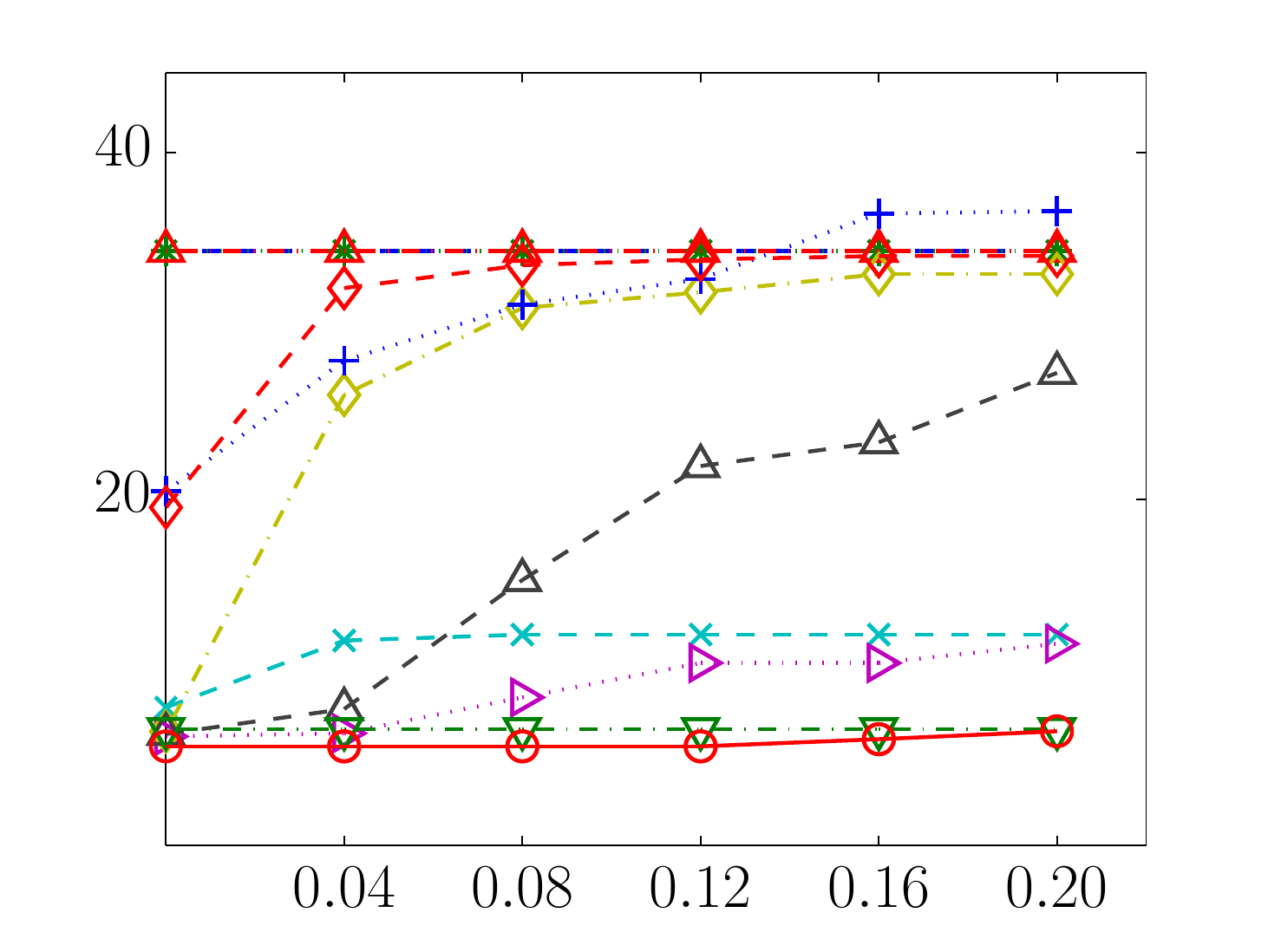} }
    \hspace*{\noiseHSpace}
    \subfloat{\includegraphics[scale = \noiseFigScale]{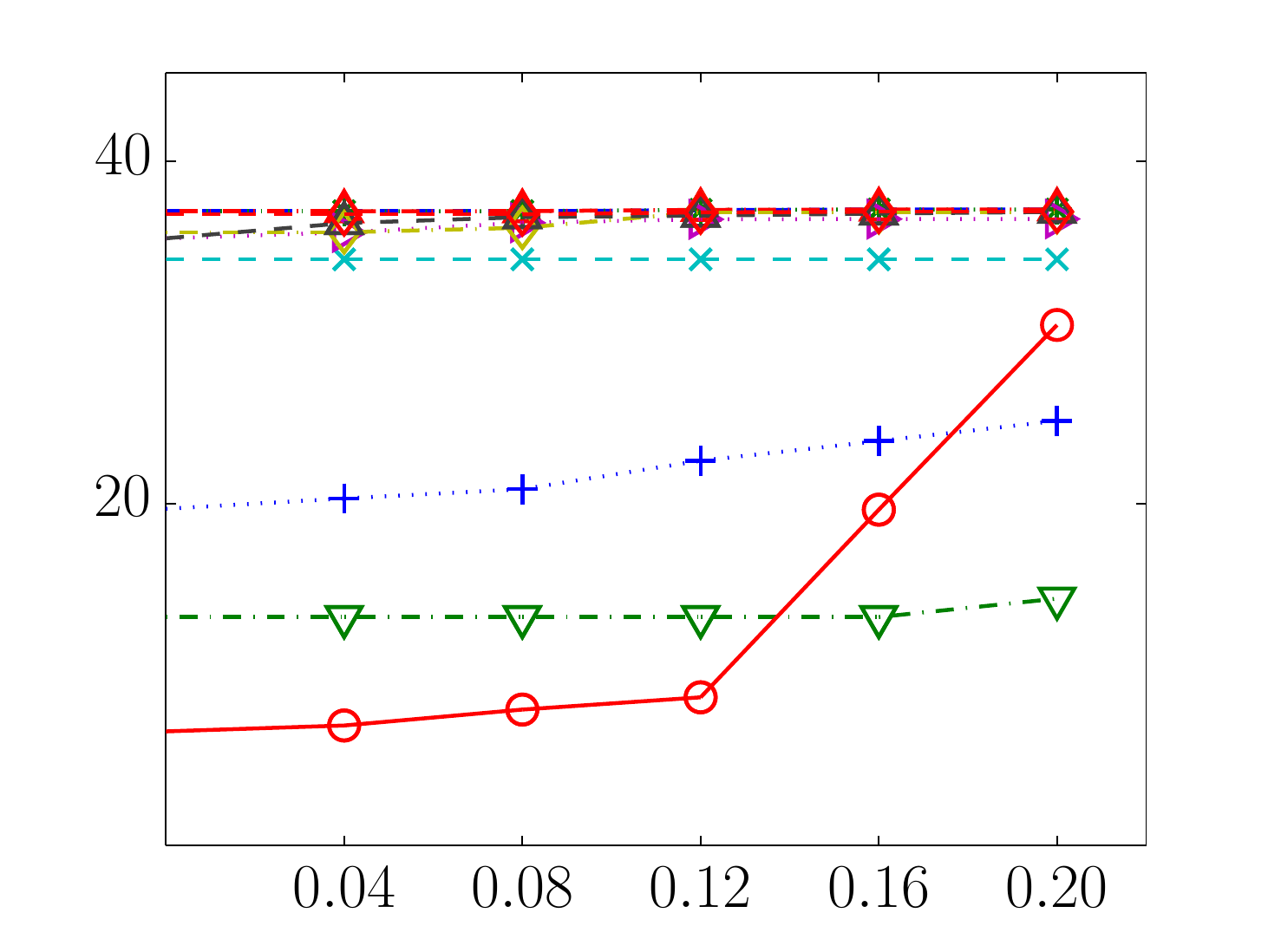}  }

    \hspace*{\noiseHMargin}
    \subfloat{\hspace*{0.8in} Wine \hspace*{.8in}}
    \hspace*{\noiseHSpace}
    \subfloat{ \hspace*{.625in} Iris \hspace*{.625in}}
    \hspace*{\noiseHSpace}
    \subfloat{ \hspace*{.725in} BCW \hspace*{.625in} }
    \hspace*{\noiseHSpace}
    \subfloat{ \hspace*{.625in} BCWD \hspace*{.625in} }

\begin{center}
\subfloat{\includegraphics[scale = \noiseLegendScale]{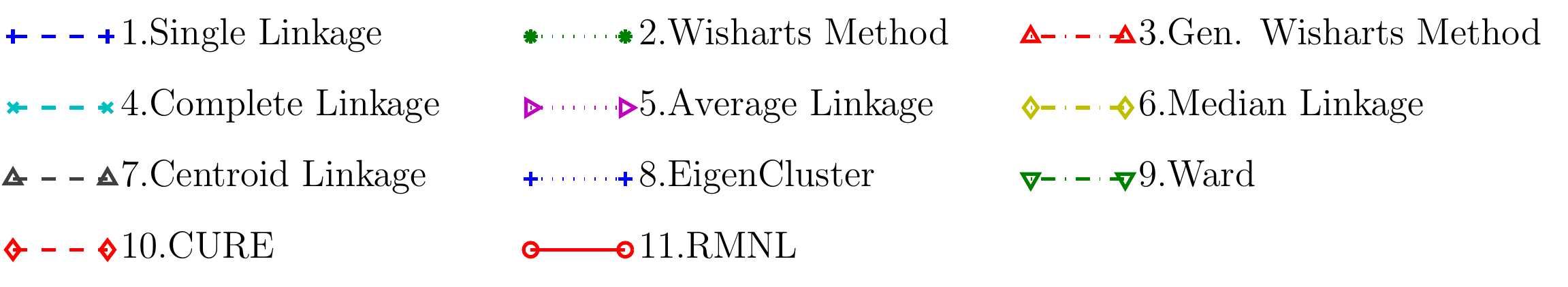}}
\end{center}
\vspace*{\noiseVSpace}
\caption{
{\small Classification Error in the presence of noise. Rows: corruption noise added to the attributes, corruption noise added to the similarities,
Gaussian noise added to the attributes. Columns: Wine, Iris, BCW, BCWD.
In each subfigure, the $x$-axis represents the noise level, and the $y$-axis represents the \% error.
}
}
\label{fig:noise}
\end{figure}

We consider three types of noise: corruption noise to the attributes,
corruption noise to the similarities, and Gaussian noise added to the attributes.
The first type of noise is generated as follows: normalize the entries in the data matrix to $[0,1]$;
randomly pick $p$ fraction of the entries;
replace each sampled entry with a random value independently generated from $N(0,1)$,
where $p$ is the parameter indicating the level of noise.
The second type of noise is generated using the same approach, but is added to the similarity matrix.
The third type of noise is generated as follows:
normalize the entries in the data matrix to $[0,1]$;
add a random value independently generated from $N(0,p^2)$ to each entry,
where $p$ is the parameter indicating the level of noise.

Figure~\ref{fig:noise} shows the results of different algorithms in the presence of noise, averaged over $30$ runs.
The rows correspond to different types of noise added, and the columns correspond to different data sets.
The first row shows the results when corruption noise is added to the attributes.
Our algorithm shows robustness to such type of noise:
its error rates remain the best or close to the best up to noise level $0.2$
on all data sets.
EigenCluster and Ward's method also show robustness, but their error rates are generally higher than those of our algorithm.
The other algorithms report high errors even when the noise level is as low as $0.04$.

The second row shows the results when corruption noise is added to the similarities.
We observe that the errors of our algorithm remain nearly the same up to noise level $0.2$ over all the data sets,
while the other algorithms report higher errors.
Some algorithms (such as Complete Linkage on Wine) show comparable performance to our algorithms when there is no noise,
but their errors generally increase rapidly as the noise level increases.
This shows that our algorithm performs much better than the other algorithms in the presence of corruptions in the similarities.

The third row shows the results when Gaussian noise is added to the attributes.
We observe that when the noise level increases, the errors of all algorithms increase.
The errors of our algorithm do not increase much:
they remain the best or close to the best up to the noise level $0.2$ on all the data sets.
Ward's method also shows robustness, since the minimum variance criterion used is insensitive to this type of noise.
The other algorithms generally show higher errors than our algorithms and Ward's method.

In conclusion, our algorithm generally outperforms the other algorithms when corruption noise is added to the data attributes or the similarities,
or when Gaussian noise is added to the data attributes. Its robustness to Gaussian noise in similarities is not so significant since such noise
with large variance can change the neighbor rankings of all points considerably.
Still, it can tolerate such noise when the noise variance is not too large.

\newcommand{\alphaFigScale}{0.3}
\newcommand{\alphaLenScale}{0.6}
\newcommand{\alphaHSpace}{-0.1in}
\newcommand{\alphaHMargin}{-.4in}
\newcommand{\alphaVSpace}{-0.2in}

\begin{figure}[!t]
    \hspace*{\alphaHMargin}
    \subfloat[Wine]{\label{fig:alpha_wine}\includegraphics[scale = \alphaFigScale]{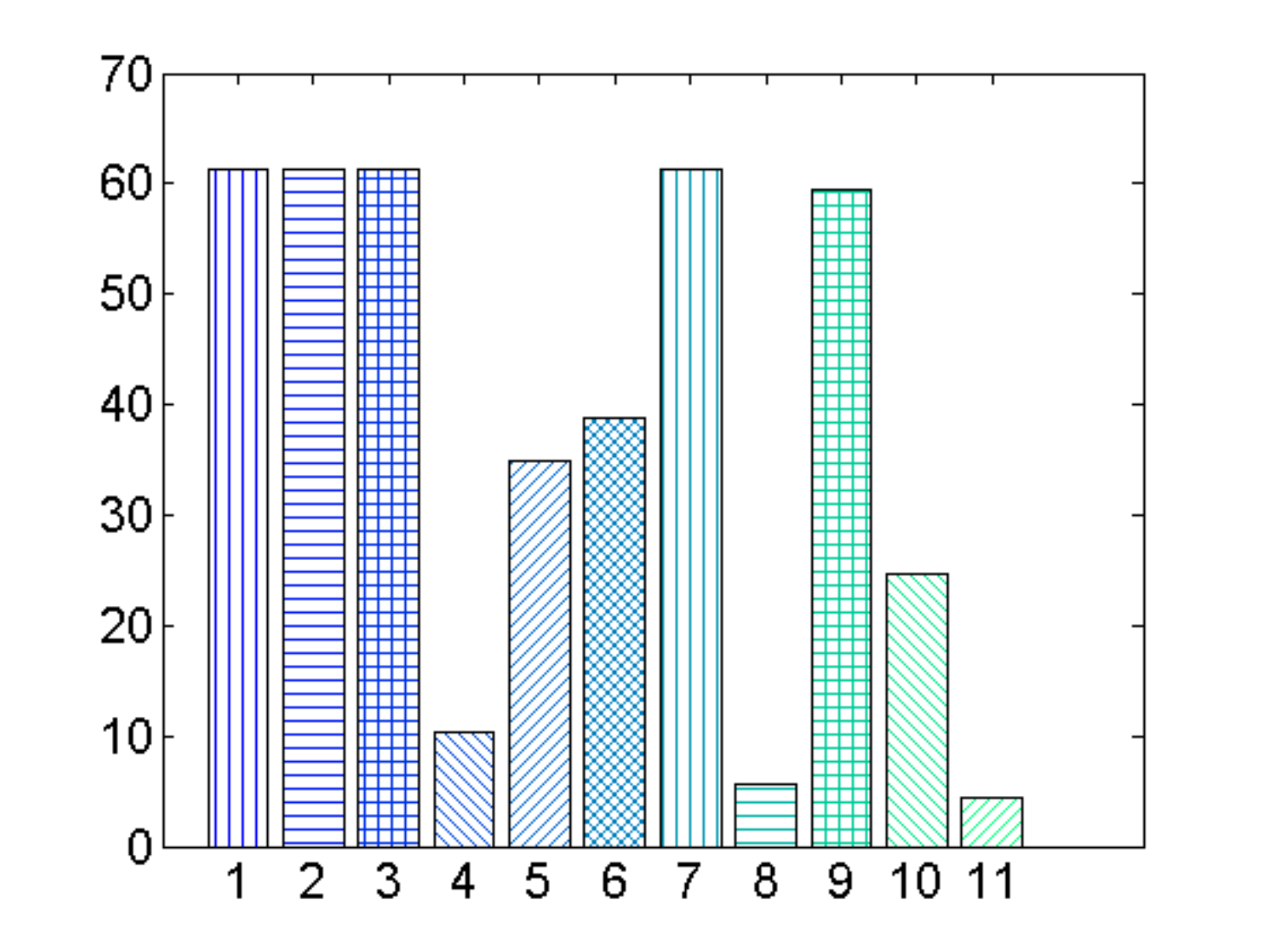} }
    \hspace*{\alphaHSpace}
    \subfloat[Iris]{\label{fig:alpha_iris}\includegraphics[scale = \alphaFigScale]{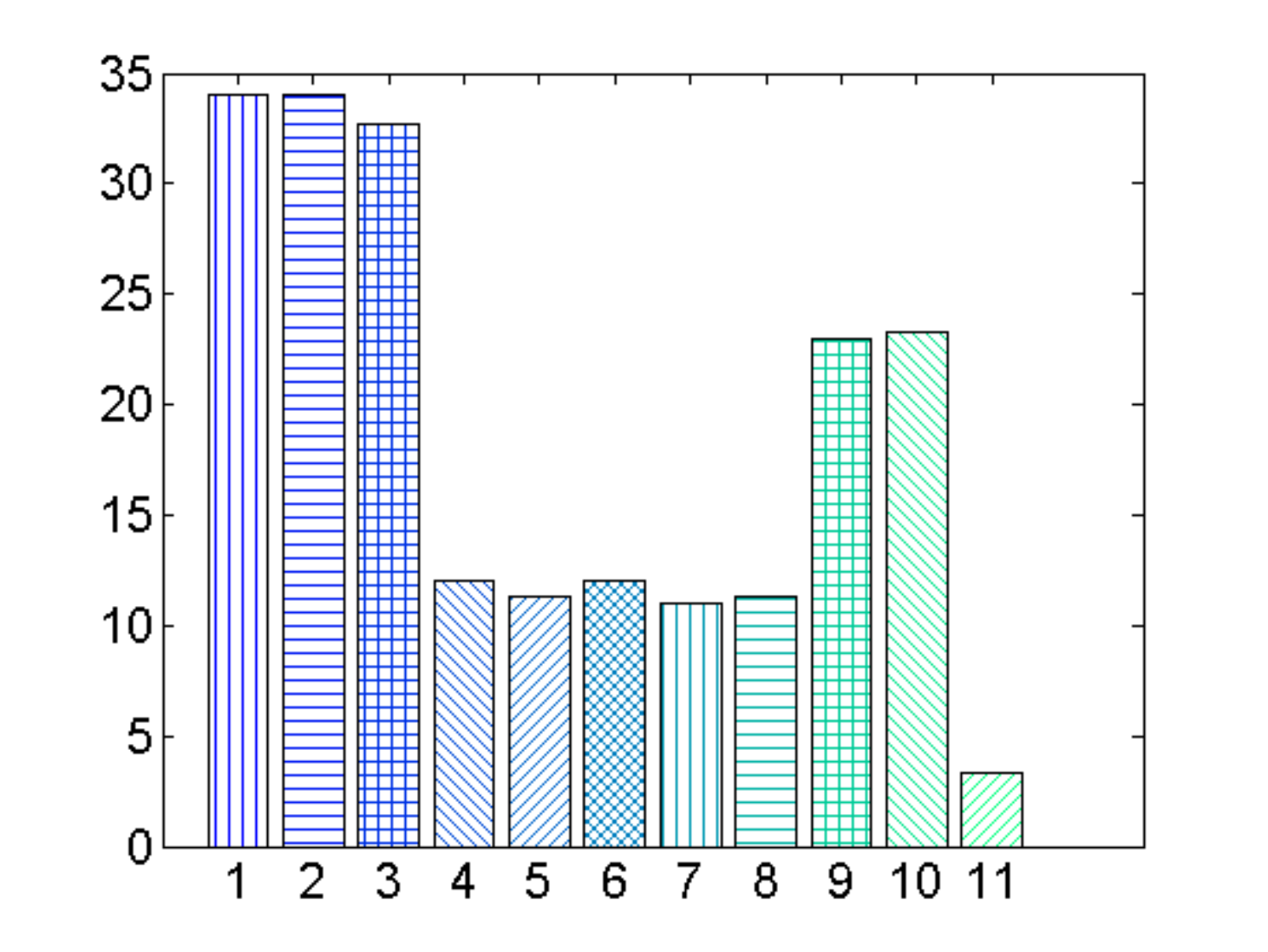}  }
    \hspace*{\alphaHSpace}
    \subfloat[BCW]{\label{fig:alpha_bcw}\includegraphics[scale=\alphaFigScale]{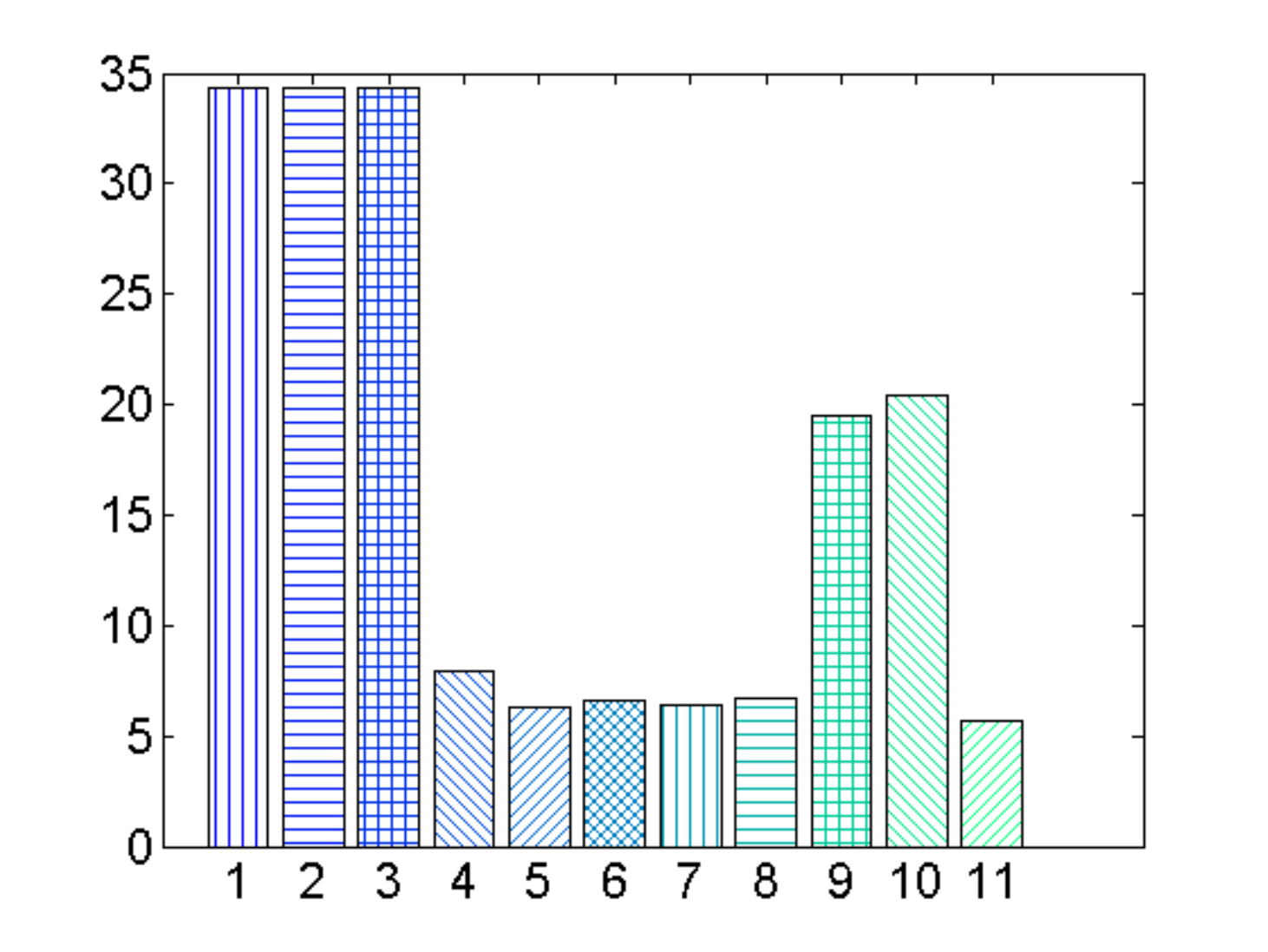}}
    \hspace*{\alphaHSpace}
    \subfloat[BCWD]{\label{fig:alpha_bcwd}\includegraphics[scale=\alphaFigScale]{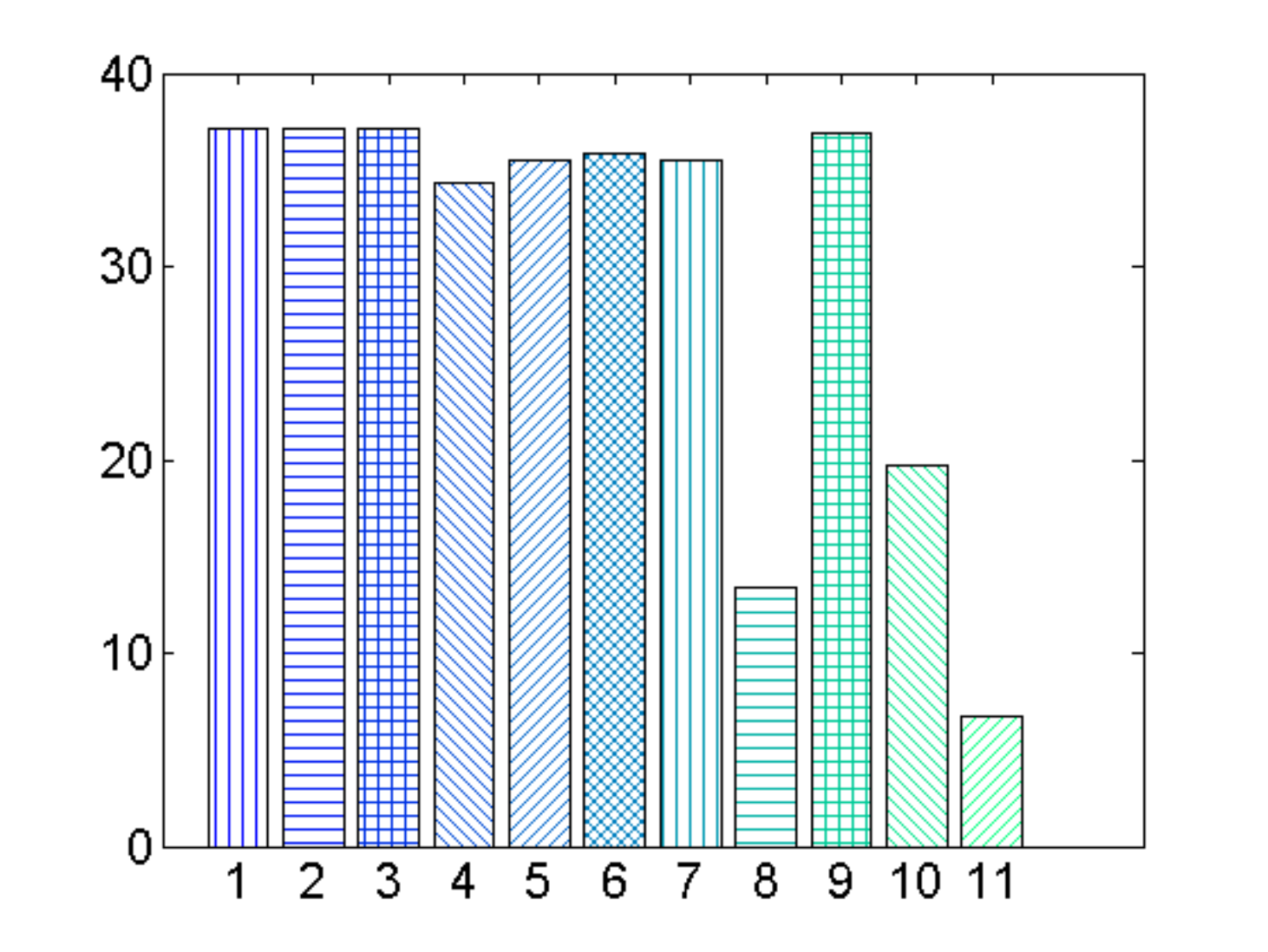}}

\caption{
{\small Classification Error of RMNL using different values of parameter $(\alpha + \nu)$. The $x$-axis represents the value of $(\alpha + \nu)$,
and the $y$-axis represents the \% error.
}
}
\label{fig:exp_alpha}
\vspace*{\alphaVSpace}
\end{figure}

\paragraph{Robustness to Parameter Tuning}
Our algorithm requires extra input parameters $\alpha$ and $\nu$.
There may be indirect ways to set their values,
for example, by estimating the size of the smallest target cluster.
Still, we are not aware of any efficient algorithm to compute the approximately correct values.
Since these parameters play an important role in our algorithm, it is crucial to show the robustness of the algorithm to parameter tuning.
Note that the two parameters are always used together as the additive term  $(\alpha + \nu)$, thus essentially the algorithm takes one parameter.
So for evaluation, we vary the parameter $(\alpha + \nu)$ linearly and run our algorithm over these values.

Figure~\ref{fig:exp_alpha} shows the performance of the algorithm for different parameter values.
We observe that the algorithm does not require the exact value of $(\alpha+\nu)$ as it shows good performance over a continuous range of values.
The range is sufficiently large for all data sets except Iris.
The range for Iris is relatively small as there is little noise in it, and thus the parameter should be set to small values.
In the other datasets we tried, we observed that it is easy to land in the right range with only a few runs.

\subsubsection{Inductive Setting} \label{sec:exp:inductive}

In this subsection, we present the evaluation results in the inductive setting.
In this setting, the algorithm generates a hierarchy on a small random sample of the data set,
and inserts the remaining points to generate a new hierarchy over the entire data set.
We repeat the sampling and evaluation for $5$ times and report the average results.

We compare our inductive algorithm with the random sample algorithm in~\citep{eriksson12RS}.
These algorithms sample some fraction of the similarities and use only these similarities.
The percentage of sampled similarities can be tuned in these algorithms,
so we compare their performance when they use the same amount of sampled similarities.

\begin{figure}[t!]
\vspace*{-.4in}
\hspace*{-1in}
\includegraphics[width=1.3\textwidth]{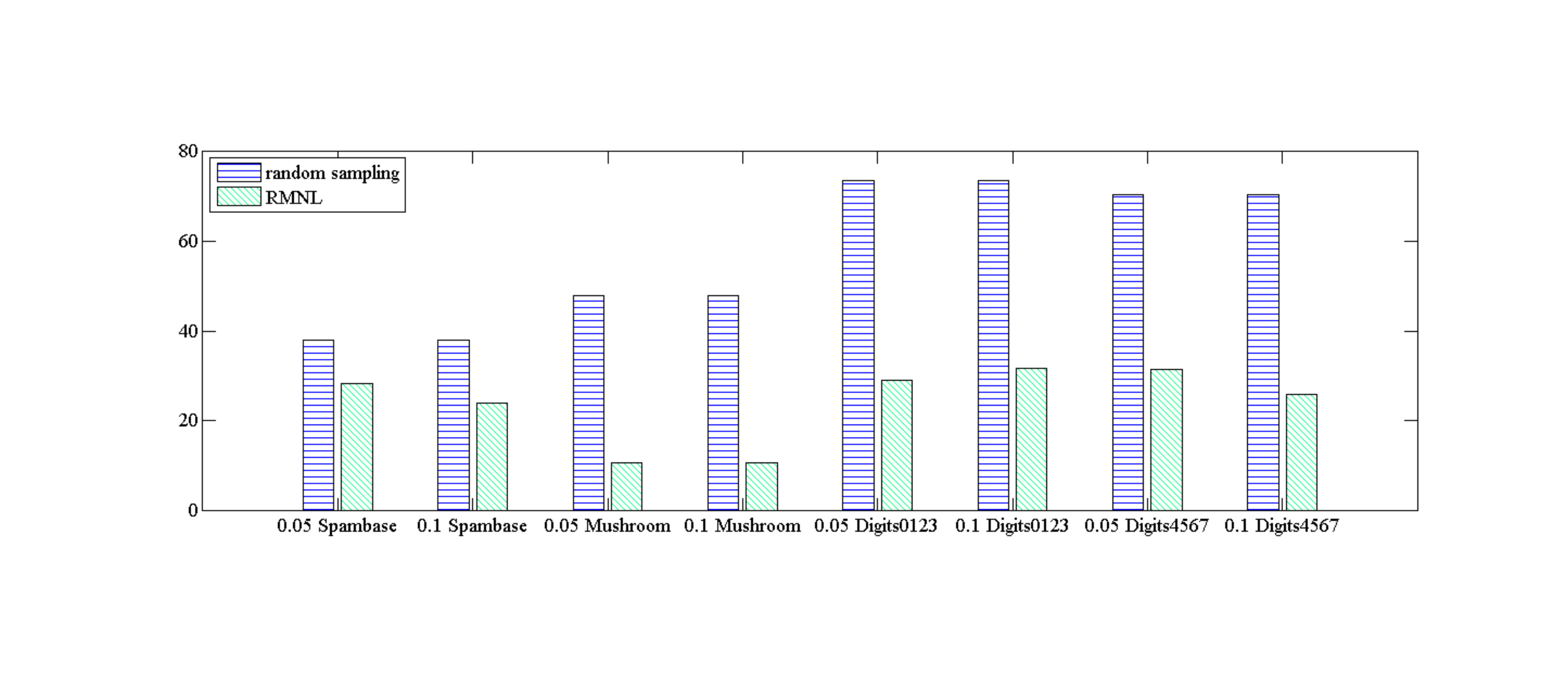}
\vspace*{-.8in}
\caption{
{\small Classification Error of different algorithms in the inductive setting. The $y$-axis represents the \% error.
The $x$-axis represents data sets, where the numbers before the names of the data sets denote the fraction of similarities used by the inductive algorithms.}
}
\label{fig:inductive_res}
\end{figure}

\begin{figure}[t!]
\hspace{-1in}
\includegraphics[width=1.3\textwidth]{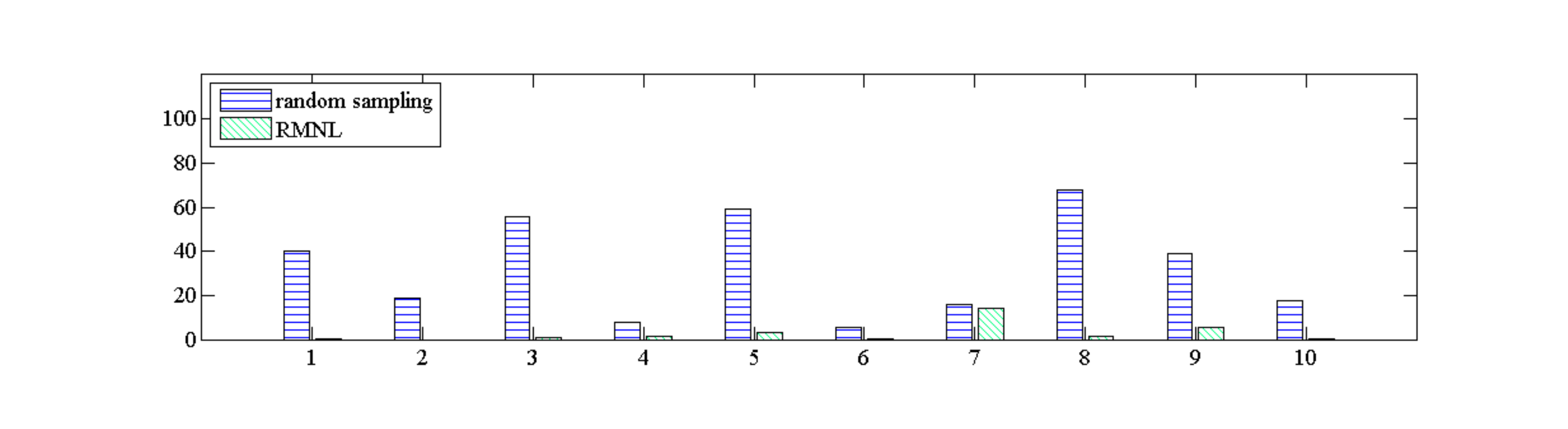}
\vspace*{-.4in}
\caption{{\small Classification Error on {PFAM1} to {PFAM10} data sets using $2.5\%$ similarities.
The $y$-axis in each case represents the \% error, and the $x$-axis represents data sets.}}
\label{fig:res_inductive_pfam}
\end{figure}

Figure~\ref{fig:inductive_res} shows the results for eight configurations (using $5\%$ or $10\%$ similarities on four different data sets).
Our algorithm consistently outperforms the random sampling algorithm.
Figure~\ref{fig:res_inductive_pfam} shows the results on PFAM1 to PFAM10, which approximately satisfy the good neighborhood property~\citep{Voevodski:2012:ACB}.
On all PFAM data sets, the errors of our algorithm are low while those of the random sample algorithm are much higher.
This shows the significant advantage of our algorithm when the data approximately satisfies the good neighborhood property.

\section{Discussion}\label{sec:dis}
In this work we propose and analyze a new
robust algorithm for bottom-up agglomerative clustering. We
show that our algorithm can be used to cluster accurately in
cases where the data satisfies a number of natural properties
and  where the traditional agglomerative algorithms fail. In
particular, if the data satisfies the good neighborhood
properties, the algorithm will be successful in generating a
hierarchy such that the target clustering is close to a pruning
of that hierarchy.

We also show how to extend our algorithm to the inductive setting, where the given data is only a small random sample of the entire data set.
Our algorithm achieves similar correctness guarantees, requiring only a small random sample whose size is independent of that of the entire data set.

We empirically show that with appropriate tuning of the noise parameters our algorithm consistently performs better than other hierarchical algorithms and are more robust to noise in the data.
We also show the efficacy of the inductive version of our algorithm as a faster alternative
when evaluation over the complete data is resource intensive.

Additionally, our subsequent work~\citep{community2013} showed
that the algorithm can be applied to the closely related
community detection task and compares favorably with existing
approaches.

It would be interesting to see if our algorithmic approach can be shown to work for other natural
properties on the input similarity function.  For example, it would be particularly interesting to
analyze a noisy version of the max stability property in~\citet{BBV08} which was shown to
be a necessary and sufficient condition for single linkage to succeed, or of the average stability
property which was shown to be a sufficient condition for average linkage to succeed.  It would
also be interesting to identify other natural conditions under different types of algorithms which
are known to provide empirical noise robustness (e.g., the Ward’s method) would provably succeed.
Finally, from an experimental point of view, an interesting open question is whether one can
provide a wrapper for the algorithm to eliminate the need for manual tuning of the noise
parameters.

\paragraph{Acknowledgments} We thank Avrim Blum for numerous useful discussions, and Konstantin Voevodski for providing us the PFAM data sets. We also thank the reviewers for their helpful comments and suggestions.

This work was supported in part by NSF grant CCF-0953192, AFOSR grant FA9550-09-1-0538, ONR grant N00014-09-1-0751,  a Google Research Award, and a Microsoft Faculty Fellowship.

\bibliographystyle{plainnat}
\bibliography{jmlr-ref}

\begin{thebibliography}{40}
\providecommand{\natexlab}[1]{#1}
\providecommand{\url}[1]{\texttt{#1}}
\expandafter\ifx\csname urlstyle\endcsname\relax
  \providecommand{\doi}[1]{doi: #1}\else
  \providecommand{\doi}{doi: \begingroup \urlstyle{rm}\Url}\fi

\bibitem[Ackerman et~al.(2013)Ackerman, Ben-David, Loker, and
  Sabato]{ackerman2013clustering}
M.~Ackerman, S.~Ben-David, D.~Loker, and S.~Sabato.
\newblock Clustering oligarchies.
\newblock In \emph{Proceedings of the International Conference on Artificial
  Intelligence and Statistics}, 2013.

\bibitem[Altschul et~al.(1990)Altschul, Gish, Miller, Myers, and
  Lipman]{blast:altsbl90}
S.~F. Altschul, W.~Gish, W.~Miller, E.~W. Myers, and D.~J. Lipman.
\newblock Basic local alignment search tool.
\newblock \emph{Journal of Molecular Biology}, 1990.

\bibitem[Bache and Lichman(2013)]{Frank+Asuncion:2010}
K.~Bache and M.~Lichman.
\newblock {UCI} machine learning repository, 2013.
\newblock URL \url{http://archive.ics.uci.edu/ml}.

\bibitem[Balcan and Liang(2013)]{community2013}
M.~F. Balcan and Y.~Liang.
\newblock Modeling and detecting community hierarchies.
\newblock In \emph{Proceedings of the International Workshop on
  Similarity-Based Pattern Analysis and Recognition}, 2013.

\bibitem[Balcan et~al.(2008)Balcan, Blum, and Vempala]{BBV08}
M.~F. Balcan, A.~Blum, and S.~Vempala.
\newblock {A discriminative framework for clustering via similarity functions}.
\newblock In \emph{Proceedings of the Annual ACM symposium on Theory of
  Computing}, 2008.

\bibitem[Balcan et~al.(2013)Balcan, Blum, and Gupta]{BBG09}
M.~F. Balcan, A.~Blum, and A.~Gupta.
\newblock Clustering under approximation stability.
\newblock \emph{Journal of ACM}, 2013.

\bibitem[Blum et~al.(2007)Blum, Coja-Oghlan, Frieze, and
  Zhou]{blum2007separating}
A.~Blum, A.~Coja-Oghlan, A.~Frieze, and S.~Zhou.
\newblock Separating populations with wide data: A spectral analysis.
\newblock In \emph{Algorithms and Computation}. 2007.

\bibitem[Bryant and Berry(2001)]{aprezian01}
D.~Bryant and V.~Berry.
\newblock A structured family of clustering and tree construction methods.
\newblock \emph{Advances in Applied Mathematics}, 2001.

\bibitem[Chaudhuri and Dasgupta(2010)]{Sanjoy2010}
K.~Chaudhuri and S.~Dasgupta.
\newblock {Rates of convergence for the cluster tree}.
\newblock \emph{Advances in Neural Information Processing Systems}, 2010.

\bibitem[Cheng et~al.(2006)Cheng, Kannan, Vempala, and Wang]{CKVW06}
D.~Cheng, R.~Kannan, S.~Vempala, and G.~Wang.
\newblock A divide-and-merge methodology for clustering.
\newblock \emph{ACM Transaction on Database Systems}, 2006.

\bibitem[Dasgupta and Long(2005)]{hierarchysanjoy05}
S.~Dasgupta and P.~Long.
\newblock Performance guarantees for hierarchical clustering.
\newblock \emph{Journal of Computer and System Sciences}, 2005.

\bibitem[Duda et~al.(2000)Duda, Hart, and Stork]{dudahart2000}
R.O. Duda, P.E. Hart, and D.G. Stork.
\newblock \emph{{Pattern classification}}.
\newblock 2000.

\bibitem[Eriksson(2012)]{eriksson12RS}
B.~Eriksson.
\newblock Hierarchical clustering using randomly selected measurements.
\newblock In \emph{Proceedings of the IEEE Statistical Signal Processing
  Workshop}, 2012.

\bibitem[Everitt et~al.(2011)Everitt, Landau, Leese, and
  Stahl]{everitt2011cluster}
B.~S. Everitt, S.~Landau, M.~Leese, and D.~Stahl.
\newblock \emph{Hierarchical Clustering}.
\newblock 2011.

\bibitem[Feldman et~al.(2008)Feldman, O'Donnell, and
  Servedio]{feldman2008learning}
J.~Feldman, R.~O'Donnell, and R.~A. Servedio.
\newblock Learning mixtures of product distributions over discrete domains.
\newblock \emph{SIAM Journal on Computing}, 2008.

\bibitem[Gallegos(2002)]{gallegos2002maximum}
M.~T. Gallegos.
\newblock Maximum likelihood clustering with outliers.
\newblock In \emph{Classification, Clustering, and Data Analysis}. Springer,
  2002.

\bibitem[Gallegos and Ritter(2005)]{gallegos2005robust}
M.~T. Gallegos and G.~Ritter.
\newblock A robust method for cluster analysis.
\newblock \emph{The Annals of Statistics}, 2005.

\bibitem[Garc{\'\i}a-Escudero and Gordaliza(1999)]{garcia1999robustness}
L.~Garc{\'\i}a-Escudero and A.~Gordaliza.
\newblock Robustness properties of k means and trimmed k means.
\newblock \emph{Journal of the American Statistical Association}, 94\penalty0
  (447), 1999.

\bibitem[Garc{\'\i}a-Escudero et~al.(2008)Garc{\'\i}a-Escudero, Gordaliza,
  Matr{\'a}n, and Mayo-Iscar]{garcia2008general}
L.~Garc{\'\i}a-Escudero, A.~Gordaliza, C.~Matr{\'a}n, and A.~Mayo-Iscar.
\newblock A general trimming approach to robust cluster analysis.
\newblock \emph{The Annals of Statistics}, 2008.

\bibitem[Garc{\'\i}a-Escudero et~al.(2010)Garc{\'\i}a-Escudero, Gordaliza,
  Matr{\'a}n, and Mayo-Iscar]{garcia2010review}
L.~Garc{\'\i}a-Escudero, A.~Gordaliza, C.~Matr{\'a}n, and A.~Mayo-Iscar.
\newblock A review of robust clustering methods.
\newblock \emph{Advances in Data Analysis and Classification}, 4\penalty0
  (2-3), 2010.

\bibitem[Gollapudi et~al.(2006)Gollapudi, Kumar, and Sivakumar]{ravik06}
S.~Gollapudi, R.~Kumar, and D.~Sivakumar.
\newblock Programmable clustering.
\newblock In \emph{Symposium on Principles of Database Systems}, 2006.

\bibitem[Gower(1967)]{gower67}
J.~C. Gower.
\newblock {A comparison of some methods of cluster analysis}.
\newblock \emph{Biometrics}, 1967.

\bibitem[Guha et~al.(1998)Guha, Rastogi, and Shim]{Guha_cure:an98}
S.~Guha, R.~Rastogi, and K.~Shim.
\newblock {CURE: an efficient clustering algorithm for large databases}.
\newblock In \emph{ACM SIGMOD Record}, 1998.

\bibitem[Hennig(2008)]{hennig2008dissolution}
C.~Hennig.
\newblock Dissolution point and isolation robustness: robustness criteria for
  general cluster analysis methods.
\newblock \emph{Journal of multivariate analysis}, 99\penalty0 (6), 2008.

\bibitem[Jain and Dubes(1981)]{Jain81dubes.algorithms}
A.~K. Jain and R.~C. Dubes.
\newblock Algorithms for clustering data, 1981.

\bibitem[Jain et~al.(1999)Jain, Murty, and Flynn]{jain1999data}
A.K. Jain, M.N. Murty, and P.J. Flynn.
\newblock Data clustering: a review.
\newblock \emph{ACM computing surveys}, 1999.

\bibitem[Johnson(1967)]{Johnson}
S.~C. Johnson.
\newblock {Hierarchical clustering schemes}.
\newblock \emph{Psychometrika}, 1967.

\bibitem[King(1967)]{KING67step-wiseclustering}
B.~King.
\newblock {Step-wise clustering procedures}.
\newblock \emph{Journal of the American Statistical Association}, 1967.

\bibitem[Kuhn(1955)]{kuhn1955hungarian}
H.~W. Kuhn.
\newblock {The Hungarian method for the assignment algorithm}.
\newblock \emph{Naval Research Logistics Quarterly}, 1955.

\bibitem[LeCun et~al.(1998)LeCun, Bottou, Bengio, and
  Haffner]{lecun1998gradient}
Y.~LeCun, L.~Bottou, Y.~Bengio, and P.~Haffner.
\newblock Gradient-based learning applied to document recognition.
\newblock \emph{Proceedings of the IEEE}, 1998.

\bibitem[Meil{\u{a}} and Heckerman(2001)]{meila_hecker01}
M.~Meil{\u{a}} and D.~Heckerman.
\newblock {An experimental comparison of model-based clustering methods}.
\newblock \emph{Machine Learning}, 2001.

\bibitem[Moitra and Saks(2013)]{moitra2013polynomial}
A.~Moitra and M.~Saks.
\newblock A polynomial time algorithm for lossy population recovery.
\newblock In \emph{Proceddings of the IEEE Annual Symposium on Foundations of
  Computer Science}, 2013.

\bibitem[Nagy(1968)]{Nagy:1968}
G.~Nagy.
\newblock State of the art in pattern recognition.
\newblock \emph{Proceedings of the IEEE}, 1968.

\bibitem[Narasimhan et~al.(2006)Narasimhan, Jojic, and Bilmes]{qcluster2005}
M.~Narasimhan, N.~Jojic, and J.~Bilmes.
\newblock {Q-clustering}.
\newblock \emph{Advances in Neural Information Processing Systems}, 2006.

\bibitem[Punta et~al.(2012)Punta, Coggill, Eberhardt, Mistry, Tate, Boursnell,
  Pang, Forslund, Ceric, Clements, Heger, Holm, Sonnhammer, Eddy, Bateman, and
  Finn]{pfam:Punta01012012}
M.~Punta, P.~C. Coggill, R.~Y. Eberhardt, J.~Mistry, J.~Tate, C.~Boursnell,
  N.~Pang, K.~Forslund, G.~Ceric, J.~Clements, A.~Heger, L.~Holm, E.~L.~L.
  Sonnhammer, S.~R. Eddy, A.~Bateman, and R.~D. Finn.
\newblock The pfam protein families database.
\newblock \emph{Nucleic Acids Research}, 2012.

\bibitem[Sneath and Sokal(1973)]{Sneath73numericaltaxonomy}
P.~H.~A. Sneath and R.~R. Sokal.
\newblock \emph{{Numerical taxonomy. The principles and practice of numerical
  classification.}}
\newblock 1973.

\bibitem[Voevodski et~al.(2012)Voevodski, Balcan, R\"{o}glin, Teng, and
  Xia]{Voevodski:2012:ACB}
K.~Voevodski, M.~F. Balcan, H.~R\"{o}glin, S.-H. Teng, and Y.~Xia.
\newblock Active clustering of biological sequences.
\newblock \emph{Journal of Machine Learning Research}, 2012.

\bibitem[Ward(1963)]{Ward1963}
J.~H. Ward.
\newblock Hierarchical grouping to optimize an objective function.
\newblock \emph{Journal of the American Statistical Association}, 1963.

\bibitem[Wigderson and Yehudayoff(2012)]{wigderson2012population}
A.~Wigderson and A.~Yehudayoff.
\newblock Population recovery and partial identification.
\newblock In \emph{Proceedings of the IEEE Annual Symposium on Foundations of
  Computer Science}, 2012.

\bibitem[Wishart(1969)]{Wishart69modeanalysis}
D.~Wishart.
\newblock {Mode analysis: a generalization of nearest neighbour which reduces
  chaining effects}.
\newblock \emph{Numerical Taxonomy}, 1969.

\end{thebibliography}

\appendix

\section{Implementation Details of Algorithm~\ref{alg:RMNL}} \label{app:Implement}

\renewcommand{\tw}{\quad}
\begin{algorithm}[!p]
\caption{Implementation Details of Robust Median Neighborhood
Linkage}\label{alg:RMNL_impl}
\begin{algorithmic}
\STATE {\bf Input:} Similarity function $\simm$ on a set of
points $S$, $n = |S|$, $\alpha > 0, \nu >0$. \STATE \STATE
\textbf{Step 1} \bw For each point, sort the other points
increasingly according to the distances. \STATE \sbw Initialize
$t = \tinit$, and $\mathcal{C}'_{t-1}$ to be a list of
singleton \sets. \STATE \sbw \textbf{while}
$|\mathcal{C}'_{t-1}|>1$ \textbf{do}
    \STATE \textbf{Step 2} \bw \tw $\mybullet$ {\tt Build a graph $F_t$ on the points in $S$ as follows:}
    \STATE \sbw \tw Set $I_t(x,y) = 1$ if $y$ is in $x$'s $t$ nearest neighbors; $I_t(x,y) = 0$ otherwise.
    \STATE \sbw \tw Set $N_t = I_t (I_t)^T$. 
    \STATE \sbw \tw Set $F_t(x, y) = 1$ if $N_t(x,y) \geq t - 2(\alpha+\nu)n$; $F_t(x,y) = 0$ otherwise.
    
    \STATE \textbf{Step 3} \bw \tw $\mybullet$ {\tt Build a graph $H_t$ on the \sets\ in $\mathcal{C}'_{t-1}$ as follows:}
    \STATE \sbw \tw Set $N\!S_t = F_t (F_t)^T$.
    \STATE \sbw \tw Set $FC_t(x,y) = 1$ if $x,y$ are in the same \set\ in $\mathcal{C}'_{t-1}$ and $F_t(x,y) = 1$; \\
           \sbw \tw  $~~~~~FC_t(x,y) = 0$ otherwise.
    \STATE \sbw \tw Set $S_t = F_t (FC_t)^T + FC_t (F_t)^T$.
    \STATE \sbw \tw \textbf{for} any $C_u, C_v \in \mathcal{C}'_{t-1}$ \textbf{do}
              \STATE \sbw \tw \tw \textbf{if }{$C_u=\{x\}$ and $C_v=\{y\}$} \textbf{then}
              \STATE \sbw \tw \tw \tw $H_t(C_u,C_v) = 1$ if $N\!S_t(x,y) > (\alpha+\nu) n$;
              \STATE \sbw \tw \tw \tw $H_t(C_u,C_v) = 0$ otherwise.
              \STATE \sbw \tw \tw \textbf{else}
              \STATE \sbw \tw \tw \tw $H_t(C_u,C_v) = 1$ if $\median_{x \in C_u, y \in C_v} S_t(x, y) > \frac{|C_u| + |C_v|}{4}$;
              \STATE \sbw \tw \tw \tw $H_t(C_u,C_v) = 0$ otherwise.
              \STATE \sbw \tw \tw \textbf{end if}
    \STATE \sbw \tw \textbf{end for}
    \STATE \textbf{Step 4} \bw \tw $\mybullet$ {\tt Merge \sets\ }
    \STATE \sbw \tw \textbf{while}{ $\exists C_u, C_v$ with $H_t(C_u,C_v)=1$ and $|C_u|+|C_v| > 4(\alpha+\nu)n$} \textbf{do}
          \STATE \sbw \tw \tw Find the pair $C_u, C_v$ with maximum $\median_{x \in C_u, y \in C_v} \frac{S_t(x, y)}{|C_u|+|C_v|}$.
          \STATE \sbw \tw \tw Merge the pair $C_u,C_v$, and update $\mathcal{C}'_{t-1}$. Recompute $FC_t, S_t$ and $H_t$.
    \STATE \sbw \tw \textbf{end while}
    
    \STATE \textbf{Step 5} \bw \tw $\mybullet$ {\tt Merge singletons}
    \STATE \sbw \tw \textbf{while}{ $\exists$ component $V$ in $H_t$ with $|\cup_{C\in V} C| \geq 4(\alpha+\nu)n$} \textbf{do}
          \STATE \sbw \tw \tw Merge \sets\ in $V$, and update $\mathcal{C}'_{t-1}$. Recompute $FC_t, S_t$ and $H_t$.
    \STATE \sbw \tw \textbf{end while}
    
    \STATE \textbf{Step 6} \bw \tw $\mybullet$ {\tt Speed up}
    \STATE \sbw \tw \textbf{if}{ $\exists$ less than $\max\{4(\alpha + \nu)n, t/2\}$ singleton \sets\ in $\mathcal{C}'_{t-1}$} \textbf{then}
        \STATE \sbw \tw \tw Merge each singleton with the non-singleton \set\ of highest median
        \STATE \sbw \tw \tw \tw similarity.
        \STATE \sbw \tw \tw Update $\mathcal{C}'_{t-1}$. Recompute $FC_t, S_t$ and $H_t$.
    \STATE \sbw \tw \textbf{end if}
    \STATE \sbw \tw $\mathcal{C}'_t=\mathcal{C}'_{t-1}$.
    
    \STATE \textbf{Step 7} \bw $\mybullet$ {\tt Increase threshold}
    \STATE \sbw \tw $t=t+1$.
\STATE \sbw \textbf{end while} \STATE \STATE {\bf Output:}
Tree $T$ with single points as leaves and internal nodes
corresponding to the merges performed.
\end{algorithmic}
\end{algorithm}

Here we give the full details of the implementation of
Algorithm~\ref{alg:RMNL}.

First, we need some auxiliary data structures to build the
graphs $F_t$ and $H_t$. See Algorithm~\ref{alg:RMNL_impl} for
the definitions of these data structures.

Second, we specify the order of merging clusters. In the merge
step in Algorithm~\ref{alg:RMNL}, the \sets\ in a sufficiently
large connected component of $H_t$ can be merged in arbitrary
order. In our implementation, we merge two connected $C_u,C_v$
in $H_t$ such that they are not both singleton \sets\ and they
have maximum $\median_{x \in C_u, y \in C_v} S_t(x,
y)/(|C_u|+|C_v|)$ (so that we are most confident about merging
them). Then we merge singleton clusters.

Third, for practical purposes, we can slightly modify the
algorithm to speed it up on practical instances~\footnote{This
does not change the time complexity and the correctness, but we
observe that it helps speed up practical instances.}. When
there are less than $4(\alpha + \nu)n$ singleton \sets, we know
that they cannot be merged together into one \set. So we can
simply merge each singleton \set\ with the non-singleton \set\
that has the highest median similarity. This will correctly
assign all but bad points under the good neighborhood
properties. Similarly, when the number of singleton \sets\ is
less than half the current threshold, we can safely merge each
singleton \set\ with the non-singleton \set\ that has the
highest median similarity.

\section{Additional Proofs for Section~\ref{sec:inductive}}\label{app:ind_good_weak}

Here we provide the details for proving that
Algorithm~\ref{alg:ind_RMNL} also succeeds for the weak good
neighborhood. First, by a similar argument as that in
Lemma~\ref{lem:ind_good0}, we can prove
Lemma~\ref{lem:ind_good_weak0} showing that for a fixed $p$ in
$X\setminus B$ and fixed $x\in A_p$, the first condition of the
weak good neighborhood is still satisfied on a sufficiently
large sample (Recall the definition of
Property~\ref{prop:weak}). Similarly, we can prove
Lemma~\ref{lem:ind_good_weak1} showing that the second
condition of the weak good neighborhood is also satisfied.
Then, the similarity $\simm$ satisfies the weak good neighborhood
property with respect to the clustering induced over the sample
(Lemma~\ref{lem:ind_good_weak}). Our final guarantee,
Theorem~\ref{thm:ind_rmnl_weak}, then follows from the lemmas.

\begin{lemma}
\label{lem:ind_good_weak0}
 Let $\simm$ be a symmetric similarity function satisfying the
weak $(\alpha, \beta, \nu)$-\neighbprop\ for the clustering
problem $(X,\ell)$. Consider any fixed $p \in  X\setminus B$
and any fixed $x \in A_p$. If the sample size satisfies
$\dimmn= \Theta\left(\frac{1}{\alpha} \ln{\frac{1}{\delta}}
\right)$, then with probability at least $1-\delta$, $x$ has at
most $2\alpha n$ neighbors outside $A_p \cap S$ out of the
$|A_p \cap S|$ nearest neighbors in $S \setminus B$.
\end{lemma}

\begin{lemma}
\label{lem:ind_good_weak1}
 Let $\simm$ be a symmetric similarity function satisfying the
weak $(\alpha, \beta, \nu)$-\neighbprop\ for the clustering
problem $(X,\ell)$. Consider any fixed $p \in  X\setminus B$
and any fixed good point $x \in A_p$. If the sample size satisfies $\dimmn= \Theta\left(\frac{1}{\alpha}
\ln{\frac{1}{\delta}} \right)$, then with probability at least
$1-\delta$, $x$ has at most $2\alpha n$ neighbors outside $C(x)
\cap S$ out of the $|A_p \cap S|$ nearest neighbors in $S
\setminus B$.
\end{lemma}

\noindent \textbf{Lemma~\ref{lem:ind_good_weak}} {\it Let
$\simm$ be a symmetric similarity function satisfying the weak
$(\alpha, \beta, \nu)$-\neighbprop\ for the clustering problem
$(X,\ell)$. Furthermore, it satisfies that for any $p \in
X\setminus B$, $|A_p| > 24 (\alpha+\nu)N$. If the sample size satisfies $\dimmn=
\Theta\left(\frac{1}{\min(\alpha,\nu)} \ln{\frac{1}{\delta
\min(\alpha,\nu)}} \right)$, then with probability at least
$1-\delta$, $\simm$ satisfies the $(2\alpha,
\frac{15}{16}\beta, 2\nu)$-\neighbprop\ with respect to the
clustering induced over the sample $S$. }

\begin{proof}
Consider the first condition of the weak \neighbprop\ property.
First, by Chernoff bounds, when $n\geq
\frac{3}{\nu}\ln\frac{4}{\delta}$, we have that with
probability at least $1-\delta/4$, at most $2\nu n$ bad points
fall into the sample. Next, by Lemma~\ref{lem:ind_good_weak0}
and union bound, when $\dimmn= \Theta\left(\frac{1}{\alpha}
\ln{\frac{n}{\delta}} \right)$ we have that with probability at
least $1-\delta/4$, for any point $p \in S\setminus B$, any
point $x \in A_p \cap S$ has at most $2\alpha n$ neighbors
outside $A_p \cap S$ out of the $|A_p \cap S|$ nearest
neighbors in $S\setminus B$. Since $|A_p| > 24(\alpha+\nu) N$,
we also have $|A_p \cap S| > 12(\alpha + \nu)n$ with
probability at least $1-\delta/4$. So the first condition of
the weak \neighbprop\ property is satisfied.

Now consider the second condition. Fix $C_i$ and a point $p\in
(C_i\setminus B)\cap S$. When $\dimmn=
\Theta\left(\frac{1}{\alpha} \ln{\frac{n}{\delta}} \right)$,
with probability at least $1-\delta/(8n)$, at least
$\frac{15}{16}\beta$ fraction of points $x$ in $A_p \cap S$
have all but at most $\alpha  N$ nearest neighbors from
$C_i\setminus B$ out of their $|C_i\setminus B|$ nearest
neighbors in $X\setminus B$. Fix such a point $x\in A_p \cap
S$. By Lemma~\ref{lem:ind_good_weak1}, with probability at
least $1-\delta/(8n^2)$, it has all but at most $2\alpha  n$
nearest neighbors from $(C_i\setminus B) \cap S$ out of their
$|(C_i\setminus B) \cap S|$ nearest neighbors in $S\setminus
B$. By union bound, we have that with probability at least
$1-\delta/4$, for any $C_i$ and any $p \in C_i\setminus B$, at
least $\frac{15}{16}\beta$ fraction of points in $A_p \cap S$
have all but at most $2 \alpha  n$ nearest neighbors from $(C_i
\setminus B) \cap S$ out of their $|(C_i \setminus B) \cap S|$
nearest neighbors in $S\setminus B$. So the second condition is
also satisfied.

Therefore, if $\dimmn= \Theta\left(\frac{1}{\min(\alpha,\nu)}
\ln{\frac{n}{\delta}} \right)$, then with probability at least
$1 - \delta$, the similarity function satisfies the $(2\alpha,
2\nu)$-\neighbprop\ property with respect to the clustering
induced over the sample $S$. The lemma then follows from the
fact that  $\dimmn= \Theta\left(\frac{1}{\min(\alpha,\nu)}
\ln{\frac{1}{\delta \min(\alpha,\nu)}} \right)$ implies
$\dimmn= \Theta\left(\frac{1}{\min(\alpha,\nu)}
\ln{\frac{n}{\delta}} \right)$.
\end{proof}

\noindent \textbf{Theorem~\ref{thm:ind_rmnl_weak}} {\it Let
$\simm$ be a symmetric similarity function satisfying the weak
$(\alpha, \beta, \nu)$-\neighbprop\ for the clustering problem
$(X,\ell)$ with $\beta\geq \frac{14}{15}$. Furthermore, it
satisfies that for any $p \in X\setminus B$, $|A_p| > 24
(\alpha+\nu)N$. Then \algref{alg:ind_RMNL} with parameters
$\dimmn= \Theta\left(\frac{1}{\min(\alpha,\nu)} \ln{
\frac{1}{\delta\cdot\min(\alpha,\nu)}} \right)$ produces a
hierarchy with a pruning that is $(\nu+\delta)$-close to the
target clustering with probability $1-\delta$.
}

\begin{proof}
Note that by \lemref{lem:ind_good_weak}, with probability at
least $1-\delta/4$, we have that $\simm$ satisfies the weak
$(2\alpha, \frac{15}{16}\beta,2\nu)$-\neighbprop\ with respect
to the clustering induced over the sample. Then by
Theorem~\ref{thm:RMNL}, \algref{alg:RMNL} outputs a hierarchy
$T$ on the sample $S$ with a pruning $\{u_1,\dots,u_k\}$ such
that $u_i \setminus B = (C_i\cap S) \setminus B$.

Now we want to show that $f_{u_1}, \dots, f_{u_k}$ have error
at most $\nu+\delta$ with probability at least $1-\delta/2$.
For convenience, let $u(x)$ be a shorthand of $u_{\ell(x)}$.
Then it is sufficient to show that with probability at least
$1-\delta/2$, a $(1-\delta)$ fraction of points $x \in X
\setminus B$ have $f_{u(x)}(x)=1$. Fix $C_i$ and a point $x \in
C_i\setminus B$. By \lemref{lem:ind_good_weak0}, with
probability at least $1-\delta^2/2$, out of the $|A_x \cap S|$
nearest neighbors of $x$ in $S\setminus B$, at most $2\alpha n$
can be outside $A_x$. Then out of the $6(\alpha + \nu) n$
nearest neighbors of $x$ in $S$, at most $2(\alpha+\nu)n$
points are outside $A_x\cap S$. By Lemma~\ref{lem:blob}, $u_i$
contains $A_x\cap S$, so $u_i$ must contain at least
$4(\alpha+\nu)n$ points in $N_S(x)$. Consequently, any ancestor
$w$ of $u_i$, including $u_i$, has more points in $N_S(x)$ than
any other sibling of $w$. Then we must have $f_w(x) =1$ for any
ancestor $w$ of $u_i$. In particular, $f_{u_i}(x)=1$. So, for
any point $x\in X\setminus B$, with probability at least
$1-\delta^2/2$ over the draw of the random sample,
$f_{u(x)}(x)=1$. By Markov inequality, with probability at
least $1-\delta/2$, a $(1-\delta)$ fraction of points $x \in X
\setminus B$ have $f_{u(x)}(x)=1$.
\end{proof}

\section{Strict Separation and Ward's Method}\label{app:ward}

\begin{figure}[h]
\centering
\includegraphics[scale = 0.8]{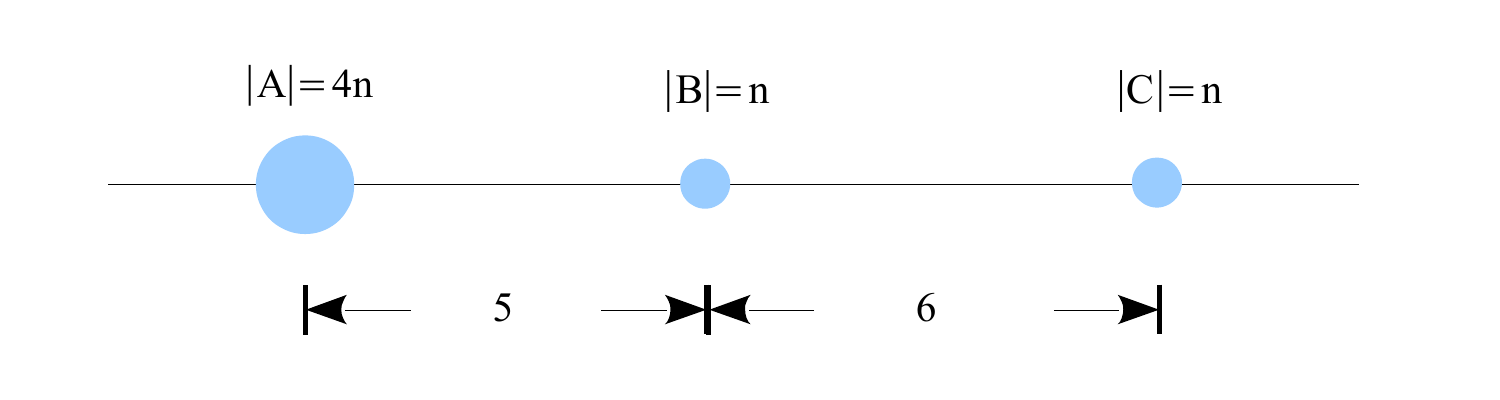}
 \caption{
 {\small
An example that satisfies the strict separation property but is not clustered successfully by Ward's minimum variance Method.
}
}
\label{fig:ward}
\end{figure}

Here we describe an example showing that Ward's minimum variance method fails in the presence of unbalanced clusters. The clustering instance satisfies the strict separation property and thus the more general good neighborhood properties, but on this instance Ward's method leads to large classification error.

The instance is presented in Figure~\ref{fig:ward}. It consists three groups of points on a line: Group $A$ has $4n$ points, Group $B$ has $n$ points, and Group $C$ has $n$ points. The distances between points in the same groups are $0$, while the distances between points in $A$ and points in $B$ are $5$, the distances between points in $B$ and points in $C$ are $6$, the distances between points in $A$ and points in $C$ are $11$.

It can be verified that the clustering $\{A \cup B, C\}$ satisfies the strict separation property. We now show that Ward's method will produce a tree that do not have this clustering as a pruning, and thus fails to cluster the instance. Recall that Ward's method starts with each point being a singleton cluster and at each step finds the pair of clusters that leads to minimum increase in total within-cluster variance after merging.  Formally, it merges the two clusters $U$ and $V$ such that 
$$
	(U,V) = \mathrm{argmin} \left[\mathrm{Var}(U\cup V) - \mathrm{Var}(U) -\mathrm{Var}(V) \right]
$$
where 
$$
	\mathrm{Var}(X) = \min_{c} \sum_{p \in X} \|p-c\|_2^2.
$$
Since the distances between points in the same groups are $0$, the method will first merge points in the same groups and forms three clusters $A, B$, and $C$. Now, merging $A$ and $B$ increases the variance by $20 n$, while merging $B$ and $C$ increases  the variance by $18 n$. Therefore, $B$ and $C$ will be merged, and thus the best pruning in the tree produced is $\{A, B\cup C\}$.  This leads to an error of $1/6 \approx 16.7\%$.

\end{document}